\PassOptionsToPackage{table, x11names}{xcolor}

\documentclass{article} %
\usepackage[table]{xcolor}
\usepackage{iclr2025_conference,times}

\usepackage{amsmath,amsfonts,bm}

\def\eqref#1{equation~\ref{#1}}

\def\1{\bm{1}}

\DeclareMathAlphabet{\mathsfit}{\encodingdefault}{\sfdefault}{m}{sl}
\SetMathAlphabet{\mathsfit}{bold}{\encodingdefault}{\sfdefault}{bx}{n}

\providecommand{\todo}[1]{{\protect\color{cyan}{[TODO: #1]}}}
\providecommand{\danqi}[1]{{\protect\color{orange}{[Danqi: #1]}}}
\providecommand{\tianyu}[1]{{\protect\color{blue}{[Tianyu: #1]}}}
\providecommand{\howard}[1]{{\protect\color{purple}{[Howard: #1]}}}

\providecommand{\todo}[1]{{\protect\color{red}{}}}
\providecommand{\danqi}[1]{{\protect\color{orange}{}}}
\providecommand{\tianyu}[1]{{\protect\color{blue}{}}}
\providecommand{\howard}[1]{{\protect\color{purple}{}}}

\providecommand{\rebuttal}[1]{#1}

\newcommand{\cmark}{{\protect\color{green}{\ding{51}}}}
\newcommand{\xmark}{{\protect\color{purple}{\ding{55}}}}

\newcommand\tf[1]{\textbf{#1}}
\newcommand\ttt[1]{\texttt{#1}}

\renewcommand{\paragraph}[1]{\vspace{0.2cm}\noindent\textbf{#1}}

\newcommand{\ours}{\textsc{HELMET}}
\newcommand{\oursfull}{\textsc{\textbf{HELMET}} (\textbf{H}ow to \textbf{E}valuate \textbf{L}ong-context \textbf{M}odels \textbf{E}ffectively and \textbf{T}horoughly)}
\newcommand{\nmodels}{59}

\newcommand{\mls}{Multi-LexSum}

\newcommand{\infbench}{$\infty${\textsc{Bench}}}

\newcommand{\base}{$^\flat$}
\newcommand{\inst}{}
\newcommand{\ruler}{RULER}

\newcommand{\subem}{SubEM}

\usepackage{hyperref}
\usepackage{url}
\usepackage{booktabs}
\usepackage{multirow}
\usepackage{booktabs}
\usepackage{pifont}
\usepackage{wrapfig}
\usepackage{cleveref}

\usepackage[pdftex]{graphicx}

\usepackage{tcolorbox}
\usepackage{enumitem}
\usepackage{array}

\usepackage{lipsum} %
\usepackage{placeins} %

\usepackage{xcolor}
\usepackage{caption}
\hypersetup{
    colorlinks,
    linkcolor={red!50!black},
    citecolor={blue!50!black},
    urlcolor={blue!80!black}
}

\title{HELMET: How to Evaluate Long-Context Language Models Effectively and Thoroughly}

\author{
    Howard Yen$^{\hspace{.1em}{\color{orange}\boldsymbol{p}}}$$\;$ Tianyu Gao$^{\hspace{.1em}{\color{orange}\boldsymbol{p}}}$$\;$ Minmin Hou$^{\hspace{.1em}{\color{blue}\boldsymbol{i}}}$$\;$ Ke Ding$^{\hspace{.1em}{\color{blue}\boldsymbol{i}}}$
    \\
    ~\tf{Daniel Fleischer}$^{\hspace{.1em}{\color{blue}\boldsymbol{i}}}$$\;$ \tf{Peter Izsak}$^{\hspace{.1em}{\color{blue}\boldsymbol{i}}}$$\;$ \tf{Moshe Wasserblat}$^{\hspace{.1em}{\color{blue}\boldsymbol{i}}}$$\;$ \tf{Danqi Chen}$^{\hspace{.1em}{\color{orange}\boldsymbol{p}}}$\\
    $^{\color{orange}\boldsymbol{p}}$Princeton Language and Intelligence, Princeton University\quad $^{\color{blue}\boldsymbol{i}}$Intel\\
    \texttt{\{hyen,tianyug,danqic\}@cs.princeton.edu}\\
    \centering
    \url{https://princeton-nlp.github.io/HELMET}
}

\iclrfinalcopy %
\begin{document}

\maketitle

\begin{abstract}
  Many benchmarks exist for evaluating long-context language models (LCLMs), yet developers often rely on synthetic tasks such as needle-in-a-haystack (NIAH) or an arbitrary subset of tasks.
However, it remains unclear whether these benchmarks reflect the diverse downstream applications of LCLMs, and such inconsistencies further complicate model comparison.
    We investigate the underlying reasons behind these practices    
    and find that existing benchmarks often provide noisy signals due to limited coverage of applications, insufficient context lengths, unreliable metrics, and incompatibility with base models. 
    In this work, we introduce \oursfull, a comprehensive benchmark encompassing seven diverse, application-centric categories. 
    We also address several issues in previous benchmarks by adding controllable lengths up to 128K tokens,
    model-based evaluation for reliable metrics,
    and few-shot prompting for robustly evaluating base models. 
    Consequently, we demonstrate that \ours{} offers more reliable and consistent rankings of frontier LCLMs. 
    Through a comprehensive study of \nmodels{} LCLMs, we find that
    (1)~synthetic  tasks like NIAH do not reliably predict downstream  performance;
    (2)~the diverse categories in \ours{} exhibit distinct trends and low correlations with each other; and
    (3)~while most LCLMs achieve perfect NIAH scores,  
    open-source models significantly lag behind closed ones when tasks require full-context reasoning or following complex instructions---the gap widens as length increases.
    Finally, we recommend using our RAG tasks for fast model development, as they are easy to run and better predict other downstream performance; ultimately, we advocate for a holistic evaluation across diverse tasks.\footnote{Our data and code are available at \url{https://github.com/princeton-nlp/HELMET}.}

\end{abstract}

\section{Introduction}

\begin{wraptable}{r}{0.5\textwidth}
    \vspace{-12pt}
    \caption{
        Most LCLMs evaluate on synthetic (Syn.) tasks. \base{}: base models. 
        $\infty$: \infbench{}.
    }
    \vspace{-5pt}
    \setlength{\tabcolsep}{0.7pt}
    \small
    \renewcommand{\arraystretch}{0.7} %
    \centering
    \resizebox{0.99\linewidth}{!}{
        \begin{tabular}{lccccccccccc}
            \toprule
            \multirow{2}{*}[-0.5ex]{Model} & \multirow{2}{*}[-0.5ex]{Syn.} & \multirow{2}{*}[-0.5ex]{PPL} & \multicolumn{2}{c}{{$\infty$}} & \multicolumn{5}{c}{\textsc{ZeroSCROLLS}} & \multirow{2}{*}[-0.5ex]{RAG} & \multirow{2}{*}[-0.5ex]{ICL} \\
               \cmidrule(lr){4-5} \cmidrule(lr){6-10}
            {} & {} & {} & {QA} & {All} & {NQA} & {QS} & {QL} & {SQ} & {All} & {} & {} \\
            \midrule
            Gemini-1.5\inst & \cmark & \cmark & \xmark & \xmark & \xmark & \xmark & \xmark & \xmark & \xmark & \xmark & \cmark \\
            GPT-4\inst & \cmark & \xmark & \xmark & \xmark & \xmark & \xmark & \xmark & \xmark & \xmark & \xmark & \xmark \\
            Claude-3.5\inst & \cmark & \xmark & \xmark & \xmark & \xmark & \xmark & \xmark & \xmark & \xmark & \xmark & \xmark \\
            \midrule
            Llama-3.1\inst & \cmark & \xmark & \cmark & \xmark & \xmark & \cmark & \cmark & \cmark & \xmark & \xmark & \xmark \\
            Phi-3\inst & \cmark & \xmark & \xmark & \xmark & \xmark & \cmark & \xmark & \cmark & \xmark & \xmark & \xmark \\
            Jamba-1.5\inst & \cmark & \xmark & \cmark & \xmark & \xmark & \xmark & \xmark & \xmark & \xmark & \xmark & \xmark \\
            Qwen2\inst & \cmark & \xmark & \xmark & \xmark & \xmark & \xmark & \xmark & \xmark & \xmark & \xmark & \xmark \\
            Command R\inst & \cmark & \xmark & \xmark & \xmark & \xmark & \xmark & \xmark & \xmark & \xmark & \cmark & \xmark\\
            \citeauthor{xiong2023effective}\inst & \xmark & \xmark & \xmark & \xmark & \cmark & \cmark & \cmark & \cmark & \cmark & \xmark & \xmark \\
            \midrule
            \citeauthor{chen2023extending}\base & \cmark & \cmark & \xmark & \xmark & \xmark & \xmark & \xmark & \xmark & \xmark & \xmark & \xmark \\
            \citeauthor{peng2024yarn}\base & \cmark & \cmark & \xmark & \xmark & \xmark & \xmark & \xmark & \xmark & \xmark & \xmark & \xmark \\
            \citeauthor{fu2024data}\base & \cmark & \cmark & \cmark & \xmark & \xmark & \xmark & \xmark & \xmark & \xmark & \xmark & \xmark \\
 
            \bottomrule
        \end{tabular}
    }
    \label{tab:model_eval_comparison}
     \vspace{-10pt}
\end{wraptable} 

Long-context language models (LCLMs) unlock 
a myriad of applications, from summarizing long documents to 
learning new tasks on the fly with thousands of examples.
Many recent benchmarks have sought to evaluate language models' long-context abilities~\citep[][\emph{inter alia}]{zhang2024inftybenchextendinglongcontext,an-etal-2024-l,shaham-etal-2023-zeroscrolls,bai-etal-2024-longbench}.
However, recent developments in long-context processing \citep{chen2023extending,xiong2023effective,peng2024yarn,fu2024data} still rely either on perplexity or on synthetic needle-in-a-haystack tasks~(NIAH; \citealp{gkamradt_llmtest_needleinahaystack_2024,hsieh2024ruler}).
Frontier LCLMs \citep{dubey2024llama3herdmodels,jambateam2024jamba15hybridtransformermambamodels,openai2023gpt4,geminiteam2024geminifamilyhighlycapable} also mostly report NIAH, sometimes with arbitrary subsets of other datasets, as shown in Table \ref{tab:model_eval_comparison}.
Such inconsistencies complicate comparisons between different models.
It is also unclear whether synthetic NIAH tasks or the few chosen datasets offer a holistic picture of models' long-context abilities and their performance in real-world applications.

\textit{Why don't model developers agree on these evaluations}? 
  We take a closer look and find that existing benchmarks suffer from many critical design flaws, including:
  
\begin{itemize}[left=5pt]
    \item \textbf{Insufficient coverage of downstream tasks}: Existing benchmarks either focus on synthetic tasks \citep{hsieh2024ruler} or include only simple question answering \citep{zhang2024inftybenchextendinglongcontext}. 
    Other works study particular aspects of LCLMs, %
    such as summarization \citep{chang2024booookscore}, in-context learning \citep{li2024longiclbench}, and retrieval-augmented generation (RAG; \citealp{lee2024longcontextlanguagemodelssubsume}), but they do not provide a holistic evaluation of LCLMs. 
    
    \item \textbf{Inadequate lengths}: Most natural language datasets in existing benchmarks (\citealp{shaham-etal-2023-zeroscrolls,an-etal-2024-l}; Table \ref{tab:datasets_length}) are too short to effectively test frontier long-context abilities (usually $\geq$128K).
    \item \textbf{Unreliable metrics}: 
    For commonly used long-document QA and summarization tasks,
    most existing benchmarks still rely on metrics like ROUGE~\citep{lin-2004-rouge}, which are often noisy and unreliable \citep{goyal2023newssummarizationevaluationera,deutsch-etal-2022-examining,chang2024booookscore}.
    \item \textbf{Incompatibility with base models}: 
    Many LCLM developments focus on base models without instruction tuning,
    but most existing benchmarks require models to be instruction-tuned---hence developers can only rely on synthetic tasks or perplexity. 
\end{itemize}

\begin{figure}[t!]
    \centering
    \includegraphics[width=0.98\linewidth]{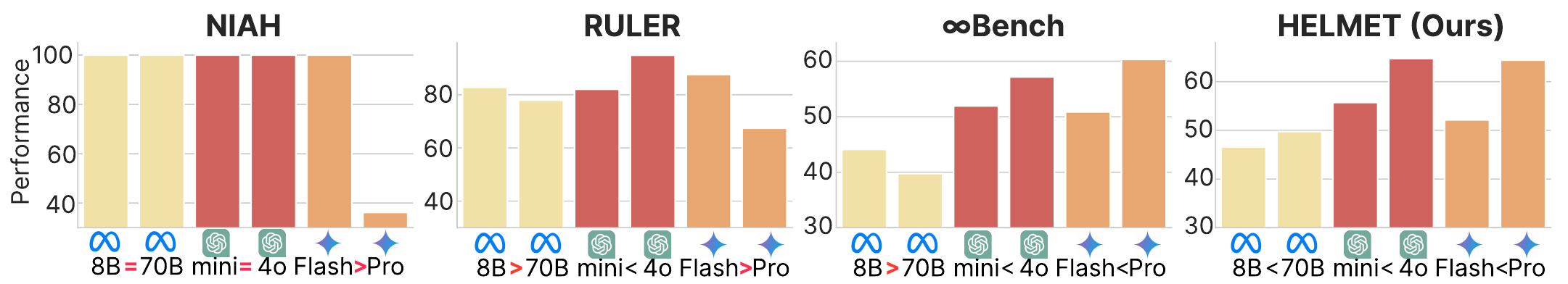}
    \vspace{-2.0pt}
    \caption{
        Long-context benchmark results of frontier LCLMs (Llama-3.1 8B/70B, GPT-4o-mini, GPT-4o-08-06, and Gemini-1.5 Flash/Pro) at 128K input length. 
        NIAH is saturated for almost all models; 
        RULER \citep{hsieh2024ruler} and \infbench{} \citep{zhang2024inftybenchextendinglongcontext} show unexpected trends for Llama-3.1 \citep{dubey2024llama3herdmodels}.
        In contrast, \ours{} demonstrates more consistent rankings of these frontier models.
    }
    \label{fig:comparison_teaser}
\end{figure}

Consequently, existing benchmarks are either not applicable to long-context works (inadequate lengths or incompatibility) or provide highly noisy signals (insufficient coverage or unreliable metrics). 
We summarize the shortcomings of  existing benchmarks in \Cref{tab:dataset_comparison}. For the three benchmarks that support a 128K\footnote{Throughout the paper, we use binary prefixes K$=2^{10}$.} context length (NIAH, \ruler{}, and \infbench{}), we use them to evaluate six frontier models and report the numbers in \Cref{fig:comparison_teaser}.
We see that NIAH does not reflect differences across  models; 
\ruler{} and \infbench{} show unexpected trends---on \ruler{}, Gemini Flash outperforms Gemini Pro; 
on \infbench{}, the 70B Llama model underperforms compared to the 8B one.
This raises concerns about the reliability of the benchmarks and how they can differentiate long-context models, which likely contributes to the lack of use of these benchmarks in model development.

To address these challenges, we present \oursfull{}.
We curate a \emph{diverse} set of application-centric long-context tasks across seven categories. 
Beyond widely adopted categories like \emph{synthetic recall}, \emph{long-document question answering (QA)}, and \emph{summarization}, we also add \emph{many-shot in-context learning} (ICL), \emph{retrieval-augmented generation} (RAG), \emph{passage re-ranking}, and \emph{generation with citations} \citep{gao2023alce}.
We further 
address the shortcomings of existing benchmarks: 
(1) we ensure all datasets support input lengths of 128K tokens and are easily extendable to longer contexts (\S\ref{dataset:overview}); 
(2) we introduce reference-based model evaluation for QA and summarization and show that it significantly improves over n-gram overlap metrics (\S\ref{dataset:metrics});
(3) we refine the prompts and in-context demonstrations used for all tasks, reducing evaluation noise caused by different output formats and allowing base models to be evaluated robustly across most categories (\S\ref{dataset:settings}).
Together, \ours{} enables a holistic evaluation of long-context capabilities and 
provides more reliable signals for model development.
\Cref{fig:comparison_teaser} demonstrates that
\ours{} can clearly differentiate models with varying  capabilities and reflect comparisons consistent with human perception.
\rebuttal{More discussion on HELMET's improvements over previous benchmarks and direct comparisons can be found in \S\ref{app:related_works_compare}.}

To understand the progress of LCLMs and how different long-context capabilities correlate with one another, we evaluate a comprehensive list of \nmodels{} LCLMs of various architectures, scales, and training approaches.
Our analysis reveals that
(1) synthetic tasks poorly indicate  ownstream performance (\S\ref{results:synthetic}),
(2) different categories in \ours{}
show distinct trends %
(\S\ref{results:realistic}), and
(3) 
open-source models significantly lag behind closed ones on tasks that require
reasoning over long contexts or following complex instructions---the gap further widens as context length increases (\S\ref{results:complex}).
Finally, we find that the RAG category strikes a good balance between ease of use, stronger correlation with downstream tasks, and compatibility with base models.
Ultimately, it is imperative to evaluate LCLMs across a diverse spectrum of categories.
We hope that our insights provide a more effective way to evaluate LCLMs for future model development and benchmarking.

\begin{table}[t!]
    \caption{
        Comparison of long-context benchmarks: ZeroSCROLLS \citep{shaham-etal-2023-zeroscrolls}, LongBench \citep{bai-etal-2024-longbench}, L-Eval \citep{an-etal-2024-l}, RULER \citep{hsieh2024ruler}, \infbench{} \citep{zhang2024inftybenchextendinglongcontext}, and our  \ours.
        $L$: input tokens.
        $^\dagger$: All datasets have $L < 128$K except one dataset. %
        $^\ddag$: L-Eval uses LLMs to compute reference-free, pairwise win-rates; we design reference-based model evaluation for specific tasks.  
    }
    \setlength{\tabcolsep}{2pt}
    \small
    \centering
        \begin{tabular}{lcccccccccc}
            \toprule
             & \multicolumn{7}{c}{Type of tasks} & \multicolumn{3}{c}{Benchmark features} \\
             \cmidrule(lr){2-8} \cmidrule(lr){9-11}
            & \multirow{2}{*}{Cite} & \multirow{2}{*}{RAG} & \multirow{2}{3em}{\centering Re-rank} & \multirow{2}{3em}{\centering Long-QA} & \multirow{2}{*}{Summ} & \multirow{2}{*}{ICL} & \multirow{2}{4em}{\centering Synthetic Recall} & \multirow{2}{4em}{\centering Robust Eval.} & \multirow{2}{*}{$L \geq$ 128k} & \multirow{2}{4em}{\centering Controll-able $L$} \\
             &  &  &  &  &  &  &  &  &  &  \\ \midrule
            ZeroSCROLLS & \xmark & \xmark & \xmark & \cmark & \cmark & \xmark & \xmark & \xmark & \xmark$^\dagger$ & \xmark \\
            LongBench & \xmark & \cmark & \xmark & \cmark & \cmark & \cmark & \cmark & \xmark & \xmark$^\dagger$ & \xmark \\
            L-Eval & \xmark & \cmark & \xmark & \cmark & \cmark & \xmark & \xmark & \cmark$^\ddag$ & \xmark$^\dagger$ & \xmark \\
            RULER & \xmark & \xmark & \xmark & \xmark & \xmark & \xmark & \cmark & \cmark & \cmark & \cmark \\
            \infbench & \xmark & \xmark & \xmark & \cmark & \cmark & \xmark & \cmark & \xmark & \cmark & \cmark\\
            \midrule
            \ours{} (Ours) & \cmark & \cmark & \cmark & \cmark & \cmark & \cmark & \cmark & \cmark & \cmark & \cmark \\ \bottomrule
        \end{tabular}
    \label{tab:dataset_comparison}
\end{table}

\section{Our Benchmark: \ours{}}

In this work, we seek to overcome the shortcomings of existing benchmarks by meeting the following desiderata: 
(1) diverse coverage across different tasks and capabilities of LCLMs,
(2) %
controllable context lengths that support more than 128K input tokens, 
and (3) reliable evaluation for both base and instruction-tuned models.
In this section, we %
describe the datasets used in \ours{} and 
how they improve upon existing evaluation benchmarks in terms of settings and metrics.
An overview of HELMET is shown in Table \ref{tab:dataset_overview}.

\subsection{Realistic and diverse long-context applications}
\label{dataset:overview}

\paragraph{Retrieval-augmented generation (RAG).}
We use open-domain question answering (ODQA)---which requires retrieving from a knowledge corpus and then generating correct answers \citep{chen-etal-2017-reading}---as a representation of retrieval-augmented generation (RAG) applications. %
We utilize Natural Questions \citep[NQ;][]{kwiatkowski2019natural}, TriviaQA \citep[TQA;][]{joshi2017triviaqa}, HotpotQA \citep{yang-etal-2018-hotpotqa}, and PopQA \citep{mallen-etal-2023-trust}.
We use the gold passage (the passage with the answer) from \citet{petroni-etal-2021-kilt}, or otherwise select any passage that contains the answer. %

Given an input length $L$, we first determine the number of passages $k$ that can fit within $L$ tokens, then retrieve $k$ passages\footnote{We use \href{https://huggingface.co/Alibaba-NLP/gte-large-en-v1.5}{\ttt{Alibaba-NLP/gte-large-en-v1.5}} for retrieval \citep{zhang2024mgtegeneralizedlongcontexttext}.}
from the corpus\footnote{We use Wikipedia 
2019-8-01 dump, split into 100-word passages \citep{petroni-etal-2021-kilt}.} that \textit{do not contain the answer} as \textit{distractors}.
This differs from previous works that \textit{randomly sample passages} from the corpus \citep{lee2024longcontextlanguagemodelssubsume} and is more realistic and challenging.
For NQ, TQA, and PopQA, %
we take the top $k-1$ distractors and insert the gold passage at six evenly distributed positions following \citet{Liu2023LostIT}.
For HotpotQA, which requires two gold passages, we combine them and the top $k-2$ distractors and randomly shuffle them into three permutations.
We use substring exact match (\subem{}; whether the answer is included in the output), 
following previous work \citep{asai2024selfrag}.
See \S\ref{app:datasets:odqa} for more details.

\paragraph{Generation with citations (Cite).}
We leverage ALCE \citep{gao2023alce} to evaluate LCLMs on a realistic application of answering questions while providing correct attributions \citep{bohnet2022attributed}.
Given multi-faceted questions and relevant passages, models are required to generate a long-text answer and cite supporting passage IDs at the end of each sentence. 
This tests models' ability to utilize the passages in the context and also to follow the instructions about citation formats.
We use the ASQA \citep{stelmakh2022asqa} and QAMPARI \citep{rubin2022qampari} subsets from ALCE.
For an input length $L$, we first determine the number of passages $k$, and use the top $k$ retrieved passages from Wikipedia as contexts. %
The model's outputs are evaluated on correctness and citation quality, and we report the average across all metrics.
See \S\ref{app:datasets:cite} for more details.

\begin{table}[t!]
    \caption{
        Overview of evaluation datasets. 
        We select datasets that cover various important long-context capabilities. %
        \subem{}: substring exact match.
    }
    \vspace{-5pt}
    \centering
    \resizebox{0.98\linewidth}{!}{
        \begin{tabular}{llll}
            \toprule
            \textbf{Category} & \textbf{Dataset} & \textbf{Metrics} & \textbf{Description} \\
            \midrule
            \multirow{4}{7em}{\textbf{Retrieval-augmented generation}}& Natural Questions & \subem{} & Factoid question answering \\
            & TriviaQA & \subem{} & Trivia question answering \\
            & PopQA & \subem{} & Long-tail entity question answering \\
            & HotpotQA & \subem{} & Multi-hop question answering \\
            \midrule
            \multirow{2}{7em}{\textbf{Generation with citations}} & ALCE ASQA & Recall, Cite & Answer ambiguous questions with citations \\
            & ALCE Qampari & Recall, Cite & Answer factoid questions with citations \\
            \midrule
            \textbf{Passage re-ranking} & MS MARCO & NDCG@10 & Rerank passage for a query \\
            \midrule
            \multirow{5}{7em}{\textbf{Many-shot in-context learning}} & TREC Coarse & Accuracy & Question type classification, 6 labels \\
            & TREC Fine & Accuracy & Question type classification, 50 labels \\
            & NLU & Accuracy & Task intent classification, 68 labels \\
            & BANKING77 & Accuracy & Banking intent classification, 77 labels \\
            & CLINC150 & Accuracy & Intent classification, 151 labels \\
            \midrule
            \multirow{3}{7em}{\textbf{Long-document QA}} & NarrativeQA & Model-based & Book and movie script QA \\
            & \infbench{} QA & ROUGE F1 & Novel QA with entity replacement \\
            & \infbench{} MC & Accuracy & Novel multiple-choice QA with entity replacement \\
            \midrule
            \multirow{2}{7em}{\textbf{Summarization}} & \infbench{} Sum & Model-based & Novel summarization with entity replacement \\
            & Multi-LexSum & Model-based & Summarizing multiple legal documents \\
            \midrule
           \multirow{4}{7em}{\textbf{Synthetic recall}} & JSON KV & \subem{} & Retrieve a key in JSON dictionary \\
            & RULER MK Needle & \subem{} & Retrieve the needle (a number) within noisy needles \\
            & RULER MK UUID & \subem{} & Retrieve the needle (a UUID) within noisy needles \\
            & RULER MV & \subem{} & Retrieve multiple values for one needle (key)\\
            \bottomrule
        \end{tabular}
    }
    \vspace{-14pt}
    \label{tab:dataset_overview}
\end{table}

\paragraph{Passage re-ranking (Re-rank).}
Re-ranking retrieved passages based on their relevance to the query 
is an important application of LCLMs~\citep{sun-etal-2023-chatgpt}.
The task requires the model to retrieve relevant information, compare, and reason over different parts of the contexts.
We use the MS MARCO dataset \citep{bajaj2018ms}, %
where each instance contains a query and
passages retrieved by BM25 \citep{robertson2009bm25} 
from the Internet. 
Each passage has annotations  of a 
relevance label---perfect, highly relevant, or not relevant.
We determine the number of passages $k$ from the input length $L$, and randomly sample  %
$k$ passages with balanced labels for each test query. 
The model is prompted with the query and the shuffled $k$ passages 
and is instructed to output the top-10 document IDs ranked by relevance. 
We report NDCG@10.
Details are in \S\ref{app:datasets:rerank}.

\paragraph{Many-shot in-context learning (ICL).}
In-context learning (ICL) is a key ability that enables LLMs to adapt to new tasks on the fly \citep{brown2020language}.
Recent studies~\citep{ratner-etal-2023-parallel,xu2024stresstesting,li2024longiclbench,bertsch2024incontext} explore performing many-shot ICL (with thousands of examples) with LCLMs.
Following \citet{bertsch2024incontext}, we focus on datasets with large label spaces: TREC-coarse, TREC-fine \citep{li-roth-2002-learning}, BANKING77 \citep{casanueva-etal-2020-efficient}, CLINC150 \citep{larson-etal-2019-evaluation}, and NLU \citep{Liu2019BenchmarkingNL}.
We adjust the number of shots to control the input length $L$, and the number of examples in each class is balanced. 
We report accuracy on the test set.

One \textit{difference} from previous works is that 
we map original natural language labels (e.g., \textit{location}) into numbered labels (i.e., \textit{0}, \textit{1}) to test how well a model can learn new tasks instead of relying on its pre-trained priors \citep{wei2023largerlanguagemodelsincontext,pan-etal-2023-context,min-etal-2022-rethinking}.
More details are in \S\ref{app:datasets:icl}.

\begin{wraptable}{r}{3.8cm}
    \vspace{0pt}
    \caption{
        Dataset lengths. 
    }
    \label{tab:datasets_length}
    \centering
    \small
    \setlength{\tabcolsep}{0.6pt}
    \begin{tabular}{l@{\hskip -12pt}rr}
        \toprule  
        \textbf{Datasets} & \textbf{Medium~} & \textbf{Max} \\
        \midrule
        \multicolumn{3}{c}{ZeroSCROLLS}\\
        \midrule
        QASPER & 6K~ & 12K \\
        GovReport & 12K~ & 33K \\
        QuALITY & 9K~ & 11K \\
        SQuALITY & 8K~ & 10K \\
        \midrule
        \multicolumn{3}{c}{\ours} \\
        \midrule
        NarrativeQA & 73K~ & 518K \\
        \infbench{} QA & 191K~ & 835K \\
        \infbench{} MC & 167K~ & 835K \\
        \infbench{} Sum & 154K~ & 835K \\
        Multi-LexSum & 90K~ & 5M \\
        \bottomrule
    \end{tabular}
    \vspace{-2em}
\end{wraptable} 

\paragraph{Long-document question answering (LongQA).}
We use NarrativeQA \citep{kocisky2018narrativeqa} and the English book QA and multiple choice (MC) subsets from \infbench{} \citep{zhang2024inftybenchextendinglongcontext} for evaluating long-document QA. %
We select those tasks 
for their abundant context lengths (\Cref{tab:datasets_length}).
We truncate the document from the end based on $L$.
We use ROUGE F1 for \infbench{} QA (answers are mostly entity names) and accuracy for \infbench{} MC. 
For NarrativeQA, where the answers can be long text and open-ended, 
we design and use a model-based evaluation (\S\ref{dataset:metrics}).

\paragraph{Summarization (Summ).}
Summarization tests LCLMs' ability to synthesize information across the contexts. 
We choose \mls{} (legal document summarization) and the English summarization task from \infbench{} (novel summarization)
for their extensive lengths (\Cref{tab:datasets_length}).
We truncate the document from the end based on the evaluation length $L$.
We use our model-based evaluation (\S\ref{dataset:metrics}) for both datasets
instead of the commonly used ROUGE, as it better reflects human judgment.

\paragraph{Synthetic recall.}
Synthetic recall tasks, such as needle-in-a-haystack (NIAH), stress test models' ability to recall relevant information (the ``needle'') from long contexts.
They have gained popularity for being
easy to use (as they can test any arbitrary length) and 
easy to control (can placing the ``needle'' at any position).
For this category,
we select multiple synthetic recall tasks from RULER (an extended version of NIAH; \citealp{hsieh2024ruler}) 
and also add a JSON KV retrieval task \citep{Liu2023LostIT}, which we find more challenging.
We intentionally select the synthetic tasks that correlate well with application-driven tasks; in-depth discussions are in \S\ref{results:synthetic}.
Following previous works, we report the percentage of the ground truth answers that are substrings in the generation (\subem{}).
Refer to \S\ref{app:datasets:synthetic} for more details.

\subsection{Reliable evaluation metrics}
\label{dataset:metrics}

Existing long-context benchmarks \citep{zhang2024inftybenchextendinglongcontext,shaham-etal-2023-zeroscrolls} largely rely on n-gram overlap metrics like ROUGE \citep{lin-2004-rouge},  which have been shown to correlate poorly with human judgment for tasks with long outputs, such as summarization \citep{goyal2023newssummarizationevaluationera,deutsch-etal-2022-examining,krishna-etal-2023-longeval}.
L-Eval \citep{an-etal-2024-l} uses LLMs to score reference-free ``win rates,'' which neglect the available answer annotations
and always require evaluating model pairs.
Instead, we design a \emph{reference-based model evaluation} method for long-document QA and summarization  that is more reliable and easy to use. 

\paragraph{Question answering.}
In NarrativeQA, 
we prompt GPT-4o\footnote{\ttt{GPT-4o-2024-05-13}} 
with the question, the ground truth, and the model output to check for fluency and correctness. 
The fluency score is either 0 (incoherent or repetitive) or 1 (fluent), and the correctness score takes on the value of 0 (incorrect), 1 (partly correct), 2 (correct but not fully relevant), and 3 (correct and relevant).
We take the product of the two as the final score, normalizing it to a range of $[0,100]$.

\paragraph{Summarization.}
Following previous works \citep{kamoi2023wice,zhang-bansal-2021-finding}, we first decompose the gold summary into atomic claims and use GPT-4o to check if each claim is supported by the generation (recall) and if each sentence in the generation is supported by the reference summary (precision).
We then compute the F1 score from the recall and precision scores.
Additionally, we ask GPT-4o to evaluate fluency (0 or 1) and take its product with the F1 score as the final score.
In each step, we prompt GPT-4o with handwritten examples.

\begin{figure}[t!]
    \centering
    \includegraphics[width=0.98\linewidth]{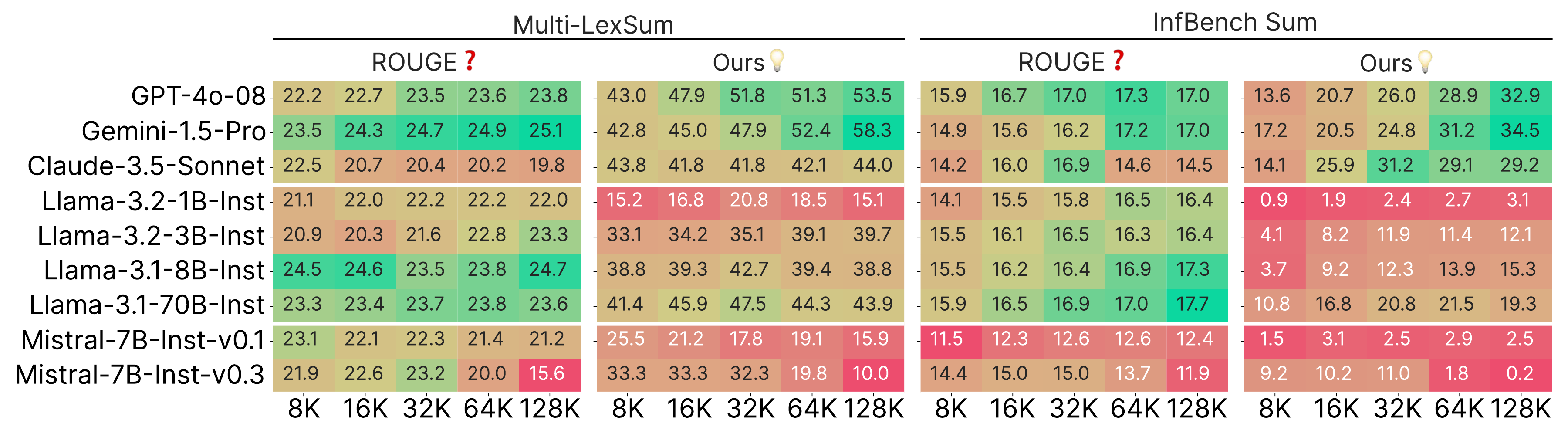}
    \caption{
        Comparison between ROUGE-L F1  and our model-based evaluation metric on summarization tasks.
        Our metric shows more consistent trends: 
        it reflects the performance gain on GPT-4o with increased input length, while ROUGE remains almost the same; our metric also clearly differentiates models while ROUGE shows little distinction.
    }
    \vspace{-5.0pt}
    \label{fig:comparison_metrics}
\end{figure}

Empirically, our reference-based model evaluation reflects more consistent trends, %
as shown in Figure \ref{fig:comparison_metrics}:
(1) Llama-3.1-8B-Inst achieves similar ROUGE scores to GPT-4o, while our evaluation reveals a significant gap.
(2) Our metric better identifies incoherent generations and shows lower performance for models with smaller context windows, such as the Mistral models.
(3) Our metric exhibits a substantially more positive trend for GPT-4o as input length increases, whereas ROUGE-L remains within a 2-point absolute difference.
\rebuttal{We further validate the model-based evaluation through human studies, which suggest our new metrics strongly correlate with human judgments: for example, on \infbench{} Sum, our metric reaches a human-model agreement of Cohen's $\kappa = 0.91$ for summary precision and $\kappa = 0.76$ for recall.}
\rebuttal{More details on the human studies are in \S\ref{app:datasets}.}

\subsection{Robust prompting and controlled evaluation settings}
\label{dataset:settings}

\paragraph{Robust prompting reduces noise and enables evaluation on base models.}
Many long-context benchmarks 
require models to follow instructions and 
only support evaluating instruction-tuned models \citep{zhang2024inftybenchextendinglongcontext,shaham-etal-2023-zeroscrolls}. %
However, many model developments do not incorporate instruction tuning~\citep{chen2023extending,fu2024data}, leaving these models reliant on perplexity-based evaluation or synthetic tasks.
To support long-context research efforts,
we design our benchmark so that at least a subset of the datasets accommodates evaluating base models.

Existing benchmarks mostly use zero-shot prompting \citep{shaham-etal-2023-zeroscrolls,zhang2024inftybenchextendinglongcontext}, which leads to inconsistent output formats, especially for base models.
For example, the model may output a long answer in RAG when a short answer is required.
We add two-shot demonstrations in the prompt for all tasks to 
address this problem.\footnote{Except for ICL (the number of shots varies) and RULER (we follow the original formatting).}
For long-document QA and summarization, we replace the original document with a placeholder phrase to reduce the number of input tokens in the ICL example.
As shown in Table \ref{tab:results_ablate_shots}, both base and instruction-tuned models significantly benefit from the demonstrations. %

Furthermore, we employ the length-instruction-enhanced evaluation from L-Eval for long-generation tasks (i.e., summarization), which has been shown to have substantially more consistent and reliable evaluations \citep{an-etal-2024-l}.
As a result, we find that our reproduction of previous datasets, such as \infbench{} QA tasks, better reflects the capabilities of LCLMs, as shown in Table \ref{tab:results_infbench_compare}. 
The use of demonstrations and improved instructions more accurately depicts how models perform in real applications.

\paragraph{Controlled input length and difficulty.}
An important dimension to consider when evaluating LCLMs is the input length $L$, as longer inputs can provide more information while challenging the model's ability to process distracting contexts.
As we discussed in \S\ref{dataset:overview}, we can control the input length $L$ for each task by either adjusting the number of retrieved passages, the number of demonstrations, or truncating the document text to fit within the specified lengths.
This allows us to study model performance at or beyond the length of current frontier LCLMs ($\geq 128$K). %

As shown in Figure \ref{fig:comparison_teaser} and \ref{fig:comparison_metrics}, \ours{} provides rankings more consistent with human perception of model performance. 
The diverse tasks, controllable lengths, and robust evaluation metrics and prompting enable a thorough examination of  LCLMs across multiple dimensions.

\section{Analysis}
\label{analysis}

We evaluate \nmodels{} LCLMs with \ours{}.
To our best knowledge, this is the most thorough and controlled comparison of long-context models on diverse applications.
These models cover closed-source models, such as GPT-4, Claude, and Gemini, as well as open-source model families, such as Llama \citep{dubey2024llama3herdmodels}, Mistral \citep{jiang2023mistral}, Phi \citep{abdin2024phi3technicalreporthighly}, and Qwen \citep{qwen2025qwen25technicalreport}.
We also consider models that use different architectures---full-attention transformers \citep{vaswani2017attention}, sliding-window attention \citep{Beltagy2020Longformer}, and hybrid models with SSM modules \citep{dao2024transformersssmsgeneralizedmodels}.
We also benchmark position extrapolation models such as 
YaRN \citep{peng2024yarn} and LongRoPE \citep{ding2024longropeextendingllmcontext}. 
We list all the models evaluated in Table \ref{tab:model_detail}.
We evaluate each model at input lengths: $L \in \{8K, 16K, 32K, 64K, 128K\}$, where $L$ is the number of Llama-2 tokens \citep{touvron2023llama2}, and use greedy decoding for all models to ensure consistency.
We randomly sample 100 to 600 examples from each dataset;
more details are in \S\ref{app:setup}.

\subsection{Simple synthetic tasks are poor predictors of real-world performance}
\label{results:synthetic}

\begin{wrapfigure}{r}{0.4\linewidth}
    \vspace{-13pt}
    \centering
    \includegraphics[width=1\linewidth]{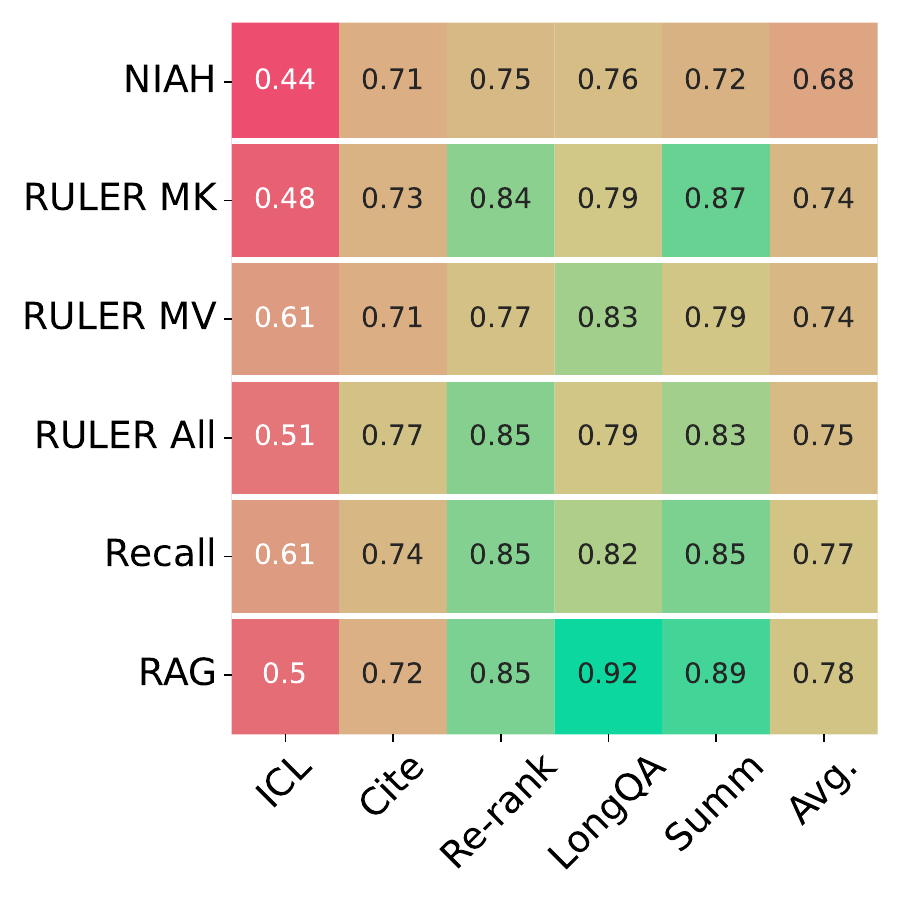}
    \vspace{-20pt}
    \caption{
        Spearman's rank correlation at 128K input length, calculated across 35 instruction-tuned models.
    }
    \vspace{-20pt}
    \label{fig:corr_syn_real}
\end{wrapfigure}
Many model developers rely on simple synthetic tasks, such as NIAH, for evaluating long-context language models, but it is unclear if these tasks accurately represent real-world performance.
To this end, we calculate Spearman's rank correlation $\rho$ between synthetic and real-world tasks for 35 instruction-tuned models.
First, Figure \ref{fig:corr_syn_real} shows that none of the synthetic tasks achieves an average correlation higher than 0.8.
We also make the following observations.

\paragraph{Not all synthetic tasks are created equal.}
The original NIAH, which places a needle in the middle of unrelated essays and asks the model to retrieve it, exhibits weak correlation with real-world tasks: 
all correlations are $\leq0.8$.
Similarly, the popular RULER average score---which includes not only NIAH variants but also synthetic aggregation, multi-hop tracing, and QA---does not yield strong correlations (all $<0.85$).

We take a closer look at different \ruler{} tasks and find that harder recall-type tasks are more reflective of real-world categories---for example, RULER MK, which places distracting needles around the target needle. Despite the overall low correlation, we believe these tasks can still serve as a useful sanity check during model development.
We compile several such RULER tasks, along with JSON KV, to form the HELMET synthetic recall set  (more discussions in \S\ref{app:correlation_syn_real}).

\begin{figure}[t!]
    \centering
    \includegraphics[width=0.9\linewidth]{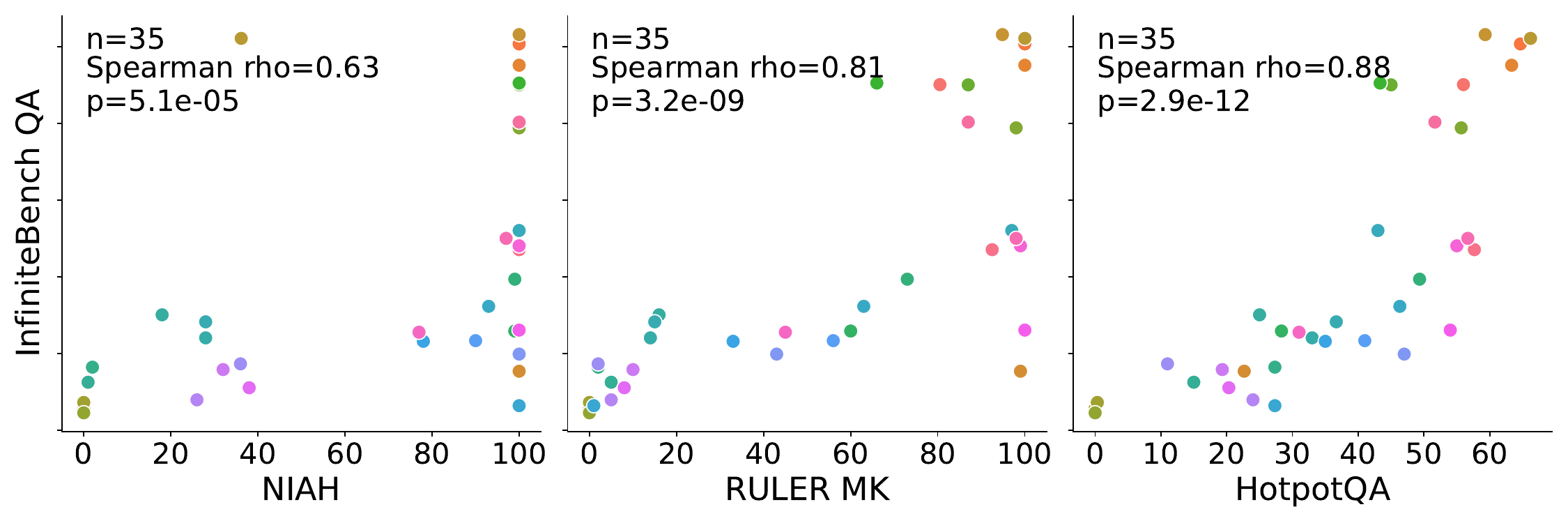}
    \vspace{-10pt}
    \caption{
        Distribution of instruction-tuned models' performance on \infbench{} QA with respect to NIAH, RULER MK, and HotpotQA.
    }
    \label{fig:corr_syn_real_dist}
    \vspace{-10pt}
\end{figure}

\paragraph{Tasks with noisier, more distracting contexts better differentiate models.}
To understand why synthetic tasks exhibit weak correlation with real-world tasks, we 
plot the performance of different models on NIAH, RULER MK (one of our recall tasks), and HotpotQA  (one of our RAG tasks) in Figure \ref{fig:corr_syn_real_dist}.
We use \infbench{} QA as a representative real-world task.
We find that most models achieve either perfect or near-zero performance on the original NIAH, leaving few data points in the middle and resulting in poor separability between models. In contrast, RULER MK, which introduces more distracting contexts, better distributes model performance between 0\% and 100\%, leading to clearer differentiation.

\paragraph{RAG is a better proxy for real-world tasks.}
Finally, we find that RAG datasets, such as HotpotQA, 
consistently achieve higher correlation with other real-world tasks. 
\Cref{fig:corr_syn_real_dist} also shows that HotpotQA exhibits an almost linear relationship with the QA dataset.
Similar to synthetic tasks, RAG tasks are easy to control and assess models’ recall abilities. However, since all passages are retrieved and relevant to the query, RAG contexts are more distracting and therefore harder to saturate.

\subsection{Diverse LCLM applications call for diverse evaluation}
\label{results:realistic}

\begin{wrapfigure}{r}{0.45\linewidth}
    \vspace{-6.0pt}
    \centering
    \includegraphics[width=1\linewidth]{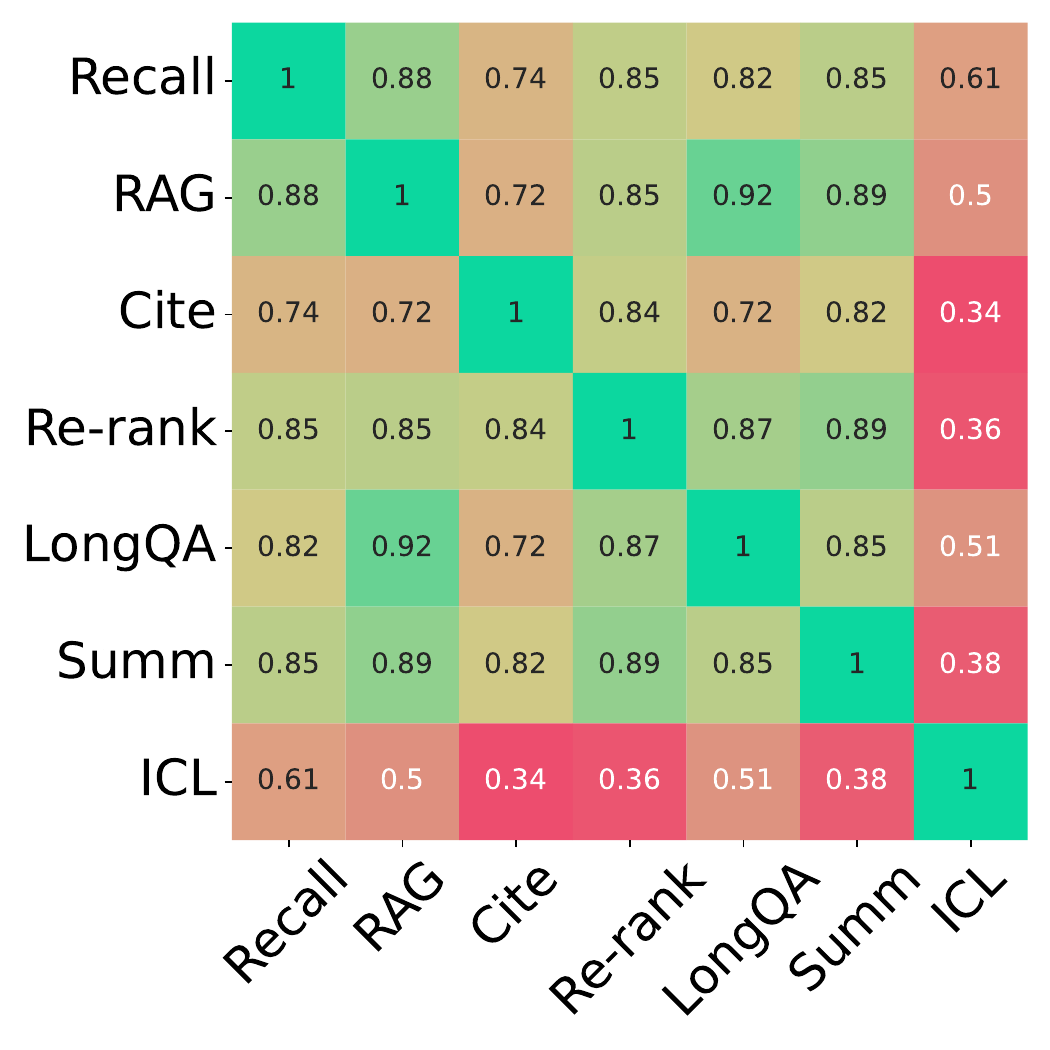}
    \vspace{-18.0pt}
    \caption{
        Spearman rank correlation between different categories at $L=$128K.
    }
    \vspace{-14.0pt}
    \label{fig:corr_category}
\end{wrapfigure}

In long-context language modeling, realistic tasks are often only used in isolation \citep{karpinska2024thousandpairsnovelchallenge,li2024longiclbench,dubey2024llama3herdmodels}, which limits the understanding of LCLMs in a broader context.
In this work, we cross-examine model performance over a wide range of real tasks, and find that different categories do not consistently correlate with each other,  as shown in Figure \ref{fig:corr_category}.

show moderate correlation due to their shared retrieval component. However, the added complexity of generating citations in ALCE results in lower correlation with other categories.Naturally, RAG and passage re-ranking moderately correlate due to the shared retrieval component
As shown in Figure \ref{fig:corr_all}, generating correct answers and producing valid citations are not strongly correlated, suggesting that instruction following and recalling facts within long contexts are distinct capabilities.

Furthermore, some categories---generation with citations and in-context learning---do not correlate well with other categories.
Intuitively, summarization tests for the model's ability to aggregate information across the entire input, while ICL evaluates its ability to learn new tasks from many examples. Such capabilities are  orthogonal to recall facts in long contexts.
Such capabilities are orthogonal to recalling facts in long contexts. Therefore, model developers should evaluate across these distinct axes to form a more holistic understanding of a model’s capabilities (see additional analysis in \S\ref{app:correlation_datasets}).

\begin{figure}[t!]
    \centering
    \includegraphics[width=0.98\linewidth]{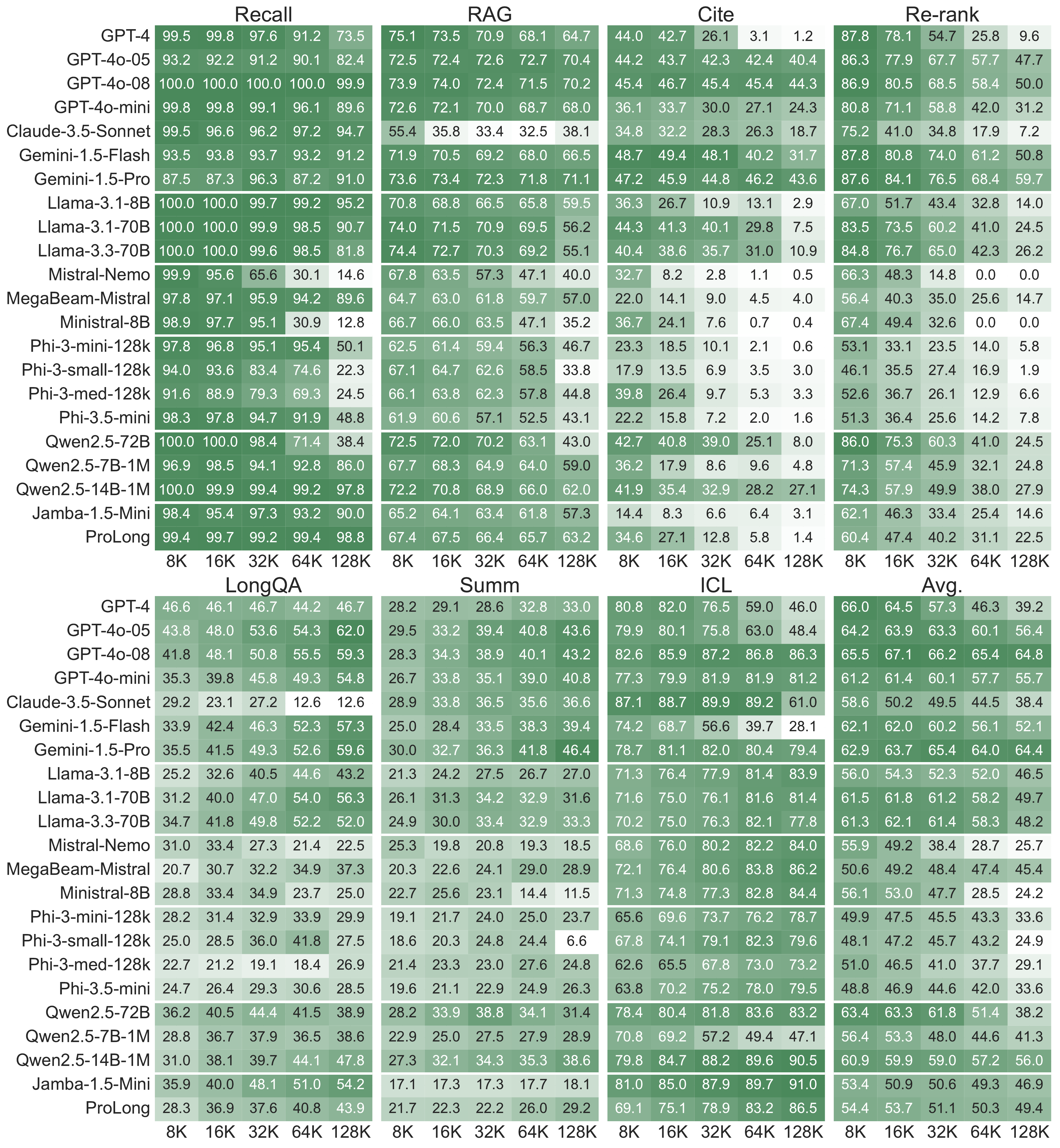}
    \caption{
        Results of \ours{}.
        All models are instruction-tuned and have a claimed context window of 128K tokens or more.
    }
    \vspace{-12.0pt}
    \label{fig:results_length_main}
\end{figure}

\subsection{Model performance across tasks and lengths}
\label{results:complex}

We show the performance of instruction-tuned models on \ours{} at  five different lengths in Figure \ref{fig:results_length_main}, and the
full results are illustrated in Figure \ref{fig:results_length_full}.
We analyze the model performance across two critical dimensions of long-context language modeling: \emph{task complexity} and \emph{input length}.

\paragraph{Open-source models lag behind closed-source models on complex tasks.}
First, we consider the performance of frontier LCLMs at the longest input length of 128K tokens.
We find that the closed-source models---notably GPT-4o-08 and Gemini-1.5-Pro---stand out as the strongest LCLMs.
Other than ICL, the closed-source models outperform the open-source models on all tasks.
The gap is relatively small on synthetic recall and LongQA, where the task is to retrieve information from the context.
There is a stark contrast in the generation with citations and re-ranking performance, where the closed-source models are 30 to 40 absolute points better than the best open-source models.

\begin{figure}[t!]
    \centering
    \includegraphics[width=0.98\linewidth]{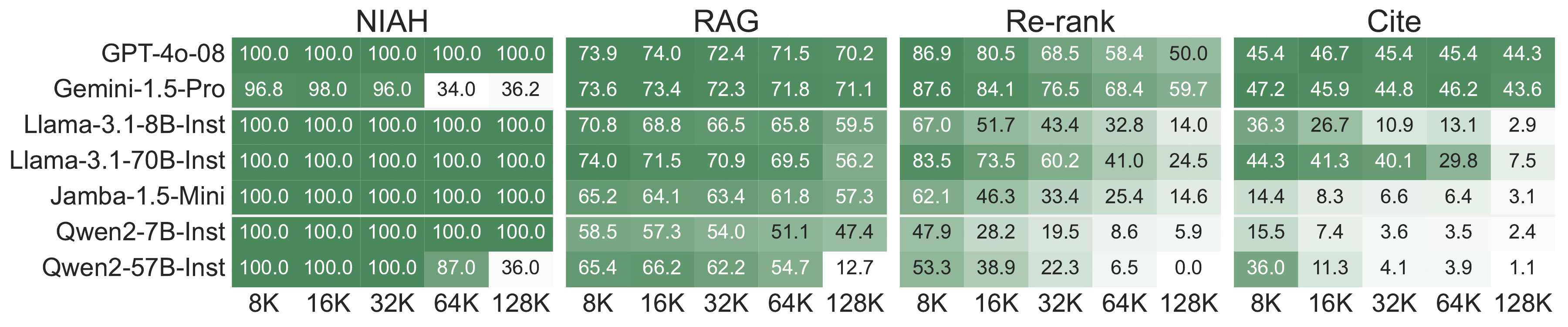}
    \caption{
        Results of selected instruct models on various lengths and increasing complexity of tasks.
        Notably, Qwen2 relies on position extrapolation, while other open-models are trained at or greater than 128K context window.
    }
    \vspace{-5.0pt}
    \label{fig:results_length_select}
\end{figure}

\paragraph{Performance degradation with longer inputs is category-dependent.}
Most frontier models largely retain performance on recall and RAG with longer inputs; however, even the best models experience significant degradation as context length increases on tasks like re-ranking and generation with citations.
As illustrated in Figure \ref{fig:results_length_select}, 
performance degradation at longer lengths becomes more pronounced as task complexity increases from left to right. On generation with citations, open-source models completely collapse at 128K, while GPT-4o remains relatively stable. This underscores the importance of evaluating models on more complex long-context applications.

\paragraph{No clear winner across all categories.}
As we observe from the previous sections, the different categories do not always correlate with each other. This is evident in the varying top-performing models across categories: for instance, GPT-4o excels in recall and generation with citations, while Gemini performs better in passage re-ranking and long-document QA.
Furthermore, many open-source models outperform closed-source models on ICL, potentially because heavy instruction tuning negatively impacts ICL. We provide qualitative examples in \autoref{tab:claude_outputs_icl}.
Thus, evaluating models across multiple axes is essential. 
In the appendix, we also present additional analysis, such as the performance of positional extrapolation methods (\S\ref{app:pos_emb}), the lost-in-the-middle phenomenon (\S\ref{app:depth}), and the performance of Claude (\S\ref{app:claude}).

\section{Related Works}
\paragraph{Long-context language models.}
Frontier models such as GPT-4 \citep{openai2023gpt4}, Gemini \citep{geminiteam2024gemini15unlockingmultimodal}, and Claude claim to have expanded their context window beyond 100K tokens.
In the open-source community, there are also efforts to train models with longer input lengths \citep{dubey2024llama3herdmodels,fu2024data,ai2024yi,gao2024prolong}, explore position  extrapolation techniques \citep{peng2024yarn,chen2023extending,ding2024longropeextendingllmcontext}, and experiment with efficient architectures \citep[][\emph{inter alia}]{Beltagy2020Longformer,bertsch2023unlimiformer,gu2024mamba,dao2024transformersssmsgeneralizedmodels,yen2024longcontext,lieber2024jambahybridtransformermambalanguage}.

\paragraph{Synthetic tasks.}
Synthetic tasks are often used to evaluate LCLMs since they can be procedurally generated, enabling arbitrarily long input lengths and controlled ``needle'' placement \citep{tay2021long,Liu2023LostIT}.
In particular, Needle-in-a-Haystack \citep[NIAH;][]{gkamradt_llmtest_needleinahaystack_2024} inserts a ``needle'' at specific depths of a long essay (i.e., the haystack) and asks the model to recall the fact.
Recent works have expanded upon it and designed new procedures to test different aspects of LCLMs \citep[][\emph{inter alia}]{hsieh2024ruler,li2024needlebenchllmsretrievalreasoning,levy2024task,arora2023zoology,laban2024summaryhaystackchallengelongcontext,goldman2024reallylongcontextneed}.
However, they do not study how results on synthetic tasks transfer to real applications.
In contrast, we evaluate both synthetic and downstream tasks and investigate how they correlate with each other.

\paragraph{Long-context benchmarks.}
As discussed in the main paper, 
existing benchmarks either 
are limit to relatively short context lengths \citep{shaham-etal-2022-scrolls,shaham-etal-2023-zeroscrolls,li-etal-2024-loogle,an-etal-2024-l,bai-etal-2024-longbench,dong-etal-2024-bamboo,kwan-etal-2024-m4le} or lack rigorous evaluation methods \citep{zhang2024inftybenchextendinglongcontext,yuan2024lvevalbalancedlongcontextbenchmark}. 
Many works instead focus on specific domains, such as question answering \citep{wang2024novelqabenchmarkingquestionanswering,karpinska2024thousandpairsnovelchallenge,wang-etal-2024-ada,wang-etal-2024-leave}, in-context learning \citep{li2024longiclbench,bertsch2024incontext,xu2024stresstesting,anil2024many,agarwal2024manyshotincontextlearning}, summarization \citep{chang2024booookscore,kim2024fables,shen2022multilexsum}, or RAG \citep{lee2024longcontextlanguagemodelssubsume}.
In this work, we construct a comprehensive benchmark that tests across diverse downstream tasks at long input lengths and also present a unified comparison and analysis across \nmodels{} LCLMs.

\section{Conclusion}

In this work, we first identify the shortcomings of long-context evaluation settings and existing benchmarks---over-reliance on synthetic tasks, limited coverage of realistic applications, and unreliable metrics among others.
We seek to address these issues by constructing \ours, an application-centric benchmark with diverse domains and reliable evaluation settings.
We then present a comprehensive evaluation of \nmodels{} frontier LCLMs across multiple dimensions, including different tasks, input lengths, and model types.
Our analysis shows that synthetic tasks poorly predict downstream performance and that different categories exhibit distinct capabilities and trends. Thus, evaluating models on a diverse set of real-world tasks is essential.
Furthermore, open-source models still lag behind closed-source models on complex tasks at longer lengths. Finally, we hope that our benchmark and comprehensive evaluation serve as a valuable resource for future research in long-context language modeling.

\subsection*{Reproducibility statement}
We have publicly released all of our code and data to ensure reproducibility.
Experimental settings are outlined in the main text (\S\ref{analysis}) and the appendix (\S\ref{app:setup}), and we will provide additional details and instructions in the code repository.
Furthermore, we plan on releasing the model generated outputs and statistics to enable further analysis and comparison.

\subsubsection*{Acknowledgments}
We would like to acknowledge Mengzhou Xia, Howard Chen, Xi Ye, Yinghui He, Lucy He, Alexander Wettig, Sadhika Malladi, and other members of Princeton Language and Intelligence for their helpful feedback and discussion. 
We also thank Adithya Bhaskar and Joie Zhang for helping with validating the model-based evaluation. 
Tianyu Gao is supported by an IBM PhD Fellowship.
This work is gratefully supported by an NSF CAREER award (IIS-2239290), a grant from Intel, and the Microsoft Accelerate Foundation Models Research (AFMR) grant program.

\bibliography{iclr2025_conference}

\begin{thebibliography}{95}
\providecommand{\natexlab}[1]{#1}
\providecommand{\url}[1]{\texttt{#1}}
\expandafter\ifx\csname urlstyle\endcsname\relax
  \providecommand{\doi}[1]{doi: #1}\else
  \providecommand{\doi}{doi: \begingroup \urlstyle{rm}\Url}\fi

\bibitem[Abdin et~al.(2024)Abdin, Aneja, Awadalla, Awadallah, Awan, Bach, Bahree, Bakhtiari, Bao, Behl, Benhaim, Bilenko, Bjorck, Bubeck, Cai, Cai, Chaudhary, Chen, Chen, Chen, Chen, Chen, Cheng, Chopra, Dai, Dixon, Eldan, Fragoso, Gao, Gao, Gao, Garg, Giorno, Goswami, Gunasekar, Haider, Hao, Hewett, Hu, Huynh, Iter, Jacobs, Javaheripi, Jin, Karampatziakis, Kauffmann, Khademi, Kim, Kim, Kurilenko, Lee, Lee, Li, Li, Liang, Liden, Lin, Lin, Liu, Liu, Liu, Liu, Liu, Luo, Madan, Mahmoudzadeh, Majercak, Mazzola, Mendes, Mitra, Modi, Nguyen, Norick, Patra, Perez-Becker, Portet, Pryzant, Qin, Radmilac, Ren, de~Rosa, Rosset, Roy, Ruwase, Saarikivi, Saied, Salim, Santacroce, Shah, Shang, Sharma, Shen, Shukla, Song, Tanaka, Tupini, Vaddamanu, Wang, Wang, Wang, Wang, Wang, Wang, Ward, Wen, Witte, Wu, Wu, Wyatt, Xiao, Xu, Xu, Xu, Xue, Yadav, Yang, Yang, Yang, Yang, Yu, Yuan, Zhang, Zhang, Zhang, Zhang, Zhang, Zhang, Zhang, and Zhou]{abdin2024phi3technicalreporthighly}
Marah Abdin, Jyoti Aneja, Hany Awadalla, Ahmed Awadallah, Ammar~Ahmad Awan, Nguyen Bach, Amit Bahree, Arash Bakhtiari, Jianmin Bao, Harkirat Behl, Alon Benhaim, Misha Bilenko, Johan Bjorck, Sébastien Bubeck, Martin Cai, Qin Cai, Vishrav Chaudhary, Dong Chen, Dongdong Chen, Weizhu Chen, Yen-Chun Chen, Yi-Ling Chen, Hao Cheng, Parul Chopra, Xiyang Dai, Matthew Dixon, Ronen Eldan, Victor Fragoso, Jianfeng Gao, Mei Gao, Min Gao, Amit Garg, Allie~Del Giorno, Abhishek Goswami, Suriya Gunasekar, Emman Haider, Junheng Hao, Russell~J. Hewett, Wenxiang Hu, Jamie Huynh, Dan Iter, Sam~Ade Jacobs, Mojan Javaheripi, Xin Jin, Nikos Karampatziakis, Piero Kauffmann, Mahoud Khademi, Dongwoo Kim, Young~Jin Kim, Lev Kurilenko, James~R. Lee, Yin~Tat Lee, Yuanzhi Li, Yunsheng Li, Chen Liang, Lars Liden, Xihui Lin, Zeqi Lin, Ce~Liu, Liyuan Liu, Mengchen Liu, Weishung Liu, Xiaodong Liu, Chong Luo, Piyush Madan, Ali Mahmoudzadeh, David Majercak, Matt Mazzola, Caio César~Teodoro Mendes, Arindam Mitra, Hardik Modi, Anh Nguyen, Brandon Norick, Barun Patra, Daniel Perez-Becker, Thomas Portet, Reid Pryzant, Heyang Qin, Marko Radmilac, Liliang Ren, Gustavo de~Rosa, Corby Rosset, Sambudha Roy, Olatunji Ruwase, Olli Saarikivi, Amin Saied, Adil Salim, Michael Santacroce, Shital Shah, Ning Shang, Hiteshi Sharma, Yelong Shen, Swadheen Shukla, Xia Song, Masahiro Tanaka, Andrea Tupini, Praneetha Vaddamanu, Chunyu Wang, Guanhua Wang, Lijuan Wang, Shuohang Wang, Xin Wang, Yu~Wang, Rachel Ward, Wen Wen, Philipp Witte, Haiping Wu, Xiaoxia Wu, Michael Wyatt, Bin Xiao, Can Xu, Jiahang Xu, Weijian Xu, Jilong Xue, Sonali Yadav, Fan Yang, Jianwei Yang, Yifan Yang, Ziyi Yang, Donghan Yu, Lu~Yuan, Chenruidong Zhang, Cyril Zhang, Jianwen Zhang, Li~Lyna Zhang, Yi~Zhang, Yue Zhang, Yunan Zhang, and Xiren Zhou.
\newblock Phi-3 technical report: A highly capable language model locally on your phone, 2024.
\newblock URL \url{https://arxiv.org/abs/2404.14219}.

\bibitem[Agarwal et~al.(2024)Agarwal, Singh, Zhang, Bohnet, Rosias, Chan, Zhang, Anand, Abbas, Nova, Co-Reyes, Chu, Behbahani, Faust, and Larochelle]{agarwal2024manyshotincontextlearning}
Rishabh Agarwal, Avi Singh, Lei~M. Zhang, Bernd Bohnet, Luis Rosias, Stephanie Chan, Biao Zhang, Ankesh Anand, Zaheer Abbas, Azade Nova, John~D. Co-Reyes, Eric Chu, Feryal Behbahani, Aleksandra Faust, and Hugo Larochelle.
\newblock Many-shot in-context learning, 2024.
\newblock URL \url{https://arxiv.org/abs/2404.11018}.

\bibitem[AI et~al.(2024)AI, :, Young, Chen, Li, Huang, Zhang, Zhang, Li, Zhu, Chen, Chang, Yu, Liu, Liu, Yue, Yang, Yang, Yu, Xie, Huang, Hu, Ren, Niu, Nie, Xu, Liu, Wang, Cai, Gu, Liu, and Dai]{ai2024yi}
01. AI, :, Alex Young, Bei Chen, Chao Li, Chengen Huang, Ge~Zhang, Guanwei Zhang, Heng Li, Jiangcheng Zhu, Jianqun Chen, Jing Chang, Kaidong Yu, Peng Liu, Qiang Liu, Shawn Yue, Senbin Yang, Shiming Yang, Tao Yu, Wen Xie, Wenhao Huang, Xiaohui Hu, Xiaoyi Ren, Xinyao Niu, Pengcheng Nie, Yuchi Xu, Yudong Liu, Yue Wang, Yuxuan Cai, Zhenyu Gu, Zhiyuan Liu, and Zonghong Dai.
\newblock Yi: Open foundation models by 01.ai, 2024.

\bibitem[An et~al.(2024)An, Gong, Zhong, Zhao, Li, Zhang, Kong, and Qiu]{an-etal-2024-l}
Chenxin An, Shansan Gong, Ming Zhong, Xingjian Zhao, Mukai Li, Jun Zhang, Lingpeng Kong, and Xipeng Qiu.
\newblock {L}-eval: Instituting standardized evaluation for long context language models.
\newblock In Lun-Wei Ku, Andre Martins, and Vivek Srikumar (eds.), \emph{Proceedings of the 62nd Annual Meeting of the Association for Computational Linguistics (Volume 1: Long Papers)}, pp.\  14388--14411, Bangkok, Thailand, August 2024. Association for Computational Linguistics.
\newblock URL \url{https://aclanthology.org/2024.acl-long.776}.

\bibitem[Anil et~al.(2024)Anil, Durmus, Sharma, Benton, Kundu, Batson, Rimsky, Tong, Mu, Ford, et~al.]{anil2024many}
Cem Anil, Esin Durmus, Mrinank Sharma, Joe Benton, Sandipan Kundu, Joshua Batson, Nina Rimsky, Meg Tong, Jesse Mu, Daniel Ford, et~al.
\newblock Many-shot jailbreaking.
\newblock \emph{Anthropic, April}, 2024.

\bibitem[Arora et~al.(2023)Arora, Eyuboglu, Timalsina, Johnson, Poli, Zou, Rudra, and Ré]{arora2023zoology}
Simran Arora, Sabri Eyuboglu, Aman Timalsina, Isys Johnson, Michael Poli, James Zou, Atri Rudra, and Christopher Ré.
\newblock Zoology: Measuring and improving recall in efficient language models, 2023.

\bibitem[Asai et~al.(2024{\natexlab{a}})Asai, Wu, Wang, Sil, and Hajishirzi]{asai2024selfrag}
Akari Asai, Zeqiu Wu, Yizhong Wang, Avirup Sil, and Hannaneh Hajishirzi.
\newblock Self-{RAG}: Learning to retrieve, generate, and critique through self-reflection.
\newblock In \emph{The Twelfth International Conference on Learning Representations}, 2024{\natexlab{a}}.
\newblock URL \url{https://openreview.net/forum?id=hSyW5go0v8}.

\bibitem[Asai et~al.(2024{\natexlab{b}})Asai, Zhong, Chen, Koh, Zettlemoyer, Hajishirzi, and tau Yih]{asai2024reliableadaptableattributablelanguage}
Akari Asai, Zexuan Zhong, Danqi Chen, Pang~Wei Koh, Luke Zettlemoyer, Hannaneh Hajishirzi, and Wen tau Yih.
\newblock Reliable, adaptable, and attributable language models with retrieval, 2024{\natexlab{b}}.
\newblock URL \url{https://arxiv.org/abs/2403.03187}.

\bibitem[Bai et~al.(2024)Bai, Lv, Zhang, Lyu, Tang, Huang, Du, Liu, Zeng, Hou, Dong, Tang, and Li]{bai-etal-2024-longbench}
Yushi Bai, Xin Lv, Jiajie Zhang, Hongchang Lyu, Jiankai Tang, Zhidian Huang, Zhengxiao Du, Xiao Liu, Aohan Zeng, Lei Hou, Yuxiao Dong, Jie Tang, and Juanzi Li.
\newblock {L}ong{B}ench: A bilingual, multitask benchmark for long context understanding.
\newblock In Lun-Wei Ku, Andre Martins, and Vivek Srikumar (eds.), \emph{Proceedings of the 62nd Annual Meeting of the Association for Computational Linguistics (Volume 1: Long Papers)}, pp.\  3119--3137, Bangkok, Thailand, August 2024. Association for Computational Linguistics.
\newblock URL \url{https://aclanthology.org/2024.acl-long.172}.

\bibitem[Bajaj et~al.(2018)Bajaj, Campos, Craswell, Deng, Gao, Liu, Majumder, McNamara, Mitra, Nguyen, Rosenberg, Song, Stoica, Tiwary, and Wang]{bajaj2018ms}
Payal Bajaj, Daniel Campos, Nick Craswell, Li~Deng, Jianfeng Gao, Xiaodong Liu, Rangan Majumder, Andrew McNamara, Bhaskar Mitra, Tri Nguyen, Mir Rosenberg, Xia Song, Alina Stoica, Saurabh Tiwary, and Tong Wang.
\newblock Ms marco: A human generated machine reading comprehension dataset, 2018.
\newblock URL \url{https://arxiv.org/abs/1611.09268}.

\bibitem[Beltagy et~al.(2020)Beltagy, Peters, and Cohan]{Beltagy2020Longformer}
Iz~Beltagy, Matthew~E. Peters, and Arman Cohan.
\newblock Longformer: The long-document {Transformer}.
\newblock \emph{arXiv:2004.05150}, 2020.
\newblock URL \url{https://arxiv.org/abs/2004.05150}.

\bibitem[Bertsch et~al.(2023)Bertsch, Alon, Neubig, and Gormley]{bertsch2023unlimiformer}
Amanda Bertsch, Uri Alon, Graham Neubig, and Matthew~R. Gormley.
\newblock Unlimiformer: Long-range transformers with unlimited length input.
\newblock In \emph{Advances in Neural Information Processing Systems (NeurIPS)}, 2023.
\newblock URL \url{https://openreview.net/forum?id=lJWUJWLCJo}.

\bibitem[Bertsch et~al.(2024)Bertsch, Ivgi, Alon, Berant, Gormley, and Neubig]{bertsch2024incontext}
Amanda Bertsch, Maor Ivgi, Uri Alon, Jonathan Berant, Matthew~R. Gormley, and Graham Neubig.
\newblock In-context learning with long-context models: An in-depth exploration.
\newblock In \emph{First Workshop on Long-Context Foundation Models @ ICML 2024}, 2024.
\newblock URL \url{https://openreview.net/forum?id=4KAmc7vUbq}.

\bibitem[Bohnet et~al.(2022)Bohnet, Tran, Verga, Aharoni, Andor, Soares, Eisenstein, Ganchev, Herzig, Hui, et~al.]{bohnet2022attributed}
Bernd Bohnet, Vinh~Q Tran, Pat Verga, Roee Aharoni, Daniel Andor, Livio~Baldini Soares, Jacob Eisenstein, Kuzman Ganchev, Jonathan Herzig, Kai Hui, et~al.
\newblock Attributed question answering: Evaluation and modeling for attributed large language models.
\newblock \emph{arXiv preprint arXiv:2212.08037}, 2022.
\newblock URL \url{https://arxiv.org/pdf/2212.08037.pdf}.

\bibitem[Brown et~al.(2020)Brown, Mann, Ryder, Subbiah, Kaplan, Dhariwal, Neelakantan, Shyam, Sastry, Askell, et~al.]{brown2020language}
Tom~B Brown, Benjamin Mann, Nick Ryder, Melanie Subbiah, Jared Kaplan, Prafulla Dhariwal, Arvind Neelakantan, Pranav Shyam, Girish Sastry, Amanda Askell, et~al.
\newblock Language models are few-shot learners.
\newblock In \emph{Advances in Neural Information Processing Systems (NeurIPS)}, 2020.
\newblock URL \url{https://papers.nips.cc/paper/2020/hash/1457c0d6bfcb4967418bfb8ac142f64a-Abstract.html}.

\bibitem[Casanueva et~al.(2020)Casanueva, Tem{\v{c}}inas, Gerz, Henderson, and Vuli{\'c}]{casanueva-etal-2020-efficient}
I{\~n}igo Casanueva, Tadas Tem{\v{c}}inas, Daniela Gerz, Matthew Henderson, and Ivan Vuli{\'c}.
\newblock Efficient intent detection with dual sentence encoders.
\newblock In Tsung-Hsien Wen, Asli Celikyilmaz, Zhou Yu, Alexandros Papangelis, Mihail Eric, Anuj Kumar, I{\~n}igo Casanueva, and Rushin Shah (eds.), \emph{Proceedings of the 2nd Workshop on Natural Language Processing for Conversational AI}, pp.\  38--45, Online, July 2020. Association for Computational Linguistics.
\newblock \doi{10.18653/v1/2020.nlp4convai-1.5}.
\newblock URL \url{https://aclanthology.org/2020.nlp4convai-1.5}.

\bibitem[Chang et~al.(2024)Chang, Lo, Goyal, and Iyyer]{chang2024booookscore}
Yapei Chang, Kyle Lo, Tanya Goyal, and Mohit Iyyer.
\newblock Booookscore: A systematic exploration of book-length summarization in the era of {LLM}s.
\newblock In \emph{The Twelfth International Conference on Learning Representations}, 2024.
\newblock URL \url{https://openreview.net/forum?id=7Ttk3RzDeu}.

\bibitem[Chen et~al.(2017)Chen, Fisch, Weston, and Bordes]{chen-etal-2017-reading}
Danqi Chen, Adam Fisch, Jason Weston, and Antoine Bordes.
\newblock Reading {W}ikipedia to answer open-domain questions.
\newblock In Regina Barzilay and Min-Yen Kan (eds.), \emph{Proceedings of the 55th Annual Meeting of the Association for Computational Linguistics (Volume 1: Long Papers)}, pp.\  1870--1879, Vancouver, Canada, July 2017. Association for Computational Linguistics.
\newblock \doi{10.18653/v1/P17-1171}.
\newblock URL \url{https://aclanthology.org/P17-1171}.

\bibitem[Chen et~al.(2023)Chen, Wong, Chen, and Tian]{chen2023extending}
Shouyuan Chen, Sherman Wong, Liangjian Chen, and Yuandong Tian.
\newblock Extending context window of large language models via positional interpolation, 2023.

\bibitem[Craswell et~al.(2020)Craswell, Mitra, Yilmaz, Campos, and Voorhees]{craswell2020overviewtrec2019deep}
Nick Craswell, Bhaskar Mitra, Emine Yilmaz, Daniel Campos, and Ellen~M. Voorhees.
\newblock Overview of the trec 2019 deep learning track, 2020.
\newblock URL \url{https://arxiv.org/abs/2003.07820}.

\bibitem[Dao(2023)]{dao2023flashattention2fasterattentionbetter}
Tri Dao.
\newblock Flashattention-2: Faster attention with better parallelism and work partitioning, 2023.
\newblock URL \url{https://arxiv.org/abs/2307.08691}.

\bibitem[Dao \& Gu(2024)Dao and Gu]{dao2024transformersssmsgeneralizedmodels}
Tri Dao and Albert Gu.
\newblock Transformers are ssms: Generalized models and efficient algorithms through structured state space duality, 2024.
\newblock URL \url{https://arxiv.org/abs/2405.21060}.

\bibitem[Deutsch et~al.(2022)Deutsch, Dror, and Roth]{deutsch-etal-2022-examining}
Daniel Deutsch, Rotem Dror, and Dan Roth.
\newblock Re-examining system-level correlations of automatic summarization evaluation metrics.
\newblock In Marine Carpuat, Marie-Catherine de~Marneffe, and Ivan~Vladimir Meza~Ruiz (eds.), \emph{Proceedings of the 2022 Conference of the North American Chapter of the Association for Computational Linguistics: Human Language Technologies}, pp.\  6038--6052, Seattle, United States, July 2022. Association for Computational Linguistics.
\newblock \doi{10.18653/v1/2022.naacl-main.442}.
\newblock URL \url{https://aclanthology.org/2022.naacl-main.442}.

\bibitem[Ding et~al.(2024)Ding, Zhang, Zhang, Xu, Shang, Xu, Yang, and Yang]{ding2024longropeextendingllmcontext}
Yiran Ding, Li~Lyna Zhang, Chengruidong Zhang, Yuanyuan Xu, Ning Shang, Jiahang Xu, Fan Yang, and Mao Yang.
\newblock Longrope: Extending llm context window beyond 2 million tokens, 2024.
\newblock URL \url{https://arxiv.org/abs/2402.13753}.

\bibitem[Dong et~al.(2024)Dong, Tang, Li, Zhao, and Wen]{dong-etal-2024-bamboo}
Zican Dong, Tianyi Tang, Junyi Li, Wayne~Xin Zhao, and Ji-Rong Wen.
\newblock {BAMBOO}: A comprehensive benchmark for evaluating long text modeling capacities of large language models.
\newblock In Nicoletta Calzolari, Min-Yen Kan, Veronique Hoste, Alessandro Lenci, Sakriani Sakti, and Nianwen Xue (eds.), \emph{Proceedings of the 2024 Joint International Conference on Computational Linguistics, Language Resources and Evaluation (LREC-COLING 2024)}, pp.\  2086--2099, Torino, Italia, May 2024. ELRA and ICCL.
\newblock URL \url{https://aclanthology.org/2024.lrec-main.188}.

\bibitem[Dubey et~al.(2024)Dubey, Jauhri, Pandey, Kadian, Al-Dahle, Letman, Mathur, Schelten, Yang, Fan, Goyal, Hartshorn, Yang, Mitra, Sravankumar, Korenev, Hinsvark, Rao, Zhang, Rodriguez, Gregerson, Spataru, Roziere, Biron, Tang, Chern, Caucheteux, Nayak, Bi, Marra, McConnell, Keller, Touret, Wu, Wong, Ferrer, Nikolaidis, Allonsius, Song, Pintz, Livshits, Esiobu, Choudhary, Mahajan, Garcia-Olano, Perino, Hupkes, Lakomkin, AlBadawy, Lobanova, Dinan, Smith, Radenovic, Zhang, Synnaeve, Lee, Anderson, Nail, Mialon, Pang, Cucurell, Nguyen, Korevaar, Xu, Touvron, Zarov, Ibarra, Kloumann, Misra, Evtimov, Copet, Lee, Geffert, Vranes, Park, Mahadeokar, Shah, van~der Linde, Billock, Hong, Lee, Fu, Chi, Huang, Liu, Wang, Yu, Bitton, Spisak, Park, Rocca, Johnstun, Saxe, Jia, Alwala, Upasani, Plawiak, Li, Heafield, Stone, El-Arini, Iyer, Malik, Chiu, Bhalla, Rantala-Yeary, van~der Maaten, Chen, Tan, Jenkins, Martin, Madaan, Malo, Blecher, Landzaat, de~Oliveira, Muzzi, Pasupuleti, Singh, Paluri, Kardas, Oldham, Rita, Pavlova, Kambadur, Lewis, Si, Singh, Hassan, Goyal, Torabi, Bashlykov, Bogoychev, Chatterji, Duchenne, Çelebi, Alrassy, Zhang, Li, Vasic, Weng, Bhargava, Dubal, Krishnan, Koura, Xu, He, Dong, Srinivasan, Ganapathy, Calderer, Cabral, Stojnic, Raileanu, Girdhar, Patel, Sauvestre, Polidoro, Sumbaly, Taylor, Silva, Hou, Wang, Hosseini, Chennabasappa, Singh, Bell, Kim, Edunov, Nie, Narang, Raparthy, Shen, Wan, Bhosale, Zhang, Vandenhende, Batra, Whitman, Sootla, Collot, Gururangan, Borodinsky, Herman, Fowler, Sheasha, Georgiou, Scialom, Speckbacher, Mihaylov, Xiao, Karn, Goswami, Gupta, Ramanathan, Kerkez, Gonguet, Do, Vogeti, Petrovic, Chu, Xiong, Fu, Meers, Martinet, Wang, Tan, Xie, Jia, Wang, Goldschlag, Gaur, Babaei, Wen, Song, Zhang, Li, Mao, Coudert, Yan, Chen, Papakipos, Singh, Grattafiori, Jain, Kelsey, Shajnfeld, Gangidi, Victoria, Goldstand, Menon, Sharma, Boesenberg, Vaughan, Baevski, Feinstein, Kallet, Sangani, Yunus, Lupu, Alvarado, Caples, Gu, Ho, Poulton, Ryan, Ramchandani, Franco, Saraf, Chowdhury, Gabriel, Bharambe, Eisenman, Yazdan, James, Maurer, Leonhardi, Huang, Loyd, Paola, Paranjape, Liu, Wu, Ni, Hancock, Wasti, Spence, Stojkovic, Gamido, Montalvo, Parker, Burton, Mejia, Wang, Kim, Zhou, Hu, Chu, Cai, Tindal, Feichtenhofer, Civin, Beaty, Kreymer, Li, Wyatt, Adkins, Xu, Testuggine, David, Parikh, Liskovich, Foss, Wang, Le, Holland, Dowling, Jamil, Montgomery, Presani, Hahn, Wood, Brinkman, Arcaute, Dunbar, Smothers, Sun, Kreuk, Tian, Ozgenel, Caggioni, Guzmán, Kanayet, Seide, Florez, Schwarz, Badeer, Swee, Halpern, Thattai, Herman, Sizov, Guangyi, Zhang, Lakshminarayanan, Shojanazeri, Zou, Wang, Zha, Habeeb, Rudolph, Suk, Aspegren, Goldman, Damlaj, Molybog, Tufanov, Veliche, Gat, Weissman, Geboski, Kohli, Asher, Gaya, Marcus, Tang, Chan, Zhen, Reizenstein, Teboul, Zhong, Jin, Yang, Cummings, Carvill, Shepard, McPhie, Torres, Ginsburg, Wang, Wu, U, Saxena, Prasad, Khandelwal, Zand, Matosich, Veeraraghavan, Michelena, Li, Huang, Chawla, Lakhotia, Huang, Chen, Garg, A, Silva, Bell, Zhang, Guo, Yu, Moshkovich, Wehrstedt, Khabsa, Avalani, Bhatt, Tsimpoukelli, Mankus, Hasson, Lennie, Reso, Groshev, Naumov, Lathi, Keneally, Seltzer, Valko, Restrepo, Patel, Vyatskov, Samvelyan, Clark, Macey, Wang, Hermoso, Metanat, Rastegari, Bansal, Santhanam, Parks, White, Bawa, Singhal, Egebo, Usunier, Laptev, Dong, Zhang, Cheng, Chernoguz, Hart, Salpekar, Kalinli, Kent, Parekh, Saab, Balaji, Rittner, Bontrager, Roux, Dollar, Zvyagina, Ratanchandani, Yuvraj, Liang, Alao, Rodriguez, Ayub, Murthy, Nayani, Mitra, Li, Hogan, Battey, Wang, Maheswari, Howes, Rinott, Bondu, Datta, Chugh, Hunt, Dhillon, Sidorov, Pan, Verma, Yamamoto, Ramaswamy, Lindsay, Lindsay, Feng, Lin, Zha, Shankar, Zhang, Zhang, Wang, Agarwal, Sajuyigbe, Chintala, Max, Chen, Kehoe, Satterfield, Govindaprasad, Gupta, Cho, Virk, Subramanian, Choudhury, Goldman, Remez, Glaser, Best, Kohler, Robinson, Li, Zhang, Matthews, Chou, Shaked, Vontimitta, Ajayi, Montanez, Mohan, Kumar, Mangla, Albiero, Ionescu, Poenaru, Mihailescu, Ivanov, Li, Wang, Jiang, Bouaziz, Constable, Tang, Wang, Wu, Wang, Xia, Wu, Gao, Chen, Hu, Jia, Qi, Li, Zhang, Zhang, Adi, Nam, Yu, Wang, Hao, Qian, He, Rait, DeVito, Rosnbrick, Wen, Yang, and Zhao]{dubey2024llama3herdmodels}
Abhimanyu Dubey, Abhinav Jauhri, Abhinav Pandey, Abhishek Kadian, Ahmad Al-Dahle, Aiesha Letman, Akhil Mathur, Alan Schelten, Amy Yang, Angela Fan, Anirudh Goyal, Anthony Hartshorn, Aobo Yang, Archi Mitra, Archie Sravankumar, Artem Korenev, Arthur Hinsvark, Arun Rao, Aston Zhang, Aurelien Rodriguez, Austen Gregerson, Ava Spataru, Baptiste Roziere, Bethany Biron, Binh Tang, Bobbie Chern, Charlotte Caucheteux, Chaya Nayak, Chloe Bi, Chris Marra, Chris McConnell, Christian Keller, Christophe Touret, Chunyang Wu, Corinne Wong, Cristian~Canton Ferrer, Cyrus Nikolaidis, Damien Allonsius, Daniel Song, Danielle Pintz, Danny Livshits, David Esiobu, Dhruv Choudhary, Dhruv Mahajan, Diego Garcia-Olano, Diego Perino, Dieuwke Hupkes, Egor Lakomkin, Ehab AlBadawy, Elina Lobanova, Emily Dinan, Eric~Michael Smith, Filip Radenovic, Frank Zhang, Gabriel Synnaeve, Gabrielle Lee, Georgia~Lewis Anderson, Graeme Nail, Gregoire Mialon, Guan Pang, Guillem Cucurell, Hailey Nguyen, Hannah Korevaar, Hu~Xu, Hugo Touvron, Iliyan Zarov, Imanol~Arrieta Ibarra, Isabel Kloumann, Ishan Misra, Ivan Evtimov, Jade Copet, Jaewon Lee, Jan Geffert, Jana Vranes, Jason Park, Jay Mahadeokar, Jeet Shah, Jelmer van~der Linde, Jennifer Billock, Jenny Hong, Jenya Lee, Jeremy Fu, Jianfeng Chi, Jianyu Huang, Jiawen Liu, Jie Wang, Jiecao Yu, Joanna Bitton, Joe Spisak, Jongsoo Park, Joseph Rocca, Joshua Johnstun, Joshua Saxe, Junteng Jia, Kalyan~Vasuden Alwala, Kartikeya Upasani, Kate Plawiak, Ke~Li, Kenneth Heafield, Kevin Stone, Khalid El-Arini, Krithika Iyer, Kshitiz Malik, Kuenley Chiu, Kunal Bhalla, Lauren Rantala-Yeary, Laurens van~der Maaten, Lawrence Chen, Liang Tan, Liz Jenkins, Louis Martin, Lovish Madaan, Lubo Malo, Lukas Blecher, Lukas Landzaat, Luke de~Oliveira, Madeline Muzzi, Mahesh Pasupuleti, Mannat Singh, Manohar Paluri, Marcin Kardas, Mathew Oldham, Mathieu Rita, Maya Pavlova, Melanie Kambadur, Mike Lewis, Min Si, Mitesh~Kumar Singh, Mona Hassan, Naman Goyal, Narjes Torabi, Nikolay Bashlykov, Nikolay Bogoychev, Niladri Chatterji, Olivier Duchenne, Onur Çelebi, Patrick Alrassy, Pengchuan Zhang, Pengwei Li, Petar Vasic, Peter Weng, Prajjwal Bhargava, Pratik Dubal, Praveen Krishnan, Punit~Singh Koura, Puxin Xu, Qing He, Qingxiao Dong, Ragavan Srinivasan, Raj Ganapathy, Ramon Calderer, Ricardo~Silveira Cabral, Robert Stojnic, Roberta Raileanu, Rohit Girdhar, Rohit Patel, Romain Sauvestre, Ronnie Polidoro, Roshan Sumbaly, Ross Taylor, Ruan Silva, Rui Hou, Rui Wang, Saghar Hosseini, Sahana Chennabasappa, Sanjay Singh, Sean Bell, Seohyun~Sonia Kim, Sergey Edunov, Shaoliang Nie, Sharan Narang, Sharath Raparthy, Sheng Shen, Shengye Wan, Shruti Bhosale, Shun Zhang, Simon Vandenhende, Soumya Batra, Spencer Whitman, Sten Sootla, Stephane Collot, Suchin Gururangan, Sydney Borodinsky, Tamar Herman, Tara Fowler, Tarek Sheasha, Thomas Georgiou, Thomas Scialom, Tobias Speckbacher, Todor Mihaylov, Tong Xiao, Ujjwal Karn, Vedanuj Goswami, Vibhor Gupta, Vignesh Ramanathan, Viktor Kerkez, Vincent Gonguet, Virginie Do, Vish Vogeti, Vladan Petrovic, Weiwei Chu, Wenhan Xiong, Wenyin Fu, Whitney Meers, Xavier Martinet, Xiaodong Wang, Xiaoqing~Ellen Tan, Xinfeng Xie, Xuchao Jia, Xuewei Wang, Yaelle Goldschlag, Yashesh Gaur, Yasmine Babaei, Yi~Wen, Yiwen Song, Yuchen Zhang, Yue Li, Yuning Mao, Zacharie~Delpierre Coudert, Zheng Yan, Zhengxing Chen, Zoe Papakipos, Aaditya Singh, Aaron Grattafiori, Abha Jain, Adam Kelsey, Adam Shajnfeld, Adithya Gangidi, Adolfo Victoria, Ahuva Goldstand, Ajay Menon, Ajay Sharma, Alex Boesenberg, Alex Vaughan, Alexei Baevski, Allie Feinstein, Amanda Kallet, Amit Sangani, Anam Yunus, Andrei Lupu, Andres Alvarado, Andrew Caples, Andrew Gu, Andrew Ho, Andrew Poulton, Andrew Ryan, Ankit Ramchandani, Annie Franco, Aparajita Saraf, Arkabandhu Chowdhury, Ashley Gabriel, Ashwin Bharambe, Assaf Eisenman, Azadeh Yazdan, Beau James, Ben Maurer, Benjamin Leonhardi, Bernie Huang, Beth Loyd, Beto~De Paola, Bhargavi Paranjape, Bing Liu, Bo~Wu, Boyu Ni, Braden Hancock, Bram Wasti, Brandon Spence, Brani Stojkovic, Brian Gamido, Britt Montalvo, Carl Parker, Carly Burton, Catalina Mejia, Changhan Wang, Changkyu Kim, Chao Zhou, Chester Hu, Ching-Hsiang Chu, Chris Cai, Chris Tindal, Christoph Feichtenhofer, Damon Civin, Dana Beaty, Daniel Kreymer, Daniel Li, Danny Wyatt, David Adkins, David Xu, Davide Testuggine, Delia David, Devi Parikh, Diana Liskovich, Didem Foss, Dingkang Wang, Duc Le, Dustin Holland, Edward Dowling, Eissa Jamil, Elaine Montgomery, Eleonora Presani, Emily Hahn, Emily Wood, Erik Brinkman, Esteban Arcaute, Evan Dunbar, Evan Smothers, Fei Sun, Felix Kreuk, Feng Tian, Firat Ozgenel, Francesco Caggioni, Francisco Guzmán, Frank Kanayet, Frank Seide, Gabriela~Medina Florez, Gabriella Schwarz, Gada Badeer, Georgia Swee, Gil Halpern, Govind Thattai, Grant Herman, Grigory Sizov, Guangyi, Zhang, Guna Lakshminarayanan, Hamid Shojanazeri, Han Zou, Hannah Wang, Hanwen Zha, Haroun Habeeb, Harrison Rudolph, Helen Suk, Henry Aspegren, Hunter Goldman, Ibrahim Damlaj, Igor Molybog, Igor Tufanov, Irina-Elena Veliche, Itai Gat, Jake Weissman, James Geboski, James Kohli, Japhet Asher, Jean-Baptiste Gaya, Jeff Marcus, Jeff Tang, Jennifer Chan, Jenny Zhen, Jeremy Reizenstein, Jeremy Teboul, Jessica Zhong, Jian Jin, Jingyi Yang, Joe Cummings, Jon Carvill, Jon Shepard, Jonathan McPhie, Jonathan Torres, Josh Ginsburg, Junjie Wang, Kai Wu, Kam~Hou U, Karan Saxena, Karthik Prasad, Kartikay Khandelwal, Katayoun Zand, Kathy Matosich, Kaushik Veeraraghavan, Kelly Michelena, Keqian Li, Kun Huang, Kunal Chawla, Kushal Lakhotia, Kyle Huang, Lailin Chen, Lakshya Garg, Lavender A, Leandro Silva, Lee Bell, Lei Zhang, Liangpeng Guo, Licheng Yu, Liron Moshkovich, Luca Wehrstedt, Madian Khabsa, Manav Avalani, Manish Bhatt, Maria Tsimpoukelli, Martynas Mankus, Matan Hasson, Matthew Lennie, Matthias Reso, Maxim Groshev, Maxim Naumov, Maya Lathi, Meghan Keneally, Michael~L. Seltzer, Michal Valko, Michelle Restrepo, Mihir Patel, Mik Vyatskov, Mikayel Samvelyan, Mike Clark, Mike Macey, Mike Wang, Miquel~Jubert Hermoso, Mo~Metanat, Mohammad Rastegari, Munish Bansal, Nandhini Santhanam, Natascha Parks, Natasha White, Navyata Bawa, Nayan Singhal, Nick Egebo, Nicolas Usunier, Nikolay~Pavlovich Laptev, Ning Dong, Ning Zhang, Norman Cheng, Oleg Chernoguz, Olivia Hart, Omkar Salpekar, Ozlem Kalinli, Parkin Kent, Parth Parekh, Paul Saab, Pavan Balaji, Pedro Rittner, Philip Bontrager, Pierre Roux, Piotr Dollar, Polina Zvyagina, Prashant Ratanchandani, Pritish Yuvraj, Qian Liang, Rachad Alao, Rachel Rodriguez, Rafi Ayub, Raghotham Murthy, Raghu Nayani, Rahul Mitra, Raymond Li, Rebekkah Hogan, Robin Battey, Rocky Wang, Rohan Maheswari, Russ Howes, Ruty Rinott, Sai~Jayesh Bondu, Samyak Datta, Sara Chugh, Sara Hunt, Sargun Dhillon, Sasha Sidorov, Satadru Pan, Saurabh Verma, Seiji Yamamoto, Sharadh Ramaswamy, Shaun Lindsay, Shaun Lindsay, Sheng Feng, Shenghao Lin, Shengxin~Cindy Zha, Shiva Shankar, Shuqiang Zhang, Shuqiang Zhang, Sinong Wang, Sneha Agarwal, Soji Sajuyigbe, Soumith Chintala, Stephanie Max, Stephen Chen, Steve Kehoe, Steve Satterfield, Sudarshan Govindaprasad, Sumit Gupta, Sungmin Cho, Sunny Virk, Suraj Subramanian, Sy~Choudhury, Sydney Goldman, Tal Remez, Tamar Glaser, Tamara Best, Thilo Kohler, Thomas Robinson, Tianhe Li, Tianjun Zhang, Tim Matthews, Timothy Chou, Tzook Shaked, Varun Vontimitta, Victoria Ajayi, Victoria Montanez, Vijai Mohan, Vinay~Satish Kumar, Vishal Mangla, Vítor Albiero, Vlad Ionescu, Vlad Poenaru, Vlad~Tiberiu Mihailescu, Vladimir Ivanov, Wei Li, Wenchen Wang, Wenwen Jiang, Wes Bouaziz, Will Constable, Xiaocheng Tang, Xiaofang Wang, Xiaojian Wu, Xiaolan Wang, Xide Xia, Xilun Wu, Xinbo Gao, Yanjun Chen, Ye~Hu, Ye~Jia, Ye~Qi, Yenda Li, Yilin Zhang, Ying Zhang, Yossi Adi, Youngjin Nam, Yu, Wang, Yuchen Hao, Yundi Qian, Yuzi He, Zach Rait, Zachary DeVito, Zef Rosnbrick, Zhaoduo Wen, Zhenyu Yang, and Zhiwei Zhao.
\newblock The llama 3 herd of models, 2024.
\newblock URL \url{https://arxiv.org/abs/2407.21783}.

\bibitem[Fu et~al.(2024)Fu, Panda, Niu, Yue, Hajishirzi, Kim, and Peng]{fu2024data}
Yao Fu, Rameswar Panda, Xinyao Niu, Xiang Yue, Hannaneh Hajishirzi, Yoon Kim, and Hao Peng.
\newblock Data engineering for scaling language models to 128k context, 2024.

\bibitem[Gao et~al.(2023)Gao, Yen, Yu, and Chen]{gao2023alce}
Tianyu Gao, Howard Yen, Jiatong Yu, and Danqi Chen.
\newblock Enabling large language models to generate text with citations.
\newblock In Houda Bouamor, Juan Pino, and Kalika Bali (eds.), \emph{Proceedings of the 2023 Conference on Empirical Methods in Natural Language Processing}, pp.\  6465--6488, Singapore, December 2023. Association for Computational Linguistics.
\newblock \doi{10.18653/v1/2023.emnlp-main.398}.
\newblock URL \url{https://aclanthology.org/2023.emnlp-main.398}.

\bibitem[Gao et~al.(2024)Gao, Wettig, Yen, and Chen]{gao2024prolong}
Tianyu Gao, Alexander Wettig, Howard Yen, and Danqi Chen.
\newblock How to train long-context language models (effectively), 2024.
\newblock URL \url{https://arxiv.org/abs/2410.02660}.

\bibitem[Goldman et~al.(2024)Goldman, Jacovi, Slobodkin, Maimon, Dagan, and Tsarfaty]{goldman2024reallylongcontextneed}
Omer Goldman, Alon Jacovi, Aviv Slobodkin, Aviya Maimon, Ido Dagan, and Reut Tsarfaty.
\newblock Is it really long context if all you need is retrieval? towards genuinely difficult long context nlp, 2024.
\newblock URL \url{https://arxiv.org/abs/2407.00402}.

\bibitem[Goyal et~al.(2023)Goyal, Li, and Durrett]{goyal2023newssummarizationevaluationera}
Tanya Goyal, Junyi~Jessy Li, and Greg Durrett.
\newblock News {Summarization and Evaluation in the Era of GPT-3}, 2023.
\newblock URL \url{https://arxiv.org/abs/2209.12356}.

\bibitem[Gu \& Dao(2024)Gu and Dao]{gu2024mamba}
Albert Gu and Tri Dao.
\newblock Mamba: Linear-time sequence modeling with selective state spaces.
\newblock In \emph{First Conference on Language Modeling}, 2024.
\newblock URL \url{https://openreview.net/forum?id=tEYskw1VY2}.

\bibitem[Hsieh et~al.(2024)Hsieh, Sun, Kriman, Acharya, Rekesh, Jia, and Ginsburg]{hsieh2024ruler}
Cheng-Ping Hsieh, Simeng Sun, Samuel Kriman, Shantanu Acharya, Dima Rekesh, Fei Jia, and Boris Ginsburg.
\newblock {RULER}: What{\textquoteright}s the real context size of your long-context language models?
\newblock In \emph{First Conference on Language Modeling}, 2024.
\newblock URL \url{https://openreview.net/forum?id=kIoBbc76Sy}.

\bibitem[Jiang et~al.(2023)Jiang, Sablayrolles, Mensch, Bamford, Chaplot, de~las Casas, Bressand, Lengyel, Lample, Saulnier, Lavaud, Lachaux, Stock, Scao, Lavril, Wang, Lacroix, and Sayed]{jiang2023mistral}
Albert~Q. Jiang, Alexandre Sablayrolles, Arthur Mensch, Chris Bamford, Devendra~Singh Chaplot, Diego de~las Casas, Florian Bressand, Gianna Lengyel, Guillaume Lample, Lucile Saulnier, Lélio~Renard Lavaud, Marie-Anne Lachaux, Pierre Stock, Teven~Le Scao, Thibaut Lavril, Thomas Wang, Timothée Lacroix, and William~El Sayed.
\newblock Mistral 7b, 2023.

\bibitem[Joshi et~al.(2017)Joshi, Choi, Weld, and Zettlemoyer]{joshi2017triviaqa}
Mandar Joshi, Eunsol Choi, Daniel Weld, and Luke Zettlemoyer.
\newblock {T}rivia{QA}: A large scale distantly supervised challenge dataset for reading comprehension.
\newblock In Regina Barzilay and Min-Yen Kan (eds.), \emph{Proceedings of the 55th Annual Meeting of the Association for Computational Linguistics (Volume 1: Long Papers)}, pp.\  1601--1611, Vancouver, Canada, July 2017. Association for Computational Linguistics.
\newblock \doi{10.18653/v1/P17-1147}.
\newblock URL \url{https://aclanthology.org/P17-1147}.

\bibitem[Kamoi et~al.(2023)Kamoi, Goyal, Rodriguez, and Durrett]{kamoi2023wice}
Ryo Kamoi, Tanya Goyal, Juan~Diego Rodriguez, and Greg Durrett.
\newblock {WiCE: Real-World Entailment for Claims in Wikipedia}.
\newblock \emph{arXiv preprint arXiv:2303.01432}, 2023.
\newblock URL \url{https://arxiv.org/abs/2303.01432}.

\bibitem[Kamradt(2024)]{gkamradt_llmtest_needleinahaystack_2024}
Garrett Kamradt.
\newblock Needle in a haystack - pressure testing llms, 2024.
\newblock URL \url{https://github.com/gkamradt/LLMTest_NeedleInAHaystack}.

\bibitem[Karpinska et~al.(2024)Karpinska, Thai, Lo, Goyal, and Iyyer]{karpinska2024thousandpairsnovelchallenge}
Marzena Karpinska, Katherine Thai, Kyle Lo, Tanya Goyal, and Mohit Iyyer.
\newblock One thousand and one pairs: A "novel" challenge for long-context language models, 2024.
\newblock URL \url{https://arxiv.org/abs/2406.16264}.

\bibitem[Kim et~al.(2024)Kim, Chang, Karpinska, Garimella, Manjunatha, Lo, Goyal, and Iyyer]{kim2024fables}
Yekyung Kim, Yapei Chang, Marzena Karpinska, Aparna Garimella, Varun Manjunatha, Kyle Lo, Tanya Goyal, and Mohit Iyyer.
\newblock {FABLES}: Evaluating faithfulness and content selection in book-length summarization.
\newblock In \emph{First Conference on Language Modeling}, 2024.
\newblock URL \url{https://openreview.net/forum?id=YfHxQSoaWU}.

\bibitem[Ko{\v{c}}isk{\'y} et~al.(2018)Ko{\v{c}}isk{\'y}, Schwarz, Blunsom, Dyer, Hermann, Melis, and Grefenstette]{kocisky2018narrativeqa}
Tom{\'a}{\v{s}} Ko{\v{c}}isk{\'y}, Jonathan Schwarz, Phil Blunsom, Chris Dyer, Karl~Moritz Hermann, G{\'a}bor Melis, and Edward Grefenstette.
\newblock The {N}arrative{QA} reading comprehension challenge.
\newblock \emph{Transactions of the Association for Computational Linguistics}, 6:\penalty0 317--328, 2018.
\newblock \doi{10.1162/tacl_a_00023}.
\newblock URL \url{https://aclanthology.org/Q18-1023}.

\bibitem[Krishna et~al.(2023)Krishna, Bransom, Kuehl, Iyyer, Dasigi, Cohan, and Lo]{krishna-etal-2023-longeval}
Kalpesh Krishna, Erin Bransom, Bailey Kuehl, Mohit Iyyer, Pradeep Dasigi, Arman Cohan, and Kyle Lo.
\newblock {L}ong{E}val: Guidelines for human evaluation of faithfulness in long-form summarization.
\newblock In Andreas Vlachos and Isabelle Augenstein (eds.), \emph{Proceedings of the 17th Conference of the European Chapter of the Association for Computational Linguistics}, pp.\  1650--1669, Dubrovnik, Croatia, May 2023. Association for Computational Linguistics.
\newblock \doi{10.18653/v1/2023.eacl-main.121}.
\newblock URL \url{https://aclanthology.org/2023.eacl-main.121/}.

\bibitem[Kwan et~al.(2024)Kwan, Zeng, Wang, Sun, Li, Jiang, Shang, Liu, and Wong]{kwan-etal-2024-m4le}
Wai-Chung Kwan, Xingshan Zeng, Yufei Wang, Yusen Sun, Liangyou Li, Yuxin Jiang, Lifeng Shang, Qun Liu, and Kam-Fai Wong.
\newblock {M}4{LE}: A multi-ability multi-range multi-task multi-domain long-context evaluation benchmark for large language models.
\newblock In Lun-Wei Ku, Andre Martins, and Vivek Srikumar (eds.), \emph{Proceedings of the 62nd Annual Meeting of the Association for Computational Linguistics (Volume 1: Long Papers)}, pp.\  15568--15592, Bangkok, Thailand, August 2024. Association for Computational Linguistics.
\newblock \doi{10.18653/v1/2024.acl-long.832}.
\newblock URL \url{https://aclanthology.org/2024.acl-long.832/}.

\bibitem[Kwiatkowski et~al.(2019)Kwiatkowski, Palomaki, Redfield, Collins, Parikh, Alberti, Epstein, Polosukhin, Devlin, Lee, Toutanova, Jones, Kelcey, Chang, Dai, Uszkoreit, Le, and Petrov]{kwiatkowski2019natural}
Tom Kwiatkowski, Jennimaria Palomaki, Olivia Redfield, Michael Collins, Ankur Parikh, Chris Alberti, Danielle Epstein, Illia Polosukhin, Jacob Devlin, Kenton Lee, Kristina Toutanova, Llion Jones, Matthew Kelcey, Ming-Wei Chang, Andrew~M. Dai, Jakob Uszkoreit, Quoc Le, and Slav Petrov.
\newblock Natural questions: A benchmark for question answering research.
\newblock \emph{Transactions of the Association for Computational Linguistics}, 7:\penalty0 452--466, 2019.
\newblock \doi{10.1162/tacl_a_00276}.
\newblock URL \url{https://aclanthology.org/Q19-1026}.

\bibitem[Laban et~al.(2024)Laban, Fabbri, Xiong, and Wu]{laban2024summaryhaystackchallengelongcontext}
Philippe Laban, Alexander~R. Fabbri, Caiming Xiong, and Chien-Sheng Wu.
\newblock Summary of a haystack: A challenge to long-context llms and rag systems, 2024.
\newblock URL \url{https://arxiv.org/abs/2407.01370}.

\bibitem[Larson et~al.(2019)Larson, Mahendran, Peper, Clarke, Lee, Hill, Kummerfeld, Leach, Laurenzano, Tang, and Mars]{larson-etal-2019-evaluation}
Stefan Larson, Anish Mahendran, Joseph~J. Peper, Christopher Clarke, Andrew Lee, Parker Hill, Jonathan~K. Kummerfeld, Kevin Leach, Michael~A. Laurenzano, Lingjia Tang, and Jason Mars.
\newblock An evaluation dataset for intent classification and out-of-scope prediction.
\newblock In Kentaro Inui, Jing Jiang, Vincent Ng, and Xiaojun Wan (eds.), \emph{Proceedings of the 2019 Conference on Empirical Methods in Natural Language Processing and the 9th International Joint Conference on Natural Language Processing (EMNLP-IJCNLP)}, pp.\  1311--1316, Hong Kong, China, November 2019. Association for Computational Linguistics.
\newblock \doi{10.18653/v1/D19-1131}.
\newblock URL \url{https://aclanthology.org/D19-1131}.

\bibitem[Lee et~al.(2024)Lee, Chen, Dai, Dua, Sachan, Boratko, Luan, Arnold, Perot, Dalmia, Hu, Lin, Pasupat, Amini, Cole, Riedel, Naim, Chang, and Guu]{lee2024longcontextlanguagemodelssubsume}
Jinhyuk Lee, Anthony Chen, Zhuyun Dai, Dheeru Dua, Devendra~Singh Sachan, Michael Boratko, Yi~Luan, Sébastien M.~R. Arnold, Vincent Perot, Siddharth Dalmia, Hexiang Hu, Xudong Lin, Panupong Pasupat, Aida Amini, Jeremy~R. Cole, Sebastian Riedel, Iftekhar Naim, Ming-Wei Chang, and Kelvin Guu.
\newblock Can long-context language models subsume retrieval, rag, sql, and more?, 2024.
\newblock URL \url{https://arxiv.org/abs/2406.13121}.

\bibitem[Levy et~al.(2024)Levy, Jacoby, and Goldberg]{levy2024task}
Mosh Levy, Alon Jacoby, and Yoav Goldberg.
\newblock Same task, more tokens: the impact of input length on the reasoning performance of large language models.
\newblock In \emph{Association for Computational Linguistics (ACL)}, 2024.

\bibitem[Li et~al.(2024{\natexlab{a}})Li, Wang, Zheng, and Zhang]{li-etal-2024-loogle}
Jiaqi Li, Mengmeng Wang, Zilong Zheng, and Muhan Zhang.
\newblock {L}oo{GLE}: Can long-context language models understand long contexts?
\newblock In Lun-Wei Ku, Andre Martins, and Vivek Srikumar (eds.), \emph{Proceedings of the 62nd Annual Meeting of the Association for Computational Linguistics (Volume 1: Long Papers)}, pp.\  16304--16333, Bangkok, Thailand, August 2024{\natexlab{a}}. Association for Computational Linguistics.
\newblock URL \url{https://aclanthology.org/2024.acl-long.859}.

\bibitem[Li et~al.(2024{\natexlab{b}})Li, Zhang, Liu, and Chen]{li2024needlebenchllmsretrievalreasoning}
Mo~Li, Songyang Zhang, Yunxin Liu, and Kai Chen.
\newblock Needlebench: Can llms do retrieval and reasoning in 1 million context window?, 2024{\natexlab{b}}.
\newblock URL \url{https://arxiv.org/abs/2407.11963}.

\bibitem[Li et~al.(2024{\natexlab{c}})Li, Zhang, Do, Yue, and Chen]{li2024longiclbench}
Tianle Li, Ge~Zhang, Quy~Duc Do, Xiang Yue, and Wenhu Chen.
\newblock Long-context llms struggle with long in-context learning.
\newblock \emph{CoRR}, abs/2404.02060, 2024{\natexlab{c}}.
\newblock URL \url{https://doi.org/10.48550/arXiv.2404.02060}.

\bibitem[Li \& Roth(2002)Li and Roth]{li-roth-2002-learning}
Xin Li and Dan Roth.
\newblock Learning question classifiers.
\newblock In \emph{{COLING} 2002: The 19th International Conference on Computational Linguistics}, 2002.
\newblock URL \url{https://aclanthology.org/C02-1150}.

\bibitem[Lieber et~al.(2024)Lieber, Lenz, Bata, Cohen, Osin, Dalmedigos, Safahi, Meirom, Belinkov, Shalev-Shwartz, Abend, Alon, Asida, Bergman, Glozman, Gokhman, Manevich, Ratner, Rozen, Shwartz, Zusman, and Shoham]{lieber2024jambahybridtransformermambalanguage}
Opher Lieber, Barak Lenz, Hofit Bata, Gal Cohen, Jhonathan Osin, Itay Dalmedigos, Erez Safahi, Shaked Meirom, Yonatan Belinkov, Shai Shalev-Shwartz, Omri Abend, Raz Alon, Tomer Asida, Amir Bergman, Roman Glozman, Michael Gokhman, Avashalom Manevich, Nir Ratner, Noam Rozen, Erez Shwartz, Mor Zusman, and Yoav Shoham.
\newblock Jamba: A hybrid transformer-mamba language model, 2024.
\newblock URL \url{https://arxiv.org/abs/2403.19887}.

\bibitem[Lin(2004)]{lin-2004-rouge}
Chin-Yew Lin.
\newblock {ROUGE}: A package for automatic evaluation of summaries.
\newblock In \emph{Text Summarization Branches Out}, pp.\  74--81, Barcelona, Spain, July 2004. Association for Computational Linguistics.
\newblock URL \url{https://aclanthology.org/W04-1013}.

\bibitem[Liu et~al.(2023)Liu, Lin, Hewitt, Paranjape, Bevilacqua, Petroni, and Liang]{Liu2023LostIT}
Nelson~F. Liu, Kevin Lin, John Hewitt, Ashwin Paranjape, Michele Bevilacqua, Fabio Petroni, and Percy Liang.
\newblock Lost in the middle: How language models use long contexts.
\newblock \emph{Transactions of the Association for Computational Linguistics}, 12:\penalty0 157--173, 2023.
\newblock URL \url{https://api.semanticscholar.org/CorpusID:259360665}.

\bibitem[Liu et~al.(2019)Liu, Eshghi, Swietojanski, and Rieser]{Liu2019BenchmarkingNL}
Xingkun Liu, Arash Eshghi, Pawel Swietojanski, and Verena Rieser.
\newblock Benchmarking natural language understanding services for building conversational agents.
\newblock \emph{ArXiv}, abs/1903.05566, 2019.
\newblock URL \url{https://api.semanticscholar.org/CorpusID:76660838}.

\bibitem[Mallen et~al.(2023)Mallen, Asai, Zhong, Das, Khashabi, and Hajishirzi]{mallen-etal-2023-trust}
Alex Mallen, Akari Asai, Victor Zhong, Rajarshi Das, Daniel Khashabi, and Hannaneh Hajishirzi.
\newblock When not to trust language models: Investigating effectiveness of parametric and non-parametric memories.
\newblock In Anna Rogers, Jordan Boyd-Graber, and Naoaki Okazaki (eds.), \emph{Association for Computational Linguistics (ACL)}, pp.\  9802--9822, Toronto, Canada, July 2023. Association for Computational Linguistics.
\newblock \doi{10.18653/v1/2023.acl-long.546}.
\newblock URL \url{https://aclanthology.org/2023.acl-long.546}.

\bibitem[Min et~al.(2022)Min, Lyu, Holtzman, Artetxe, Lewis, Hajishirzi, and Zettlemoyer]{min-etal-2022-rethinking}
Sewon Min, Xinxi Lyu, Ari Holtzman, Mikel Artetxe, Mike Lewis, Hannaneh Hajishirzi, and Luke Zettlemoyer.
\newblock Rethinking the role of demonstrations: What makes in-context learning work?
\newblock In Yoav Goldberg, Zornitsa Kozareva, and Yue Zhang (eds.), \emph{Proceedings of the 2022 Conference on Empirical Methods in Natural Language Processing}, pp.\  11048--11064, Abu Dhabi, United Arab Emirates, December 2022. Association for Computational Linguistics.
\newblock \doi{10.18653/v1/2022.emnlp-main.759}.
\newblock URL \url{https://aclanthology.org/2022.emnlp-main.759}.

\bibitem[Ni et~al.(2022)Ni, Qu, Lu, Dai, Hernandez~Abrego, Ma, Zhao, Luan, Hall, Chang, and Yang]{ni-etal-2022-large}
Jianmo Ni, Chen Qu, Jing Lu, Zhuyun Dai, Gustavo Hernandez~Abrego, Ji~Ma, Vincent Zhao, Yi~Luan, Keith Hall, Ming-Wei Chang, and Yinfei Yang.
\newblock Large dual encoders are generalizable retrievers.
\newblock In \emph{Empirical Methods in Natural Language Processing (EMNLP)}, pp.\  9844--9855, 2022.
\newblock URL \url{https://aclanthology.org/2022.emnlp-main.669}.

\bibitem[OpenAI(2023)]{openai2023gpt4}
OpenAI.
\newblock {GPT-4 Technical Report}, 2023.

\bibitem[Pan et~al.(2023)Pan, Gao, Chen, and Chen]{pan-etal-2023-context}
Jane Pan, Tianyu Gao, Howard Chen, and Danqi Chen.
\newblock What in-context learning {``}learns{''} in-context: Disentangling task recognition and task learning.
\newblock In Anna Rogers, Jordan Boyd-Graber, and Naoaki Okazaki (eds.), \emph{Findings of the Association for Computational Linguistics: ACL 2023}, pp.\  8298--8319, Toronto, Canada, July 2023. Association for Computational Linguistics.
\newblock \doi{10.18653/v1/2023.findings-acl.527}.
\newblock URL \url{https://aclanthology.org/2023.findings-acl.527}.

\bibitem[Peng et~al.(2024)Peng, Quesnelle, Fan, and Shippole]{peng2024yarn}
Bowen Peng, Jeffrey Quesnelle, Honglu Fan, and Enrico Shippole.
\newblock Ya{RN}: Efficient context window extension of large language models.
\newblock In \emph{The Twelfth International Conference on Learning Representations}, 2024.
\newblock URL \url{https://openreview.net/forum?id=wHBfxhZu1u}.

\bibitem[Petroni et~al.(2021)Petroni, Piktus, Fan, Lewis, Yazdani, De~Cao, Thorne, Jernite, Karpukhin, Maillard, Plachouras, Rockt{\"a}schel, and Riedel]{petroni-etal-2021-kilt}
Fabio Petroni, Aleksandra Piktus, Angela Fan, Patrick Lewis, Majid Yazdani, Nicola De~Cao, James Thorne, Yacine Jernite, Vladimir Karpukhin, Jean Maillard, Vassilis Plachouras, Tim Rockt{\"a}schel, and Sebastian Riedel.
\newblock {KILT}: a benchmark for knowledge intensive language tasks.
\newblock In \emph{Proceedings of the 2021 Conference of the North American Chapter of the Association for Computational Linguistics: Human Language Technologies}, pp.\  2523--2544, Online, June 2021. Association for Computational Linguistics.
\newblock \doi{10.18653/v1/2021.naacl-main.200}.
\newblock URL \url{https://aclanthology.org/2021.naacl-main.200}.

\bibitem[Qwen et~al.(2025)Qwen, :, Yang, Yang, Zhang, Hui, Zheng, Yu, Li, Liu, Huang, Wei, Lin, Yang, Tu, Zhang, Yang, Yang, Zhou, Lin, Dang, Lu, Bao, Yang, Yu, Li, Xue, Zhang, Zhu, Men, Lin, Li, Tang, Xia, Ren, Ren, Fan, Su, Zhang, Wan, Liu, Cui, Zhang, and Qiu]{qwen2025qwen25technicalreport}
Qwen, :, An~Yang, Baosong Yang, Beichen Zhang, Binyuan Hui, Bo~Zheng, Bowen Yu, Chengyuan Li, Dayiheng Liu, Fei Huang, Haoran Wei, Huan Lin, Jian Yang, Jianhong Tu, Jianwei Zhang, Jianxin Yang, Jiaxi Yang, Jingren Zhou, Junyang Lin, Kai Dang, Keming Lu, Keqin Bao, Kexin Yang, Le~Yu, Mei Li, Mingfeng Xue, Pei Zhang, Qin Zhu, Rui Men, Runji Lin, Tianhao Li, Tianyi Tang, Tingyu Xia, Xingzhang Ren, Xuancheng Ren, Yang Fan, Yang Su, Yichang Zhang, Yu~Wan, Yuqiong Liu, Zeyu Cui, Zhenru Zhang, and Zihan Qiu.
\newblock Qwen2.5 technical report, 2025.
\newblock URL \url{https://arxiv.org/abs/2412.15115}.

\bibitem[Rajpurkar et~al.(2016)Rajpurkar, Zhang, Lopyrev, and Liang]{rajpurkar2016squad}
Pranav Rajpurkar, Jian Zhang, Konstantin Lopyrev, and Percy Liang.
\newblock {SQ}u{AD}: 100,000+ questions for machine comprehension of text.
\newblock In \emph{Empirical Methods in Natural Language Processing (EMNLP)}, 2016.
\newblock URL \url{https://aclanthology.org/D16-1264/}.

\bibitem[Ratner et~al.(2023)Ratner, Levine, Belinkov, Ram, Magar, Abend, Karpas, Shashua, Leyton-Brown, and Shoham]{ratner-etal-2023-parallel}
Nir Ratner, Yoav Levine, Yonatan Belinkov, Ori Ram, Inbal Magar, Omri Abend, Ehud Karpas, Amnon Shashua, Kevin Leyton-Brown, and Yoav Shoham.
\newblock Parallel context windows for large language models.
\newblock In Anna Rogers, Jordan Boyd-Graber, and Naoaki Okazaki (eds.), \emph{Proceedings of the 61st Annual Meeting of the Association for Computational Linguistics (Volume 1: Long Papers)}, pp.\  6383--6402, Toronto, Canada, July 2023. Association for Computational Linguistics.
\newblock \doi{10.18653/v1/2023.acl-long.352}.
\newblock URL \url{https://aclanthology.org/2023.acl-long.352}.

\bibitem[Robertson \& Zaragoza(2009)Robertson and Zaragoza]{robertson2009bm25}
Stephen Robertson and Hugo Zaragoza.
\newblock The probabilistic relevance framework: Bm25 and beyond.
\newblock \emph{Found. Trends Inf. Retr.}, 3\penalty0 (4):\penalty0 333–389, apr 2009.
\newblock ISSN 1554-0669.
\newblock \doi{10.1561/1500000019}.
\newblock URL \url{https://doi.org/10.1561/1500000019}.

\bibitem[Rubin et~al.(2022)Rubin, Yoran, Wolfson, Herzig, and Berant]{rubin2022qampari}
Samuel Joseph Amouyal~Ohad Rubin, Ori Yoran, Tomer Wolfson, Jonathan Herzig, and Jonathan Berant.
\newblock {QAMPARI: An Open-domain Question Answering Benchmark for Questions with Many Answers from Multiple Paragraphs}.
\newblock \emph{arXiv preprint arXiv:2205.12665}, 2022.
\newblock URL \url{https://arxiv.org/abs/2205.12665}.

\bibitem[Shaham et~al.(2022)Shaham, Segal, Ivgi, Efrat, Yoran, Haviv, Gupta, Xiong, Geva, Berant, and Levy]{shaham-etal-2022-scrolls}
Uri Shaham, Elad Segal, Maor Ivgi, Avia Efrat, Ori Yoran, Adi Haviv, Ankit Gupta, Wenhan Xiong, Mor Geva, Jonathan Berant, and Omer Levy.
\newblock {SCROLLS}: Standardized {C}ompa{R}ison over long language sequences.
\newblock In \emph{Proceedings of the 2022 Conference on Empirical Methods in Natural Language Processing}, pp.\  12007--12021, Abu Dhabi, United Arab Emirates, December 2022. Association for Computational Linguistics.
\newblock URL \url{https://aclanthology.org/2022.emnlp-main.823}.

\bibitem[Shaham et~al.(2023)Shaham, Ivgi, Efrat, Berant, and Levy]{shaham-etal-2023-zeroscrolls}
Uri Shaham, Maor Ivgi, Avia Efrat, Jonathan Berant, and Omer Levy.
\newblock {Z}ero{SCROLLS}: A zero-shot benchmark for long text understanding.
\newblock In Houda Bouamor, Juan Pino, and Kalika Bali (eds.), \emph{Findings of the Association for Computational Linguistics: EMNLP 2023}, pp.\  7977--7989, Singapore, December 2023. Association for Computational Linguistics.
\newblock \doi{10.18653/v1/2023.findings-emnlp.536}.
\newblock URL \url{https://aclanthology.org/2023.findings-emnlp.536}.

\bibitem[Shen et~al.(2022)Shen, Lo, Yu, Dahlberg, Schlanger, and Downey]{shen2022multilexsum}
Zejiang Shen, Kyle Lo, Lauren Yu, Nathan Dahlberg, Margo Schlanger, and Doug Downey.
\newblock Multi-lexsum: Real-world summaries of civil rights lawsuits at multiple granularities.
\newblock In \emph{Thirty-sixth Conference on Neural Information Processing Systems Datasets and Benchmarks Track}, 2022.
\newblock URL \url{https://openreview.net/forum?id=z1d8fUiS8Cr}.

\bibitem[Stelmakh et~al.(2022)Stelmakh, Luan, Dhingra, and Chang]{stelmakh2022asqa}
Ivan Stelmakh, Yi~Luan, Bhuwan Dhingra, and Ming-Wei Chang.
\newblock {ASQA}: Factoid questions meet long-form answers.
\newblock In \emph{Proceedings of the 2022 Conference on Empirical Methods in Natural Language Processing}, pp.\  8273--8288, Abu Dhabi, United Arab Emirates, December 2022. Association for Computational Linguistics.
\newblock \doi{10.18653/v1/2022.emnlp-main.566}.
\newblock URL \url{https://aclanthology.org/2022.emnlp-main.566}.

\bibitem[Su et~al.(2021)Su, Lu, Pan, Murtadha, Wen, and Liu]{su2021roformer}
Jianlin Su, Yu~Lu, Shengfeng Pan, Ahmed Murtadha, Bo~Wen, and Yunfeng Liu.
\newblock Roformer: Enhanced transformer with rotary position embedding, 2021.
\newblock URL \url{https://arxiv.org/abs/2104.09864}.

\bibitem[Sun et~al.(2023)Sun, Yan, Ma, Wang, Ren, Chen, Yin, and Ren]{sun-etal-2023-chatgpt}
Weiwei Sun, Lingyong Yan, Xinyu Ma, Shuaiqiang Wang, Pengjie Ren, Zhumin Chen, Dawei Yin, and Zhaochun Ren.
\newblock Is {C}hat{GPT} good at search? investigating large language models as re-ranking agents.
\newblock In Houda Bouamor, Juan Pino, and Kalika Bali (eds.), \emph{Proceedings of the 2023 Conference on Empirical Methods in Natural Language Processing}, pp.\  14918--14937, Singapore, December 2023. Association for Computational Linguistics.
\newblock \doi{10.18653/v1/2023.emnlp-main.923}.
\newblock URL \url{https://aclanthology.org/2023.emnlp-main.923}.

\bibitem[Tay et~al.(2021)Tay, Dehghani, Abnar, Shen, Bahri, Pham, Rao, Yang, Ruder, and Metzler]{tay2021long}
Yi~Tay, Mostafa Dehghani, Samira Abnar, Yikang Shen, Dara Bahri, Philip Pham, Jinfeng Rao, Liu Yang, Sebastian Ruder, and Donald Metzler.
\newblock Long range arena : A benchmark for efficient transformers.
\newblock In \emph{International Conference on Learning Representations}, 2021.
\newblock URL \url{https://openreview.net/forum?id=qVyeW-grC2k}.

\bibitem[Team et~al.(2024{\natexlab{a}})Team, Anil, Borgeaud, Alayrac, Yu, Soricut, Schalkwyk, Dai, Hauth, Millican, Silver, Johnson, Antonoglou, Schrittwieser, Glaese, Chen, Pitler, Lillicrap, Lazaridou, Firat, Molloy, Isard, Barham, Hennigan, Lee, Viola, Reynolds, Xu, Doherty, Collins, Meyer, Rutherford, Moreira, Ayoub, Goel, Krawczyk, Du, Chi, Cheng, Ni, Shah, Kane, Chan, Faruqui, Severyn, Lin, Li, Cheng, Ittycheriah, Mahdieh, Chen, Sun, Tran, Bagri, Lakshminarayanan, Liu, Orban, Güra, Zhou, Song, Boffy, Ganapathy, Zheng, Choe, Ágoston Weisz, Zhu, Lu, Gopal, Kahn, Kula, Pitman, Shah, Taropa, Merey, Baeuml, Chen, Shafey, Zhang, Sercinoglu, Tucker, Piqueras, Krikun, Barr, Savinov, Danihelka, Roelofs, White, Andreassen, von Glehn, Yagati, Kazemi, Gonzalez, Khalman, Sygnowski, Frechette, Smith, Culp, Proleev, Luan, Chen, Lottes, Schucher, Lebron, Rrustemi, Clay, Crone, Kocisky, Zhao, Perz, Yu, Howard, Bloniarz, Rae, Lu, Sifre, Maggioni, Alcober, Garrette, Barnes, Thakoor, Austin, Barth-Maron, Wong, Joshi, Chaabouni, Fatiha, Ahuja, Tomar, Senter, Chadwick, Kornakov, Attaluri, Iturrate, Liu, Li, Cogan, Chen, Jia, Gu, Zhang, Grimstad, Hartman, Garcia, Pillai, Devlin, Laskin, de~Las~Casas, Valter, Tao, Blanco, Badia, Reitter, Chen, Brennan, Rivera, Brin, Iqbal, Surita, Labanowski, Rao, Winkler, Parisotto, Gu, Olszewska, Addanki, Miech, Louis, Teplyashin, Brown, Catt, Balaguer, Xiang, Wang, Ashwood, Briukhov, Webson, Ganapathy, Sanghavi, Kannan, Chang, Stjerngren, Djolonga, Sun, Bapna, Aitchison, Pejman, Michalewski, Yu, Wang, Love, Ahn, Bloxwich, Han, Humphreys, Sellam, Bradbury, Godbole, Samangooei, Damoc, Kaskasoli, Arnold, Vasudevan, Agrawal, Riesa, Lepikhin, Tanburn, Srinivasan, Lim, Hodkinson, Shyam, Ferret, Hand, Garg, Paine, Li, Li, Giang, Neitz, Abbas, York, Reid, Cole, Chowdhery, Das, Rogozińska, Nikolaev, Sprechmann, Nado, Zilka, Prost, He, Monteiro, Mishra, Welty, Newlan, Jia, Allamanis, Hu, de~Liedekerke, Gilmer, Saroufim, Rijhwani, Hou, Shrivastava, Baddepudi, Goldin, Ozturel, Cassirer, Xu, Sohn, Sachan, Amplayo, Swanson, Petrova, Narayan, Guez, Brahma, Landon, Patel, Zhao, Villela, Wang, Jia, Rahtz, Giménez, Yeung, Keeling, Georgiev, Mincu, Wu, Haykal, Saputro, Vodrahalli, Qin, Cankara, Sharma, Fernando, Hawkins, Neyshabur, Kim, Hutter, Agrawal, Castro-Ros, van~den Driessche, Wang, Yang, yiin Chang, Komarek, McIlroy, Lučić, Zhang, Farhan, Sharman, Natsev, Michel, Bansal, Qiao, Cao, Shakeri, Butterfield, Chung, Rubenstein, Agrawal, Mensch, Soparkar, Lenc, Chung, Pope, Maggiore, Kay, Jhakra, Wang, Maynez, Phuong, Tobin, Tacchetti, Trebacz, Robinson, Katariya, Riedel, Bailey, Xiao, Ghelani, Aroyo, Slone, Houlsby, Xiong, Yang, Gribovskaya, Adler, Wirth, Lee, Li, Kagohara, Pavagadhi, Bridgers, Bortsova, Ghemawat, Ahmed, Liu, Powell, Bolina, Iinuma, Zablotskaia, Besley, Chung, Dozat, Comanescu, Si, Greer, Su, Polacek, Kaufman, Tokumine, Hu, Buchatskaya, Miao, Elhawaty, Siddhant, Tomasev, Xing, Greer, Miller, Ashraf, Roy, Zhang, Ma, Filos, Besta, Blevins, Klimenko, Yeh, Changpinyo, Mu, Chang, Pajarskas, Muir, Cohen, Lan, Haridasan, Marathe, Hansen, Douglas, Samuel, Wang, Austin, Lan, Jiang, Chiu, Lorenzo, Sjösund, Cevey, Gleicher, Avrahami, Boral, Srinivasan, Selo, May, Aisopos, Hussenot, Soares, Baumli, Chang, Recasens, Caine, Pritzel, Pavetic, Pardo, Gergely, Frye, Ramasesh, Horgan, Badola, Kassner, Roy, Dyer, Campos, Tomala, Tang, Badawy, White, Mustafa, Lang, Jindal, Vikram, Gong, Caelles, Hemsley, Thornton, Feng, Stokowiec, Zheng, Thacker, Çağlar Ünlü, Zhang, Saleh, Svensson, Bileschi, Patil, Anand, Ring, Tsihlas, Vezer, Selvi, Shevlane, Rodriguez, Kwiatkowski, Daruki, Rong, Dafoe, FitzGerald, Gu-Lemberg, Khan, Hendricks, Pellat, Feinberg, Cobon-Kerr, Sainath, Rauh, Hashemi, Ives, Hasson, Noland, Cao, Byrd, Hou, Wang, Sottiaux, Paganini, Lespiau, Moufarek, Hassan, Shivakumar, van Amersfoort, Mandhane, Joshi, Goyal, Tung, Brock, Sheahan, Misra, Li, Rakićević, Dehghani, Liu, Mittal, Oh, Noury, Sezener, Huot, Lamm, Cao, Chen, Mudgal, Stella, Brooks, Vasudevan, Liu, Chain, Melinkeri, Cohen, Wang, Seymore, Zubkov, Goel, Yue, Krishnakumaran, Albert, Hurley, Sano, Mohananey, Joughin, Filonov, Kepa, Eldawy, Lim, Rishi, Badiezadegan, Bos, Chang, Jain, Padmanabhan, Puttagunta, Krishna, Baker, Kalb, Bedapudi, Kurzrok, Lei, Yu, Litvin, Zhou, Wu, Sobell, Siciliano, Papir, Neale, Bragagnolo, Toor, Chen, Anklin, Wang, Feng, Gholami, Ling, Liu, Walter, Moghaddam, Kishore, Adamek, Mercado, Mallinson, Wandekar, Cagle, Ofek, Garrido, Lombriser, Mukha, Sun, Mohammad, Matak, Qian, Peswani, Janus, Yuan, Schelin, David, Garg, He, Duzhyi, Älgmyr, Lottaz, Li, Yadav, Xu, Chinien, Shivanna, Chuklin, Li, Spadine, Wolfe, Mohamed, Das, Dai, He, von Dincklage, Upadhyay, Maurya, Chi, Krause, Salama, Rabinovitch, M, Selvan, Dektiarev, Ghiasi, Guven, Gupta, Liu, Sharma, Shtacher, Paul, Akerlund, Aubet, Huang, Zhu, Zhu, Teixeira, Fritze, Bertolini, Marinescu, Bölle, Paulus, Gupta, Latkar, Chang, Sanders, Wilson, Wu, Tan, Thiet, Doshi, Lall, Mishra, Chen, Luong, Benjamin, Lee, Andrejczuk, Rabiej, Ranjan, Styrc, Yin, Simon, Harriott, Bansal, Robsky, Bacon, Greene, Mirylenka, Zhou, Sarvana, Goyal, Andermatt, Siegler, Horn, Israel, Pongetti, Chen, Selvatici, Silva, Wang, Tolins, Guu, Yogev, Cai, Agostini, Shah, Nguyen, Donnaile, Pereira, Friso, Stambler, Kurzrok, Kuang, Romanikhin, Geller, Yan, Jang, Lee, Fica, Malmi, Tan, Banica, Balle, Pham, Huang, Avram, Shi, Singh, Hidey, Ahuja, Saxena, Dooley, Potharaju, O'Neill, Gokulchandran, Foley, Zhao, Dusenberry, Liu, Mehta, Kotikalapudi, Safranek-Shrader, Goodman, Kessinger, Globen, Kolhar, Gorgolewski, Ibrahim, Song, Eichenbaum, Brovelli, Potluri, Lahoti, Baetu, Ghorbani, Chen, Crawford, Pal, Sridhar, Gurita, Mujika, Petrovski, Cedoz, Li, Chen, Santo, Goyal, Punjabi, Kappaganthu, Kwak, LV, Velury, Choudhury, Hall, Shah, Figueira, Thomas, Lu, Zhou, Kumar, Jurdi, Chikkerur, Ma, Yu, Kwak, Ähdel, Rajayogam, Choma, Liu, Barua, Ji, Park, Hellendoorn, Bailey, Bilal, Zhou, Khatir, Sutton, Rzadkowski, Macintosh, Shagin, Medina, Liang, Zhou, Shah, Bi, Dankovics, Banga, Lehmann, Bredesen, Lin, Hoffmann, Lai, Chung, Yang, Balani, Bražinskas, Sozanschi, Hayes, Alcalde, Makarov, Chen, Stella, Snijders, Mandl, Kärrman, Nowak, Wu, Dyck, Vaidyanathan, R, Mallet, Rudominer, Johnston, Mittal, Udathu, Christensen, Verma, Irving, Santucci, Elsayed, Davoodi, Georgiev, Tenney, Hua, Cideron, Leurent, Alnahlawi, Georgescu, Wei, Zheng, Scandinaro, Jiang, Snoek, Sundararajan, Wang, Ontiveros, Karo, Cole, Rajashekhar, Tumeh, Ben-David, Jain, Uesato, Datta, Bunyan, Wu, Zhang, Stanczyk, Zhang, Steiner, Naskar, Azzam, Johnson, Paszke, Chiu, Elias, Mohiuddin, Muhammad, Miao, Lee, Vieillard, Park, Zhang, Stanway, Garmon, Karmarkar, Dong, Lee, Kumar, Zhou, Evens, Isaac, Irving, Loper, Fink, Arkatkar, Chen, Shafran, Petrychenko, Chen, Jia, Levskaya, Zhu, Grabowski, Mao, Magni, Yao, Snaider, Casagrande, Palmer, Suganthan, Castaño, Giannoumis, Kim, Rybiński, Sreevatsa, Prendki, Soergel, Goedeckemeyer, Gierke, Jafari, Gaba, Wiesner, Wright, Wei, Vashisht, Kulizhskaya, Hoover, Le, Li, Iwuanyanwu, Liu, Ramirez, Khorlin, Cui, LIN, Wu, Aguilar, Pallo, Chakladar, Perng, Abellan, Zhang, Dasgupta, Kushman, Penchev, Repina, Wu, van~der Weide, Ponnapalli, Kaplan, Simsa, Li, Dousse, Yang, Piper, Ie, Pasumarthi, Lintz, Vijayakumar, Andor, Valenzuela, Lui, Paduraru, Peng, Lee, Zhang, Greene, Nguyen, Kurylowicz, Hardin, Dixon, Janzer, Choo, Feng, Zhang, Singhal, Du, McKinnon, Antropova, Bolukbasi, Keller, Reid, Finchelstein, Raad, Crocker, Hawkins, Dadashi, Gaffney, Franko, Bulanova, Leblond, Chung, Askham, Cobo, Xu, Fischer, Xu, Sorokin, Alberti, Lin, Evans, Dimitriev, Forbes, Banarse, Tung, Omernick, Bishop, Sterneck, Jain, Xia, Amid, Piccinno, Wang, Banzal, Mankowitz, Polozov, Krakovna, Brown, Bateni, Duan, Firoiu, Thotakuri, Natan, Geist, tan Girgin, Li, Ye, Roval, Tojo, Kwong, Lee-Thorp, Yew, Sinopalnikov, Ramos, Mellor, Sharma, Wu, Miller, Sonnerat, Vnukov, Greig, Beattie, Caveness, Bai, Eisenschlos, Korchemniy, Tsai, Jasarevic, Kong, Dao, Zheng, Liu, Yang, Zhu, Teh, Sanmiya, Gladchenko, Trdin, Toyama, Rosen, Tavakkol, Xue, Elkind, Woodman, Carpenter, Papamakarios, Kemp, Kafle, Grunina, Sinha, Talbert, Wu, Owusu-Afriyie, Du, Thornton, Pont-Tuset, Narayana, Li, Fatehi, Wieting, Ajmeri, Uria, Ko, Knight, Héliou, Niu, Gu, Pang, Li, Levine, Stolovich, Santamaria-Fernandez, Goenka, Yustalim, Strudel, Elqursh, Deck, Lee, Li, Levin, Hoffmann, Holtmann-Rice, Bachem, Arora, Koh, Yeganeh, Põder, Tariq, Sun, Ionita, Seyedhosseini, Tafti, Liu, Gulati, Liu, Ye, Chrzaszcz, Wang, Sethi, Li, Brown, Singh, Fan, Parisi, Stanton, Koverkathu, Choquette-Choo, Li, Lu, Ittycheriah, Shroff, Varadarajan, Bahargam, Willoughby, Gaddy, Desjardins, Cornero, Robenek, Mittal, Albrecht, Shenoy, Moiseev, Jacobsson, Ghaffarkhah, Rivière, Walton, Crepy, Parrish, Zhou, Farabet, Radebaugh, Srinivasan, van~der Salm, Fidjeland, Scellato, Latorre-Chimoto, Klimczak-Plucińska, Bridson, de~Cesare, Hudson, Mendolicchio, Walker, Morris, Mauger, Guseynov, Reid, Odoom, Loher, Cotruta, Yenugula, Grewe, Petrushkina, Duerig, Sanchez, Yadlowsky, Shen, Globerson, Webb, Dua, Li, Bhupatiraju, Hurt, Qureshi, Agarwal, Shani, Eyal, Khare, Belle, Wang, Tekur, Kale, Wei, Sang, Saeta, Liechty, Sun, Zhao, Lee, Nayak, Fritz, Vuyyuru, Aslanides, Vyas, Wicke, Ma, Eltyshev, Martin, Cate, Manyika, Amiri, Kim, Xiong, Kang, Luisier, Tripuraneni, Madras, Guo, Waters, Wang, Ainslie, Baldridge, Zhang, Pruthi, Bauer, Yang, Mansour, Gelman, Xu, Polovets, Liu, Cai, Chen, Sheng, Xue, Ozair, Angermueller, Li, Sinha, Wang, Wiesinger, Koukoumidis, Tian, Iyer, Gurumurthy, Goldenson, Shah, Blake, Yu, Urbanowicz, Palomaki, Fernando, Durden, Mehta, Momchev, Rahimtoroghi, Georgaki, Raul, Ruder, Redshaw, Lee, Zhou, Jalan, Li, Hechtman, Schuh, Nasr, Milan, Mikulik, Franco, Green, Nguyen, Kelley, Mahendru, Hu, Howland, Vargas, Hui, Bansal, Rao, Ghiya, Wang, Ye, Sarr, Preston, Elish, Li, Kaku, Gupta,
  Pasupat, Juan, Someswar, M., Chen, Amini, Fabrikant, Chu, Dong, Muthal, Buthpitiya, Jauhari, Hua, Khandelwal, Hitron, Ren, Rinaldi, Drath, Dabush, Jiang, Godhia, Sachs, Chen, Fan, Taitelbaum, Noga, Dai, Wang, Liang, Hamer, Ferng, Elkind, Atias, Lee, Listík, Carlen, van~de Kerkhof, Pikus, Zaher, Müller, Zykova, Stefanec, Gatsko, Hirnschall, Sethi, Xu, Ahuja, Tsai, Stefanoiu, Feng, Dhandhania, Katyal, Gupta, Parulekar, Pitta, Zhao, Bhatia, Bhavnani, Alhadlaq, Li, Danenberg, Tu, Pine, Filippova, Ghosh, Limonchik, Urala, Lanka, Clive, Sun, Li, Wu, Hongtongsak, Li, Thakkar, Omarov, Majmundar, Alverson, Kucharski, Patel, Jain, Zabelin, Pelagatti, Kohli, Kumar, Kim, Sankar, Shah, Ramachandruni, Zeng, Bariach, Weidinger, Vu, Andreev, He, Hui, Kashem, Subramanya, Hsiao, Hassabis, Kavukcuoglu, Sadovsky, Le, Strohman, Wu, Petrov, Dean, and Vinyals]{geminiteam2024geminifamilyhighlycapable}
Gemini Team, Rohan Anil, Sebastian Borgeaud, Jean-Baptiste Alayrac, Jiahui Yu, Radu Soricut, Johan Schalkwyk, Andrew~M. Dai, Anja Hauth, Katie Millican, David Silver, Melvin Johnson, Ioannis Antonoglou, Julian Schrittwieser, Amelia Glaese, Jilin Chen, Emily Pitler, Timothy Lillicrap, Angeliki Lazaridou, Orhan Firat, James Molloy, Michael Isard, Paul~R. Barham, Tom Hennigan, Benjamin Lee, Fabio Viola, Malcolm Reynolds, Yuanzhong Xu, Ryan Doherty, Eli Collins, Clemens Meyer, Eliza Rutherford, Erica Moreira, Kareem Ayoub, Megha Goel, Jack Krawczyk, Cosmo Du, Ed~Chi, Heng-Tze Cheng, Eric Ni, Purvi Shah, Patrick Kane, Betty Chan, Manaal Faruqui, Aliaksei Severyn, Hanzhao Lin, YaGuang Li, Yong Cheng, Abe Ittycheriah, Mahdis Mahdieh, Mia Chen, Pei Sun, Dustin Tran, Sumit Bagri, Balaji Lakshminarayanan, Jeremiah Liu, Andras Orban, Fabian Güra, Hao Zhou, Xinying Song, Aurelien Boffy, Harish Ganapathy, Steven Zheng, HyunJeong Choe, Ágoston Weisz, Tao Zhu, Yifeng Lu, Siddharth Gopal, Jarrod Kahn, Maciej Kula, Jeff Pitman, Rushin Shah, Emanuel Taropa, Majd~Al Merey, Martin Baeuml, Zhifeng Chen, Laurent~El Shafey, Yujing Zhang, Olcan Sercinoglu, George Tucker, Enrique Piqueras, Maxim Krikun, Iain Barr, Nikolay Savinov, Ivo Danihelka, Becca Roelofs, Anaïs White, Anders Andreassen, Tamara von Glehn, Lakshman Yagati, Mehran Kazemi, Lucas Gonzalez, Misha Khalman, Jakub Sygnowski, Alexandre Frechette, Charlotte Smith, Laura Culp, Lev Proleev, Yi~Luan, Xi~Chen, James Lottes, Nathan Schucher, Federico Lebron, Alban Rrustemi, Natalie Clay, Phil Crone, Tomas Kocisky, Jeffrey Zhao, Bartek Perz, Dian Yu, Heidi Howard, Adam Bloniarz, Jack~W. Rae, Han Lu, Laurent Sifre, Marcello Maggioni, Fred Alcober, Dan Garrette, Megan Barnes, Shantanu Thakoor, Jacob Austin, Gabriel Barth-Maron, William Wong, Rishabh Joshi, Rahma Chaabouni, Deeni Fatiha, Arun Ahuja, Gaurav~Singh Tomar, Evan Senter, Martin Chadwick, Ilya Kornakov, Nithya Attaluri, Iñaki Iturrate, Ruibo Liu, Yunxuan Li, Sarah Cogan, Jeremy Chen, Chao Jia, Chenjie Gu, Qiao Zhang, Jordan Grimstad, Ale~Jakse Hartman, Xavier Garcia, Thanumalayan~Sankaranarayana Pillai, Jacob Devlin, Michael Laskin, Diego de~Las~Casas, Dasha Valter, Connie Tao, Lorenzo Blanco, Adrià~Puigdomènech Badia, David Reitter, Mianna Chen, Jenny Brennan, Clara Rivera, Sergey Brin, Shariq Iqbal, Gabriela Surita, Jane Labanowski, Abhi Rao, Stephanie Winkler, Emilio Parisotto, Yiming Gu, Kate Olszewska, Ravi Addanki, Antoine Miech, Annie Louis, Denis Teplyashin, Geoff Brown, Elliot Catt, Jan Balaguer, Jackie Xiang, Pidong Wang, Zoe Ashwood, Anton Briukhov, Albert Webson, Sanjay Ganapathy, Smit Sanghavi, Ajay Kannan, Ming-Wei Chang, Axel Stjerngren, Josip Djolonga, Yuting Sun, Ankur Bapna, Matthew Aitchison, Pedram Pejman, Henryk Michalewski, Tianhe Yu, Cindy Wang, Juliette Love, Junwhan Ahn, Dawn Bloxwich, Kehang Han, Peter Humphreys, Thibault Sellam, James Bradbury, Varun Godbole, Sina Samangooei, Bogdan Damoc, Alex Kaskasoli, Sébastien M.~R. Arnold, Vijay Vasudevan, Shubham Agrawal, Jason Riesa, Dmitry Lepikhin, Richard Tanburn, Srivatsan Srinivasan, Hyeontaek Lim, Sarah Hodkinson, Pranav Shyam, Johan Ferret, Steven Hand, Ankush Garg, Tom~Le Paine, Jian Li, Yujia Li, Minh Giang, Alexander Neitz, Zaheer Abbas, Sarah York, Machel Reid, Elizabeth Cole, Aakanksha Chowdhery, Dipanjan Das, Dominika Rogozińska, Vitaliy Nikolaev, Pablo Sprechmann, Zachary Nado, Lukas Zilka, Flavien Prost, Luheng He, Marianne Monteiro, Gaurav Mishra, Chris Welty, Josh Newlan, Dawei Jia, Miltiadis Allamanis, Clara~Huiyi Hu, Raoul de~Liedekerke, Justin Gilmer, Carl Saroufim, Shruti Rijhwani, Shaobo Hou, Disha Shrivastava, Anirudh Baddepudi, Alex Goldin, Adnan Ozturel, Albin Cassirer, Yunhan Xu, Daniel Sohn, Devendra Sachan, Reinald~Kim Amplayo, Craig Swanson, Dessie Petrova, Shashi Narayan, Arthur Guez, Siddhartha Brahma, Jessica Landon, Miteyan Patel, Ruizhe Zhao, Kevin Villela, Luyu Wang, Wenhao Jia, Matthew Rahtz, Mai Giménez, Legg Yeung, James Keeling, Petko Georgiev, Diana Mincu, Boxi Wu, Salem Haykal, Rachel Saputro, Kiran Vodrahalli, James Qin, Zeynep Cankara, Abhanshu Sharma, Nick Fernando, Will Hawkins, Behnam Neyshabur, Solomon Kim, Adrian Hutter, Priyanka Agrawal, Alex Castro-Ros, George van~den Driessche, Tao Wang, Fan Yang, Shuo yiin Chang, Paul Komarek, Ross McIlroy, Mario Lučić, Guodong Zhang, Wael Farhan, Michael Sharman, Paul Natsev, Paul Michel, Yamini Bansal, Siyuan Qiao, Kris Cao, Siamak Shakeri, Christina Butterfield, Justin Chung, Paul~Kishan Rubenstein, Shivani Agrawal, Arthur Mensch, Kedar Soparkar, Karel Lenc, Timothy Chung, Aedan Pope, Loren Maggiore, Jackie Kay, Priya Jhakra, Shibo Wang, Joshua Maynez, Mary Phuong, Taylor Tobin, Andrea Tacchetti, Maja Trebacz, Kevin Robinson, Yash Katariya, Sebastian Riedel, Paige Bailey, Kefan Xiao, Nimesh Ghelani, Lora Aroyo, Ambrose Slone, Neil Houlsby, Xuehan Xiong, Zhen Yang, Elena Gribovskaya, Jonas Adler, Mateo Wirth, Lisa Lee, Music Li, Thais Kagohara, Jay Pavagadhi, Sophie Bridgers, Anna Bortsova, Sanjay Ghemawat, Zafarali Ahmed, Tianqi Liu, Richard Powell, Vijay Bolina, Mariko Iinuma, Polina Zablotskaia, James Besley, Da-Woon Chung, Timothy Dozat, Ramona Comanescu, Xiance Si, Jeremy Greer, Guolong Su, Martin Polacek, Raphaël~Lopez Kaufman, Simon Tokumine, Hexiang Hu, Elena Buchatskaya, Yingjie Miao, Mohamed Elhawaty, Aditya Siddhant, Nenad Tomasev, Jinwei Xing, Christina Greer, Helen Miller, Shereen Ashraf, Aurko Roy, Zizhao Zhang, Ada Ma, Angelos Filos, Milos Besta, Rory Blevins, Ted Klimenko, Chih-Kuan Yeh, Soravit Changpinyo, Jiaqi Mu, Oscar Chang, Mantas Pajarskas, Carrie Muir, Vered Cohen, Charline~Le Lan, Krishna Haridasan, Amit Marathe, Steven Hansen, Sholto Douglas, Rajkumar Samuel, Mingqiu Wang, Sophia Austin, Chang Lan, Jiepu Jiang, Justin Chiu, Jaime~Alonso Lorenzo, Lars~Lowe Sjösund, Sébastien Cevey, Zach Gleicher, Thi Avrahami, Anudhyan Boral, Hansa Srinivasan, Vittorio Selo, Rhys May, Konstantinos Aisopos, Léonard Hussenot, Livio~Baldini Soares, Kate Baumli, Michael~B. Chang, Adrià Recasens, Ben Caine, Alexander Pritzel, Filip Pavetic, Fabio Pardo, Anita Gergely, Justin Frye, Vinay Ramasesh, Dan Horgan, Kartikeya Badola, Nora Kassner, Subhrajit Roy, Ethan Dyer, Víctor~Campos Campos, Alex Tomala, Yunhao Tang, Dalia~El Badawy, Elspeth White, Basil Mustafa, Oran Lang, Abhishek Jindal, Sharad Vikram, Zhitao Gong, Sergi Caelles, Ross Hemsley, Gregory Thornton, Fangxiaoyu Feng, Wojciech Stokowiec, Ce~Zheng, Phoebe Thacker, Çağlar Ünlü, Zhishuai Zhang, Mohammad Saleh, James Svensson, Max Bileschi, Piyush Patil, Ankesh Anand, Roman Ring, Katerina Tsihlas, Arpi Vezer, Marco Selvi, Toby Shevlane, Mikel Rodriguez, Tom Kwiatkowski, Samira Daruki, Keran Rong, Allan Dafoe, Nicholas FitzGerald, Keren Gu-Lemberg, Mina Khan, Lisa~Anne Hendricks, Marie Pellat, Vladimir Feinberg, James Cobon-Kerr, Tara Sainath, Maribeth Rauh, Sayed~Hadi Hashemi, Richard Ives, Yana Hasson, Eric Noland, Yuan Cao, Nathan Byrd, Le~Hou, Qingze Wang, Thibault Sottiaux, Michela Paganini, Jean-Baptiste Lespiau, Alexandre Moufarek, Samer Hassan, Kaushik Shivakumar, Joost van Amersfoort, Amol Mandhane, Pratik Joshi, Anirudh Goyal, Matthew Tung, Andrew Brock, Hannah Sheahan, Vedant Misra, Cheng Li, Nemanja Rakićević, Mostafa Dehghani, Fangyu Liu, Sid Mittal, Junhyuk Oh, Seb Noury, Eren Sezener, Fantine Huot, Matthew Lamm, Nicola~De Cao, Charlie Chen, Sidharth Mudgal, Romina Stella, Kevin Brooks, Gautam Vasudevan, Chenxi Liu, Mainak Chain, Nivedita Melinkeri, Aaron Cohen, Venus Wang, Kristie Seymore, Sergey Zubkov, Rahul Goel, Summer Yue, Sai Krishnakumaran, Brian Albert, Nate Hurley, Motoki Sano, Anhad Mohananey, Jonah Joughin, Egor Filonov, Tomasz Kepa, Yomna Eldawy, Jiawern Lim, Rahul Rishi, Shirin Badiezadegan, Taylor Bos, Jerry Chang, Sanil Jain, Sri Gayatri~Sundara Padmanabhan, Subha Puttagunta, Kalpesh Krishna, Leslie Baker, Norbert Kalb, Vamsi Bedapudi, Adam Kurzrok, Shuntong Lei, Anthony Yu, Oren Litvin, Xiang Zhou, Zhichun Wu, Sam Sobell, Andrea Siciliano, Alan Papir, Robby Neale, Jonas Bragagnolo, Tej Toor, Tina Chen, Valentin Anklin, Feiran Wang, Richie Feng, Milad Gholami, Kevin Ling, Lijuan Liu, Jules Walter, Hamid Moghaddam, Arun Kishore, Jakub Adamek, Tyler Mercado, Jonathan Mallinson, Siddhinita Wandekar, Stephen Cagle, Eran Ofek, Guillermo Garrido, Clemens Lombriser, Maksim Mukha, Botu Sun, Hafeezul~Rahman Mohammad, Josip Matak, Yadi Qian, Vikas Peswani, Pawel Janus, Quan Yuan, Leif Schelin, Oana David, Ankur Garg, Yifan He, Oleksii Duzhyi, Anton Älgmyr, Timothée Lottaz, Qi~Li, Vikas Yadav, Luyao Xu, Alex Chinien, Rakesh Shivanna, Aleksandr Chuklin, Josie Li, Carrie Spadine, Travis Wolfe, Kareem Mohamed, Subhabrata Das, Zihang Dai, Kyle He, Daniel von Dincklage, Shyam Upadhyay, Akanksha Maurya, Luyan Chi, Sebastian Krause, Khalid Salama, Pam~G Rabinovitch, Pavan Kumar~Reddy M, Aarush Selvan, Mikhail Dektiarev, Golnaz Ghiasi, Erdem Guven, Himanshu Gupta, Boyi Liu, Deepak Sharma, Idan~Heimlich Shtacher, Shachi Paul, Oscar Akerlund, François-Xavier Aubet, Terry Huang, Chen Zhu, Eric Zhu, Elico Teixeira, Matthew Fritze, Francesco Bertolini, Liana-Eleonora Marinescu, Martin Bölle, Dominik Paulus, Khyatti Gupta, Tejasi Latkar, Max Chang, Jason Sanders, Roopa Wilson, Xuewei Wu, Yi-Xuan Tan, Lam~Nguyen Thiet, Tulsee Doshi, Sid Lall, Swaroop Mishra, Wanming Chen, Thang Luong, Seth Benjamin, Jasmine Lee, Ewa Andrejczuk, Dominik Rabiej, Vipul Ranjan, Krzysztof Styrc, Pengcheng Yin, Jon Simon, Malcolm~Rose Harriott, Mudit Bansal, Alexei Robsky, Geoff Bacon, David Greene, Daniil Mirylenka, Chen Zhou, Obaid Sarvana, Abhimanyu Goyal, Samuel Andermatt, Patrick Siegler, Ben Horn, Assaf Israel, Francesco Pongetti, Chih-Wei~"Louis" Chen, Marco Selvatici, Pedro Silva, Kathie Wang, Jackson Tolins, Kelvin Guu, Roey Yogev, Xiaochen Cai, Alessandro Agostini, Maulik Shah, Hung Nguyen, Noah~Ó Donnaile, Sébastien Pereira, Linda Friso, Adam Stambler, Adam Kurzrok, Chenkai Kuang, Yan
  Romanikhin, Mark Geller, ZJ~Yan, Kane Jang, Cheng-Chun Lee, Wojciech Fica, Eric Malmi, Qijun Tan, Dan Banica, Daniel Balle, Ryan Pham, Yanping Huang, Diana Avram, Hongzhi Shi, Jasjot Singh, Chris Hidey, Niharika Ahuja, Pranab Saxena, Dan Dooley, Srividya~Pranavi Potharaju, Eileen O'Neill, Anand Gokulchandran, Ryan Foley, Kai Zhao, Mike Dusenberry, Yuan Liu, Pulkit Mehta, Ragha Kotikalapudi, Chalence Safranek-Shrader, Andrew Goodman, Joshua Kessinger, Eran Globen, Prateek Kolhar, Chris Gorgolewski, Ali Ibrahim, Yang Song, Ali Eichenbaum, Thomas Brovelli, Sahitya Potluri, Preethi Lahoti, Cip Baetu, Ali Ghorbani, Charles Chen, Andy Crawford, Shalini Pal, Mukund Sridhar, Petru Gurita, Asier Mujika, Igor Petrovski, Pierre-Louis Cedoz, Chenmei Li, Shiyuan Chen, Niccolò~Dal Santo, Siddharth Goyal, Jitesh Punjabi, Karthik Kappaganthu, Chester Kwak, Pallavi LV, Sarmishta Velury, Himadri Choudhury, Jamie Hall, Premal Shah, Ricardo Figueira, Matt Thomas, Minjie Lu, Ting Zhou, Chintu Kumar, Thomas Jurdi, Sharat Chikkerur, Yenai Ma, Adams Yu, Soo Kwak, Victor Ähdel, Sujeevan Rajayogam, Travis Choma, Fei Liu, Aditya Barua, Colin Ji, Ji~Ho Park, Vincent Hellendoorn, Alex Bailey, Taylan Bilal, Huanjie Zhou, Mehrdad Khatir, Charles Sutton, Wojciech Rzadkowski, Fiona Macintosh, Konstantin Shagin, Paul Medina, Chen Liang, Jinjing Zhou, Pararth Shah, Yingying Bi, Attila Dankovics, Shipra Banga, Sabine Lehmann, Marissa Bredesen, Zifan Lin, John~Eric Hoffmann, Jonathan Lai, Raynald Chung, Kai Yang, Nihal Balani, Arthur Bražinskas, Andrei Sozanschi, Matthew Hayes, Héctor~Fernández Alcalde, Peter Makarov, Will Chen, Antonio Stella, Liselotte Snijders, Michael Mandl, Ante Kärrman, Paweł Nowak, Xinyi Wu, Alex Dyck, Krishnan Vaidyanathan, Raghavender R, Jessica Mallet, Mitch Rudominer, Eric Johnston, Sushil Mittal, Akhil Udathu, Janara Christensen, Vishal Verma, Zach Irving, Andreas Santucci, Gamaleldin Elsayed, Elnaz Davoodi, Marin Georgiev, Ian Tenney, Nan Hua, Geoffrey Cideron, Edouard Leurent, Mahmoud Alnahlawi, Ionut Georgescu, Nan Wei, Ivy Zheng, Dylan Scandinaro, Heinrich Jiang, Jasper Snoek, Mukund Sundararajan, Xuezhi Wang, Zack Ontiveros, Itay Karo, Jeremy Cole, Vinu Rajashekhar, Lara Tumeh, Eyal Ben-David, Rishub Jain, Jonathan Uesato, Romina Datta, Oskar Bunyan, Shimu Wu, John Zhang, Piotr Stanczyk, Ye~Zhang, David Steiner, Subhajit Naskar, Michael Azzam, Matthew Johnson, Adam Paszke, Chung-Cheng Chiu, Jaume~Sanchez Elias, Afroz Mohiuddin, Faizan Muhammad, Jin Miao, Andrew Lee, Nino Vieillard, Jane Park, Jiageng Zhang, Jeff Stanway, Drew Garmon, Abhijit Karmarkar, Zhe Dong, Jong Lee, Aviral Kumar, Luowei Zhou, Jonathan Evens, William Isaac, Geoffrey Irving, Edward Loper, Michael Fink, Isha Arkatkar, Nanxin Chen, Izhak Shafran, Ivan Petrychenko, Zhe Chen, Johnson Jia, Anselm Levskaya, Zhenkai Zhu, Peter Grabowski, Yu~Mao, Alberto Magni, Kaisheng Yao, Javier Snaider, Norman Casagrande, Evan Palmer, Paul Suganthan, Alfonso Castaño, Irene Giannoumis, Wooyeol Kim, Mikołaj Rybiński, Ashwin Sreevatsa, Jennifer Prendki, David Soergel, Adrian Goedeckemeyer, Willi Gierke, Mohsen Jafari, Meenu Gaba, Jeremy Wiesner, Diana~Gage Wright, Yawen Wei, Harsha Vashisht, Yana Kulizhskaya, Jay Hoover, Maigo Le, Lu~Li, Chimezie Iwuanyanwu, Lu~Liu, Kevin Ramirez, Andrey Khorlin, Albert Cui, Tian LIN, Marcus Wu, Ricardo Aguilar, Keith Pallo, Abhishek Chakladar, Ginger Perng, Elena~Allica Abellan, Mingyang Zhang, Ishita Dasgupta, Nate Kushman, Ivo Penchev, Alena Repina, Xihui Wu, Tom van~der Weide, Priya Ponnapalli, Caroline Kaplan, Jiri Simsa, Shuangfeng Li, Olivier Dousse, Fan Yang, Jeff Piper, Nathan Ie, Rama Pasumarthi, Nathan Lintz, Anitha Vijayakumar, Daniel Andor, Pedro Valenzuela, Minnie Lui, Cosmin Paduraru, Daiyi Peng, Katherine Lee, Shuyuan Zhang, Somer Greene, Duc~Dung Nguyen, Paula Kurylowicz, Cassidy Hardin, Lucas Dixon, Lili Janzer, Kiam Choo, Ziqiang Feng, Biao Zhang, Achintya Singhal, Dayou Du, Dan McKinnon, Natasha Antropova, Tolga Bolukbasi, Orgad Keller, David Reid, Daniel Finchelstein, Maria~Abi Raad, Remi Crocker, Peter Hawkins, Robert Dadashi, Colin Gaffney, Ken Franko, Anna Bulanova, Rémi Leblond, Shirley Chung, Harry Askham, Luis~C. Cobo, Kelvin Xu, Felix Fischer, Jun Xu, Christina Sorokin, Chris Alberti, Chu-Cheng Lin, Colin Evans, Alek Dimitriev, Hannah Forbes, Dylan Banarse, Zora Tung, Mark Omernick, Colton Bishop, Rachel Sterneck, Rohan Jain, Jiawei Xia, Ehsan Amid, Francesco Piccinno, Xingyu Wang, Praseem Banzal, Daniel~J. Mankowitz, Alex Polozov, Victoria Krakovna, Sasha Brown, MohammadHossein Bateni, Dennis Duan, Vlad Firoiu, Meghana Thotakuri, Tom Natan, Matthieu Geist, Ser tan Girgin, Hui Li, Jiayu Ye, Ofir Roval, Reiko Tojo, Michael Kwong, James Lee-Thorp, Christopher Yew, Danila Sinopalnikov, Sabela Ramos, John Mellor, Abhishek Sharma, Kathy Wu, David Miller, Nicolas Sonnerat, Denis Vnukov, Rory Greig, Jennifer Beattie, Emily Caveness, Libin Bai, Julian Eisenschlos, Alex Korchemniy, Tomy Tsai, Mimi Jasarevic, Weize Kong, Phuong Dao, Zeyu Zheng, Frederick Liu, Fan Yang, Rui Zhu, Tian~Huey Teh, Jason Sanmiya, Evgeny Gladchenko, Nejc Trdin, Daniel Toyama, Evan Rosen, Sasan Tavakkol, Linting Xue, Chen Elkind, Oliver Woodman, John Carpenter, George Papamakarios, Rupert Kemp, Sushant Kafle, Tanya Grunina, Rishika Sinha, Alice Talbert, Diane Wu, Denese Owusu-Afriyie, Cosmo Du, Chloe Thornton, Jordi Pont-Tuset, Pradyumna Narayana, Jing Li, Saaber Fatehi, John Wieting, Omar Ajmeri, Benigno Uria, Yeongil Ko, Laura Knight, Amélie Héliou, Ning Niu, Shane Gu, Chenxi Pang, Yeqing Li, Nir Levine, Ariel Stolovich, Rebeca Santamaria-Fernandez, Sonam Goenka, Wenny Yustalim, Robin Strudel, Ali Elqursh, Charlie Deck, Hyo Lee, Zonglin Li, Kyle Levin, Raphael Hoffmann, Dan Holtmann-Rice, Olivier Bachem, Sho Arora, Christy Koh, Soheil~Hassas Yeganeh, Siim Põder, Mukarram Tariq, Yanhua Sun, Lucian Ionita, Mojtaba Seyedhosseini, Pouya Tafti, Zhiyu Liu, Anmol Gulati, Jasmine Liu, Xinyu Ye, Bart Chrzaszcz, Lily Wang, Nikhil Sethi, Tianrun Li, Ben Brown, Shreya Singh, Wei Fan, Aaron Parisi, Joe Stanton, Vinod Koverkathu, Christopher~A. Choquette-Choo, Yunjie Li, TJ~Lu, Abe Ittycheriah, Prakash Shroff, Mani Varadarajan, Sanaz Bahargam, Rob Willoughby, David Gaddy, Guillaume Desjardins, Marco Cornero, Brona Robenek, Bhavishya Mittal, Ben Albrecht, Ashish Shenoy, Fedor Moiseev, Henrik Jacobsson, Alireza Ghaffarkhah, Morgane Rivière, Alanna Walton, Clément Crepy, Alicia Parrish, Zongwei Zhou, Clement Farabet, Carey Radebaugh, Praveen Srinivasan, Claudia van~der Salm, Andreas Fidjeland, Salvatore Scellato, Eri Latorre-Chimoto, Hanna Klimczak-Plucińska, David Bridson, Dario de~Cesare, Tom Hudson, Piermaria Mendolicchio, Lexi Walker, Alex Morris, Matthew Mauger, Alexey Guseynov, Alison Reid, Seth Odoom, Lucia Loher, Victor Cotruta, Madhavi Yenugula, Dominik Grewe, Anastasia Petrushkina, Tom Duerig, Antonio Sanchez, Steve Yadlowsky, Amy Shen, Amir Globerson, Lynette Webb, Sahil Dua, Dong Li, Surya Bhupatiraju, Dan Hurt, Haroon Qureshi, Ananth Agarwal, Tomer Shani, Matan Eyal, Anuj Khare, Shreyas~Rammohan Belle, Lei Wang, Chetan Tekur, Mihir~Sanjay Kale, Jinliang Wei, Ruoxin Sang, Brennan Saeta, Tyler Liechty, Yi~Sun, Yao Zhao, Stephan Lee, Pandu Nayak, Doug Fritz, Manish~Reddy Vuyyuru, John Aslanides, Nidhi Vyas, Martin Wicke, Xiao Ma, Evgenii Eltyshev, Nina Martin, Hardie Cate, James Manyika, Keyvan Amiri, Yelin Kim, Xi~Xiong, Kai Kang, Florian Luisier, Nilesh Tripuraneni, David Madras, Mandy Guo, Austin Waters, Oliver Wang, Joshua Ainslie, Jason Baldridge, Han Zhang, Garima Pruthi, Jakob Bauer, Feng Yang, Riham Mansour, Jason Gelman, Yang Xu, George Polovets, Ji~Liu, Honglong Cai, Warren Chen, XiangHai Sheng, Emily Xue, Sherjil Ozair, Christof Angermueller, Xiaowei Li, Anoop Sinha, Weiren Wang, Julia Wiesinger, Emmanouil Koukoumidis, Yuan Tian, Anand Iyer, Madhu Gurumurthy, Mark Goldenson, Parashar Shah, MK~Blake, Hongkun Yu, Anthony Urbanowicz, Jennimaria Palomaki, Chrisantha Fernando, Ken Durden, Harsh Mehta, Nikola Momchev, Elahe Rahimtoroghi, Maria Georgaki, Amit Raul, Sebastian Ruder, Morgan Redshaw, Jinhyuk Lee, Denny Zhou, Komal Jalan, Dinghua Li, Blake Hechtman, Parker Schuh, Milad Nasr, Kieran Milan, Vladimir Mikulik, Juliana Franco, Tim Green, Nam Nguyen, Joe Kelley, Aroma Mahendru, Andrea Hu, Joshua Howland, Ben Vargas, Jeffrey Hui, Kshitij Bansal, Vikram Rao, Rakesh Ghiya, Emma Wang, Ke~Ye, Jean~Michel Sarr, Melanie~Moranski Preston, Madeleine Elish, Steve Li, Aakash Kaku, Jigar Gupta, Ice Pasupat, Da-Cheng Juan, Milan Someswar, Tejvi M., Xinyun Chen, Aida Amini, Alex Fabrikant, Eric Chu, Xuanyi Dong, Amruta Muthal, Senaka Buthpitiya, Sarthak Jauhari, Nan Hua, Urvashi Khandelwal, Ayal Hitron, Jie Ren, Larissa Rinaldi, Shahar Drath, Avigail Dabush, Nan-Jiang Jiang, Harshal Godhia, Uli Sachs, Anthony Chen, Yicheng Fan, Hagai Taitelbaum, Hila Noga, Zhuyun Dai, James Wang, Chen Liang, Jenny Hamer, Chun-Sung Ferng, Chenel Elkind, Aviel Atias, Paulina Lee, Vít Listík, Mathias Carlen, Jan van~de Kerkhof, Marcin Pikus, Krunoslav Zaher, Paul Müller, Sasha Zykova, Richard Stefanec, Vitaly Gatsko, Christoph Hirnschall, Ashwin Sethi, Xingyu~Federico Xu, Chetan Ahuja, Beth Tsai, Anca Stefanoiu, Bo~Feng, Keshav Dhandhania, Manish Katyal, Akshay Gupta, Atharva Parulekar, Divya Pitta, Jing Zhao, Vivaan Bhatia, Yashodha Bhavnani, Omar Alhadlaq, Xiaolin Li, Peter Danenberg, Dennis Tu, Alex Pine, Vera Filippova, Abhipso Ghosh, Ben Limonchik, Bhargava Urala, Chaitanya~Krishna Lanka, Derik Clive, Yi~Sun, Edward Li, Hao Wu, Kevin Hongtongsak, Ianna Li, Kalind Thakkar, Kuanysh Omarov, Kushal Majmundar, Michael Alverson, Michael Kucharski, Mohak Patel, Mudit Jain, Maksim Zabelin, Paolo Pelagatti, Rohan Kohli, Saurabh Kumar, Joseph Kim, Swetha Sankar, Vineet Shah, Lakshmi Ramachandruni, Xiangkai Zeng, Ben Bariach, Laura Weidinger, Tu~Vu, Alek Andreev, Antoine He, Kevin Hui, Sheleem Kashem, Amar Subramanya, Sissie Hsiao, Demis Hassabis,
  Koray Kavukcuoglu, Adam Sadovsky, Quoc Le, Trevor Strohman, Yonghui Wu, Slav Petrov, Jeffrey Dean, and Oriol Vinyals.
\newblock Gemini: A family of highly capable multimodal models, 2024{\natexlab{a}}.
\newblock URL \url{https://arxiv.org/abs/2312.11805}.

\bibitem[Team et~al.(2024{\natexlab{b}})Team, Georgiev, Lei, Burnell, Bai, Gulati, Tanzer, Vincent, Pan, Wang, Mariooryad, Ding, Geng, Alcober, Frostig, Omernick, Walker, Paduraru, Sorokin, Tacchetti, Gaffney, Daruki, Sercinoglu, Gleicher, Love, Voigtlaender, Jain, Surita, Mohamed, Blevins, Ahn, Zhu, Kawintiranon, Firat, Gu, Zhang, Rahtz, Faruqui, Clay, Gilmer, Co-Reyes, Penchev, Zhu, Morioka, Hui, Haridasan, Campos, Mahdieh, Guo, Hassan, Kilgour, Vezer, Cheng, de~Liedekerke, Goyal, Barham, Strouse, Noury, Adler, Sundararajan, Vikram, Lepikhin, Paganini, Garcia, Yang, Valter, Trebacz, Vodrahalli, Asawaroengchai, Ring, Kalb, Soares, Brahma, Steiner, Yu, Mentzer, He, Gonzalez, Xu, Kaufman, Shafey, Oh, Hennigan, van~den Driessche, Odoom, Lucic, Roelofs, Lall, Marathe, Chan, Ontanon, He, Teplyashin, Lai, Crone, Damoc, Ho, Riedel, Lenc, Yeh, Chowdhery, Xu, Kazemi, Amid, Petrushkina, Swersky, Khodaei, Chen, Larkin, Pinto, Yan, Badia, Patil, Hansen, Orr, Arnold, Grimstad, Dai, Douglas, Sinha, Yadav, Chen, Gribovskaya, Austin, Zhao, Patel, Komarek, Austin, Borgeaud, Friso, Goyal, Caine, Cao, Chung, Lamm, Barth-Maron, Kagohara, Olszewska, Chen, Shivakumar, Agarwal, Godhia, Rajwar, Snaider, Dotiwalla, Liu, Barua, Ungureanu, Zhang, Batsaikhan, Wirth, Qin, Danihelka, Doshi, Chadwick, Chen, Jain, Le, Kar, Gurumurthy, Li, Sang, Liu, Lamprou, Munoz, Lintz, Mehta, Howard, Reynolds, Aroyo, Wang, Blanco, Cassirer, Griffith, Das, Lee, Sygnowski, Fisher, Besley, Powell, Ahmed, Paulus, Reitter, Borsos, Joshi, Pope, Hand, Selo, Jain, Sethi, Goel, Makino, May, Yang, Schalkwyk, Butterfield, Hauth, Goldin, Hawkins, Senter, Brin, Woodman, Ritter, Noland, Giang, Bolina, Lee, Blyth, Mackinnon, Reid, Sarvana, Silver, Chen, Wang, Maggiore, Chang, Attaluri, Thornton, Chiu, Bunyan, Levine, Chung, Eltyshev, Si, Lillicrap, Brady, Aggarwal, Wu, Xu, McIlroy, Badola, Sandhu, Moreira, Stokowiec, Hemsley, Li, Tudor, Shyam, Rahimtoroghi, Haykal, Sprechmann, Zhou, Mincu, Li, Addanki, Krishna, Wu, Frechette, Eyal, Dafoe, Lacey, Whang, Avrahami, Zhang, Taropa, Lin, Toyama, Rutherford, Sano, Choe, Tomala, Safranek-Shrader, Kassner, Pajarskas, Harvey, Sechrist, Fortunato, Lyu, Elsayed, Kuang, Lottes, Chu, Jia, Chen, Humphreys, Baumli, Tao, Samuel, dos Santos, Andreassen, Rakićević, Grewe, Kumar, Winkler, Caton, Brock, Dalmia, Sheahan, Barr, Miao, Natsev, Devlin, Behbahani, Prost, Sun, Myaskovsky, Pillai, Hurt, Lazaridou, Xiong, Zheng, Pardo, Li, Horgan, Stanton, Ambar, Xia, Lince, Wang, Mustafa, Webson, Lee, Anil, Wicke, Dozat, Sinha, Piqueras, Dabir, Upadhyay, Boral, Hendricks, Fry, Djolonga, Su, Walker, Labanowski, Huang, Misra, Chen, Skerry-Ryan, Singh, Rijhwani, Yu, Castro-Ros, Changpinyo, Datta, Bagri, Hrafnkelsson, Maggioni, Zheng, Sulsky, Hou, Paine, Yang, Riesa, Rogozinska, Marcus, Badawy, Zhang, Wang, Miller, Greer, Sjos, Nova, Zen, Chaabouni, Rosca, Jiang, Chen, Liu, Sainath, Krikun, Polozov, Lespiau, Newlan, Cankara, Kwak, Xu, Chen, Coenen, Meyer, Tsihlas, Ma, Gottweis, Xing, Gu, Miao, Frank, Cankara, Ganapathy, Dasgupta, Hughes-Fitt, Chen, Reid, Rong, Fan, van Amersfoort, Zhuang, Cohen, Gu, Mohananey, Ilic, Tobin, Wieting, Bortsova, Thacker, Wang, Caveness, Chiu, Sezener, Kaskasoli, Baker, Millican, Elhawaty, Aisopos, Lebsack, Byrd, Dai, Jia, Wiethoff, Davoodi, Weston, Yagati, Ahuja, Gao, Pundak, Zhang, Azzam, Sim, Caelles, Keeling, Sharma, Swing, Li, Liu, Bostock, Bansal, Nado, Anand, Lipschultz, Karmarkar, Proleev, Ittycheriah, Yeganeh, Polovets, Faust, Sun, Rrustemi, Li, Shivanna, Liu, Welty, Lebron, Baddepudi, Krause, Parisotto, Soricut, Xu, Bloxwich, Johnson, Neyshabur, Mao-Jones, Wang, Ramasesh, Abbas, Guez, Segal, Nguyen, Svensson, Hou, York, Milan, Bridgers, Gworek, Tagliasacchi, Lee-Thorp, Chang, Guseynov, Hartman, Kwong, Zhao, Kashem, Cole, Miech, Tanburn, Phuong, Pavetic, Cevey, Comanescu, Ives, Yang, Du, Li, Zhang, Iinuma, Hu, Roy, Bijwadia, Zhu, Martins, Saputro, Gergely, Zheng, Jia, Antonoglou, Sadovsky, Gu, Bi, Andreev, Samangooei, Khan, Kocisky, Filos, Kumar, Bishop, Yu, Hodkinson, Mittal, Shah, Moufarek, Cheng, Bloniarz, Lee, Pejman, Michel, Spencer, Feinberg, Xiong, Savinov, Smith, Shakeri, Tran, Chesus, Bohnet, Tucker, von Glehn, Muir, Mao, Kazawa, Slone, Soparkar, Shrivastava, Cobon-Kerr, Sharman, Pavagadhi, Araya, Misiunas, Ghelani, Laskin, Barker, Li, Briukhov, Houlsby, Glaese, Lakshminarayanan, Schucher, Tang, Collins, Lim, Feng, Recasens, Lai, Magni, Cao, Siddhant, Ashwood, Orbay, Dehghani, Brennan, He, Xu, Gao, Saroufim, Molloy, Wu, Arnold, Chang, Schrittwieser, Buchatskaya, Radpour, Polacek, Giordano, Bapna, Tokumine, Hellendoorn, Sottiaux, Cogan, Severyn, Saleh, Thakoor, Shefey, Qiao, Gaba, yiin Chang, Swanson, Zhang, Lee, Rubenstein, Song, Kwiatkowski, Koop, Kannan, Kao, Schuh, Stjerngren, Ghiasi, Gibson, Vilnis, Yuan, Ferreira, Kamath, Klimenko, Franko, Xiao, Bhattacharya, Patel, Wang, Morris, Strudel, Sharma, Choy, Hashemi, Landon, Finkelstein, Jhakra, Frye, Barnes, Mauger, Daun, Baatarsukh, Tung, Farhan, Michalewski, Viola, de~Chaumont~Quitry, Lan, Hudson, Wang, Fischer, Zheng, White, Dragan, baptiste Alayrac, Ni, Pritzel, Iwanicki, Isard, Bulanova, Zilka, Dyer, Sachan, Srinivasan, Muckenhirn, Cai, Mandhane, Tariq, Rae, Wang, Ayoub, FitzGerald, Zhao, Han, Alberti, Garrette, Krishnakumar, Gimenez, Levskaya, Sohn, Matak, Iturrate, Chang, Xiang, Cao, Ranka, Brown, Hutter, Mirrokni, Chen, Yao, Egyed, Galilee, Liechty, Kallakuri, Palmer, Ghemawat, Liu, Tao, Thornton, Green, Jasarevic, Lin, Cotruta, Tan, Fiedel, Yu, Chi, Neitz, Heitkaemper, Sinha, Zhou, Sun, Kaed, Hulse, Mishra, Georgaki, Kudugunta, Farabet, Shafran, Vlasic, Tsitsulin, Ananthanarayanan, Carin, Su, Sun, V, Carvajal, Broder, Comsa, Repina, Wong, Chen, Hawkins, Filonov, Loher, Hirnschall, Wang, Ye, Burns, Cate, Wright, Piccinini, Zhang, Lin, Gog, Kulizhskaya, Sreevatsa, Song, Cobo, Iyer, Tekur, Garrido, Xiao, Kemp, Zheng, Li, Agarwal, Ngani, Goshvadi, Santamaria-Fernandez, Fica, Chen, Gorgolewski, Sun, Garg, Ye, Eslami, Hua, Simon, Joshi, Kim, Tenney, Potluri, Thiet, Yuan, Luisier, Chronopoulou, Scellato, Srinivasan, Chen, Koverkathu, Dalibard, Xu, Saeta, Anderson, Sellam, Fernando, Huot, Jung, Varadarajan, Quinn, Raul, Le, Habalov, Clark, Jalan, Bullard, Singhal, Luong, Wang, Rajayogam, Eisenschlos, Jia, Finchelstein, Yakubovich, Balle, Fink, Agarwal, Li, Dvijotham, Pal, Kang, Konzelmann, Beattie, Dousse, Wu, Crocker, Elkind, Jonnalagadda, Lee, Holtmann-Rice, Kallarackal, Liu, Vnukov, Vats, Invernizzi, Jafari, Zhou, Taylor, Prendki, Wu, Eccles, Liu, Kopparapu, Beaufays, Angermueller, Marzoca, Sarcar, Dib, Stanway, Perbet, Trdin, Sterneck, Khorlin, Li, Wu, Goenka, Madras, Goldshtein, Gierke, Zhou, Liu, Liang, White, Li, Singh, Bahargam, Epstein, Basu, Lao, Ozturel, Crous, Zhai, Lu, Tung, Gaur, Walton, Dixon, Zhang, Globerson, Uy, Bolt, Wiles, Nasr, Shumailov, Selvi, Piccinno, Aguilar, McCarthy, Khalman, Shukla, Galic, Carpenter, Villela, Zhang, Richardson, Martens, Bosnjak, Belle, Seibert, Alnahlawi, McWilliams, Singh, Louis, Ding, Popovici, Simicich, Knight, Mehta, Gupta, Shi, Fatehi, Mitrovic, Grills, Pagadora, Petrova, Eisenbud, Zhang, Yates, Mittal, Tripuraneni, Assael, Brovelli, Jain, Velimirovic, Akbulut, Mu, Macherey, Kumar, Xu, Qureshi, Comanici, Wiesner, Gong, Ruddock, Bauer, Felt, GP, Arnab, Zelle, Rothfuss, Rosgen, Shenoy, Seybold, Li, Mudigonda, Erdogan, Xia, Simsa, Michi, Yao, Yew, Kan, Caswell, Radebaugh, Elisseeff, Valenzuela, McKinney, Paterson, Cui, Latorre-Chimoto, Kim, Zeng, Durden, Ponnapalli, Sosea, Choquette-Choo, Manyika, Robenek, Vashisht, Pereira, Lam, Velic, Owusu-Afriyie, Lee, Bolukbasi, Parrish, Lu, Park, Venkatraman, Talbert, Rosique, Cheng, Sozanschi, Paszke, Kumar, Austin, Li, Salama, Kim, Dukkipati, Baryshnikov, Kaplanis, Sheng, Chervonyi, Unlu, de~Las~Casas, Askham, Tunyasuvunakool, Gimeno, Poder, Kwak, Miecnikowski, Mirrokni, Dimitriev, Parisi, Liu, Tsai, Shevlane, Kouridi, Garmon, Goedeckemeyer, Brown, Vijayakumar, Elqursh, Jazayeri, Huang, Carthy, Hoover, Kim, Kumar, Chen, Biles, Bingham, Rosen, Wang, Tan, Engel, Pongetti, de~Cesare, Hwang, Yu, Pullman, Narayanan, Levin, Gopal, Li, Aharoni, Trinh, Lo, Casagrande, Vij, Matthey, Ramadhana, Matthews, Carey, Johnson, Goranova, Shah, Ashraf, Dasgupta, Larsen, Wang, Vuyyuru, Jiang, Ijazi, Osawa, Smith, Boppana, Bilal, Koizumi, Xu, Altun, Shabat, Bariach, Korchemniy, Choo, Ronneberger, Iwuanyanwu, Zhao, Soergel, Hsieh, Cai, Iqbal, Sundermeyer, Chen, Bursztein, Malaviya, Biadsy, Shroff, Dhillon, Latkar, Dyer, Forbes, Nicosia, Nikolaev, Greene, Georgiev, Wang, Martin, Sedghi, Zhang, Banzal, Fritz, Rao, Wang, Zhang, Patraucean, Du, Mordatch, Jurin, Liu, Dubey, Mohan, Nowakowski, Ion, Wei, Tojo, Raad, Hudson, Keshava, Agrawal, Ramirez, Wu, Nguyen, Liu, Sewak, Petrini, Choi, Philips, Wang, Bica, Garg, Wilkiewicz, Agrawal, Li, Guo, Xue, Shaik, Leach, Khan, Wiesinger, Jerome, Chakladar, Wang, Ornduff, Abu, Ghaffarkhah, Wainwright, Cortes, Liu, Maynez, Terzis, Samangouei, Mansour, Kepa, Aubet, Algymr, Banica, Weisz, Orban, Senges, Andrejczuk, Geller, Santo, Anklin, Merey, Baeuml, Strohman, Bai, Petrov, Wu, Hassabis, Kavukcuoglu, Dean, and Vinyals]{geminiteam2024gemini15unlockingmultimodal}
Gemini Team, Petko Georgiev, Ving~Ian Lei, Ryan Burnell, Libin Bai, Anmol Gulati, Garrett Tanzer, Damien Vincent, Zhufeng Pan, Shibo Wang, Soroosh Mariooryad, Yifan Ding, Xinyang Geng, Fred Alcober, Roy Frostig, Mark Omernick, Lexi Walker, Cosmin Paduraru, Christina Sorokin, Andrea Tacchetti, Colin Gaffney, Samira Daruki, Olcan Sercinoglu, Zach Gleicher, Juliette Love, Paul Voigtlaender, Rohan Jain, Gabriela Surita, Kareem Mohamed, Rory Blevins, Junwhan Ahn, Tao Zhu, Kornraphop Kawintiranon, Orhan Firat, Yiming Gu, Yujing Zhang, Matthew Rahtz, Manaal Faruqui, Natalie Clay, Justin Gilmer, JD~Co-Reyes, Ivo Penchev, Rui Zhu, Nobuyuki Morioka, Kevin Hui, Krishna Haridasan, Victor Campos, Mahdis Mahdieh, Mandy Guo, Samer Hassan, Kevin Kilgour, Arpi Vezer, Heng-Tze Cheng, Raoul de~Liedekerke, Siddharth Goyal, Paul Barham, DJ~Strouse, Seb Noury, Jonas Adler, Mukund Sundararajan, Sharad Vikram, Dmitry Lepikhin, Michela Paganini, Xavier Garcia, Fan Yang, Dasha Valter, Maja Trebacz, Kiran Vodrahalli, Chulayuth Asawaroengchai, Roman Ring, Norbert Kalb, Livio~Baldini Soares, Siddhartha Brahma, David Steiner, Tianhe Yu, Fabian Mentzer, Antoine He, Lucas Gonzalez, Bibo Xu, Raphael~Lopez Kaufman, Laurent~El Shafey, Junhyuk Oh, Tom Hennigan, George van~den Driessche, Seth Odoom, Mario Lucic, Becca Roelofs, Sid Lall, Amit Marathe, Betty Chan, Santiago Ontanon, Luheng He, Denis Teplyashin, Jonathan Lai, Phil Crone, Bogdan Damoc, Lewis Ho, Sebastian Riedel, Karel Lenc, Chih-Kuan Yeh, Aakanksha Chowdhery, Yang Xu, Mehran Kazemi, Ehsan Amid, Anastasia Petrushkina, Kevin Swersky, Ali Khodaei, Gowoon Chen, Chris Larkin, Mario Pinto, Geng Yan, Adria~Puigdomenech Badia, Piyush Patil, Steven Hansen, Dave Orr, Sebastien M.~R. Arnold, Jordan Grimstad, Andrew Dai, Sholto Douglas, Rishika Sinha, Vikas Yadav, Xi~Chen, Elena Gribovskaya, Jacob Austin, Jeffrey Zhao, Kaushal Patel, Paul Komarek, Sophia Austin, Sebastian Borgeaud, Linda Friso, Abhimanyu Goyal, Ben Caine, Kris Cao, Da-Woon Chung, Matthew Lamm, Gabe Barth-Maron, Thais Kagohara, Kate Olszewska, Mia Chen, Kaushik Shivakumar, Rishabh Agarwal, Harshal Godhia, Ravi Rajwar, Javier Snaider, Xerxes Dotiwalla, Yuan Liu, Aditya Barua, Victor Ungureanu, Yuan Zhang, Bat-Orgil Batsaikhan, Mateo Wirth, James Qin, Ivo Danihelka, Tulsee Doshi, Martin Chadwick, Jilin Chen, Sanil Jain, Quoc Le, Arjun Kar, Madhu Gurumurthy, Cheng Li, Ruoxin Sang, Fangyu Liu, Lampros Lamprou, Rich Munoz, Nathan Lintz, Harsh Mehta, Heidi Howard, Malcolm Reynolds, Lora Aroyo, Quan Wang, Lorenzo Blanco, Albin Cassirer, Jordan Griffith, Dipanjan Das, Stephan Lee, Jakub Sygnowski, Zach Fisher, James Besley, Richard Powell, Zafarali Ahmed, Dominik Paulus, David Reitter, Zalan Borsos, Rishabh Joshi, Aedan Pope, Steven Hand, Vittorio Selo, Vihan Jain, Nikhil Sethi, Megha Goel, Takaki Makino, Rhys May, Zhen Yang, Johan Schalkwyk, Christina Butterfield, Anja Hauth, Alex Goldin, Will Hawkins, Evan Senter, Sergey Brin, Oliver Woodman, Marvin Ritter, Eric Noland, Minh Giang, Vijay Bolina, Lisa Lee, Tim Blyth, Ian Mackinnon, Machel Reid, Obaid Sarvana, David Silver, Alexander Chen, Lily Wang, Loren Maggiore, Oscar Chang, Nithya Attaluri, Gregory Thornton, Chung-Cheng Chiu, Oskar Bunyan, Nir Levine, Timothy Chung, Evgenii Eltyshev, Xiance Si, Timothy Lillicrap, Demetra Brady, Vaibhav Aggarwal, Boxi Wu, Yuanzhong Xu, Ross McIlroy, Kartikeya Badola, Paramjit Sandhu, Erica Moreira, Wojciech Stokowiec, Ross Hemsley, Dong Li, Alex Tudor, Pranav Shyam, Elahe Rahimtoroghi, Salem Haykal, Pablo Sprechmann, Xiang Zhou, Diana Mincu, Yujia Li, Ravi Addanki, Kalpesh Krishna, Xiao Wu, Alexandre Frechette, Matan Eyal, Allan Dafoe, Dave Lacey, Jay Whang, Thi Avrahami, Ye~Zhang, Emanuel Taropa, Hanzhao Lin, Daniel Toyama, Eliza Rutherford, Motoki Sano, HyunJeong Choe, Alex Tomala, Chalence Safranek-Shrader, Nora Kassner, Mantas Pajarskas, Matt Harvey, Sean Sechrist, Meire Fortunato, Christina Lyu, Gamaleldin Elsayed, Chenkai Kuang, James Lottes, Eric Chu, Chao Jia, Chih-Wei Chen, Peter Humphreys, Kate Baumli, Connie Tao, Rajkumar Samuel, Cicero~Nogueira dos Santos, Anders Andreassen, Nemanja Rakićević, Dominik Grewe, Aviral Kumar, Stephanie Winkler, Jonathan Caton, Andrew Brock, Sid Dalmia, Hannah Sheahan, Iain Barr, Yingjie Miao, Paul Natsev, Jacob Devlin, Feryal Behbahani, Flavien Prost, Yanhua Sun, Artiom Myaskovsky, Thanumalayan~Sankaranarayana Pillai, Dan Hurt, Angeliki Lazaridou, Xi~Xiong, Ce~Zheng, Fabio Pardo, Xiaowei Li, Dan Horgan, Joe Stanton, Moran Ambar, Fei Xia, Alejandro Lince, Mingqiu Wang, Basil Mustafa, Albert Webson, Hyo Lee, Rohan Anil, Martin Wicke, Timothy Dozat, Abhishek Sinha, Enrique Piqueras, Elahe Dabir, Shyam Upadhyay, Anudhyan Boral, Lisa~Anne Hendricks, Corey Fry, Josip Djolonga, Yi~Su, Jake Walker, Jane Labanowski, Ronny Huang, Vedant Misra, Jeremy Chen, RJ~Skerry-Ryan, Avi Singh, Shruti Rijhwani, Dian Yu, Alex Castro-Ros, Beer Changpinyo, Romina Datta, Sumit Bagri, Arnar~Mar Hrafnkelsson, Marcello Maggioni, Daniel Zheng, Yury Sulsky, Shaobo Hou, Tom~Le Paine, Antoine Yang, Jason Riesa, Dominika Rogozinska, Dror Marcus, Dalia~El Badawy, Qiao Zhang, Luyu Wang, Helen Miller, Jeremy Greer, Lars~Lowe Sjos, Azade Nova, Heiga Zen, Rahma Chaabouni, Mihaela Rosca, Jiepu Jiang, Charlie Chen, Ruibo Liu, Tara Sainath, Maxim Krikun, Alex Polozov, Jean-Baptiste Lespiau, Josh Newlan, Zeyncep Cankara, Soo Kwak, Yunhan Xu, Phil Chen, Andy Coenen, Clemens Meyer, Katerina Tsihlas, Ada Ma, Juraj Gottweis, Jinwei Xing, Chenjie Gu, Jin Miao, Christian Frank, Zeynep Cankara, Sanjay Ganapathy, Ishita Dasgupta, Steph Hughes-Fitt, Heng Chen, David Reid, Keran Rong, Hongmin Fan, Joost van Amersfoort, Vincent Zhuang, Aaron Cohen, Shixiang~Shane Gu, Anhad Mohananey, Anastasija Ilic, Taylor Tobin, John Wieting, Anna Bortsova, Phoebe Thacker, Emma Wang, Emily Caveness, Justin Chiu, Eren Sezener, Alex Kaskasoli, Steven Baker, Katie Millican, Mohamed Elhawaty, Kostas Aisopos, Carl Lebsack, Nathan Byrd, Hanjun Dai, Wenhao Jia, Matthew Wiethoff, Elnaz Davoodi, Albert Weston, Lakshman Yagati, Arun Ahuja, Isabel Gao, Golan Pundak, Susan Zhang, Michael Azzam, Khe~Chai Sim, Sergi Caelles, James Keeling, Abhanshu Sharma, Andy Swing, YaGuang Li, Chenxi Liu, Carrie~Grimes Bostock, Yamini Bansal, Zachary Nado, Ankesh Anand, Josh Lipschultz, Abhijit Karmarkar, Lev Proleev, Abe Ittycheriah, Soheil~Hassas Yeganeh, George Polovets, Aleksandra Faust, Jiao Sun, Alban Rrustemi, Pen Li, Rakesh Shivanna, Jeremiah Liu, Chris Welty, Federico Lebron, Anirudh Baddepudi, Sebastian Krause, Emilio Parisotto, Radu Soricut, Zheng Xu, Dawn Bloxwich, Melvin Johnson, Behnam Neyshabur, Justin Mao-Jones, Renshen Wang, Vinay Ramasesh, Zaheer Abbas, Arthur Guez, Constant Segal, Duc~Dung Nguyen, James Svensson, Le~Hou, Sarah York, Kieran Milan, Sophie Bridgers, Wiktor Gworek, Marco Tagliasacchi, James Lee-Thorp, Michael Chang, Alexey Guseynov, Ale~Jakse Hartman, Michael Kwong, Ruizhe Zhao, Sheleem Kashem, Elizabeth Cole, Antoine Miech, Richard Tanburn, Mary Phuong, Filip Pavetic, Sebastien Cevey, Ramona Comanescu, Richard Ives, Sherry Yang, Cosmo Du, Bo~Li, Zizhao Zhang, Mariko Iinuma, Clara~Huiyi Hu, Aurko Roy, Shaan Bijwadia, Zhenkai Zhu, Danilo Martins, Rachel Saputro, Anita Gergely, Steven Zheng, Dawei Jia, Ioannis Antonoglou, Adam Sadovsky, Shane Gu, Yingying Bi, Alek Andreev, Sina Samangooei, Mina Khan, Tomas Kocisky, Angelos Filos, Chintu Kumar, Colton Bishop, Adams Yu, Sarah Hodkinson, Sid Mittal, Premal Shah, Alexandre Moufarek, Yong Cheng, Adam Bloniarz, Jaehoon Lee, Pedram Pejman, Paul Michel, Stephen Spencer, Vladimir Feinberg, Xuehan Xiong, Nikolay Savinov, Charlotte Smith, Siamak Shakeri, Dustin Tran, Mary Chesus, Bernd Bohnet, George Tucker, Tamara von Glehn, Carrie Muir, Yiran Mao, Hideto Kazawa, Ambrose Slone, Kedar Soparkar, Disha Shrivastava, James Cobon-Kerr, Michael Sharman, Jay Pavagadhi, Carlos Araya, Karolis Misiunas, Nimesh Ghelani, Michael Laskin, David Barker, Qiujia Li, Anton Briukhov, Neil Houlsby, Mia Glaese, Balaji Lakshminarayanan, Nathan Schucher, Yunhao Tang, Eli Collins, Hyeontaek Lim, Fangxiaoyu Feng, Adria Recasens, Guangda Lai, Alberto Magni, Nicola~De Cao, Aditya Siddhant, Zoe Ashwood, Jordi Orbay, Mostafa Dehghani, Jenny Brennan, Yifan He, Kelvin Xu, Yang Gao, Carl Saroufim, James Molloy, Xinyi Wu, Seb Arnold, Solomon Chang, Julian Schrittwieser, Elena Buchatskaya, Soroush Radpour, Martin Polacek, Skye Giordano, Ankur Bapna, Simon Tokumine, Vincent Hellendoorn, Thibault Sottiaux, Sarah Cogan, Aliaksei Severyn, Mohammad Saleh, Shantanu Thakoor, Laurent Shefey, Siyuan Qiao, Meenu Gaba, Shuo yiin Chang, Craig Swanson, Biao Zhang, Benjamin Lee, Paul~Kishan Rubenstein, Gan Song, Tom Kwiatkowski, Anna Koop, Ajay Kannan, David Kao, Parker Schuh, Axel Stjerngren, Golnaz Ghiasi, Gena Gibson, Luke Vilnis, Ye~Yuan, Felipe~Tiengo Ferreira, Aishwarya Kamath, Ted Klimenko, Ken Franko, Kefan Xiao, Indro Bhattacharya, Miteyan Patel, Rui Wang, Alex Morris, Robin Strudel, Vivek Sharma, Peter Choy, Sayed~Hadi Hashemi, Jessica Landon, Mara Finkelstein, Priya Jhakra, Justin Frye, Megan Barnes, Matthew Mauger, Dennis Daun, Khuslen Baatarsukh, Matthew Tung, Wael Farhan, Henryk Michalewski, Fabio Viola, Felix de~Chaumont~Quitry, Charline~Le Lan, Tom Hudson, Qingze Wang, Felix Fischer, Ivy Zheng, Elspeth White, Anca Dragan, Jean baptiste Alayrac, Eric Ni, Alexander Pritzel, Adam Iwanicki, Michael Isard, Anna Bulanova, Lukas Zilka, Ethan Dyer, Devendra Sachan, Srivatsan Srinivasan, Hannah Muckenhirn, Honglong Cai, Amol Mandhane, Mukarram Tariq, Jack~W. Rae, Gary Wang, Kareem Ayoub, Nicholas FitzGerald, Yao Zhao, Woohyun Han, Chris Alberti, Dan Garrette, Kashyap Krishnakumar, Mai Gimenez, Anselm Levskaya, Daniel Sohn, Josip Matak, Inaki Iturrate, Michael~B. Chang, Jackie Xiang, Yuan Cao, Nishant Ranka, Geoff Brown, Adrian Hutter, Vahab Mirrokni, Nanxin Chen, Kaisheng Yao, Zoltan Egyed, Francois Galilee, Tyler Liechty, Praveen Kallakuri, Evan Palmer, Sanjay Ghemawat,
  Jasmine Liu, David Tao, Chloe Thornton, Tim Green, Mimi Jasarevic, Sharon Lin, Victor Cotruta, Yi-Xuan Tan, Noah Fiedel, Hongkun Yu, Ed~Chi, Alexander Neitz, Jens Heitkaemper, Anu Sinha, Denny Zhou, Yi~Sun, Charbel Kaed, Brice Hulse, Swaroop Mishra, Maria Georgaki, Sneha Kudugunta, Clement Farabet, Izhak Shafran, Daniel Vlasic, Anton Tsitsulin, Rajagopal Ananthanarayanan, Alen Carin, Guolong Su, Pei Sun, Shashank V, Gabriel Carvajal, Josef Broder, Iulia Comsa, Alena Repina, William Wong, Warren~Weilun Chen, Peter Hawkins, Egor Filonov, Lucia Loher, Christoph Hirnschall, Weiyi Wang, Jingchen Ye, Andrea Burns, Hardie Cate, Diana~Gage Wright, Federico Piccinini, Lei Zhang, Chu-Cheng Lin, Ionel Gog, Yana Kulizhskaya, Ashwin Sreevatsa, Shuang Song, Luis~C. Cobo, Anand Iyer, Chetan Tekur, Guillermo Garrido, Zhuyun Xiao, Rupert Kemp, Huaixiu~Steven Zheng, Hui Li, Ananth Agarwal, Christel Ngani, Kati Goshvadi, Rebeca Santamaria-Fernandez, Wojciech Fica, Xinyun Chen, Chris Gorgolewski, Sean Sun, Roopal Garg, Xinyu Ye, S.~M.~Ali Eslami, Nan Hua, Jon Simon, Pratik Joshi, Yelin Kim, Ian Tenney, Sahitya Potluri, Lam~Nguyen Thiet, Quan Yuan, Florian Luisier, Alexandra Chronopoulou, Salvatore Scellato, Praveen Srinivasan, Minmin Chen, Vinod Koverkathu, Valentin Dalibard, Yaming Xu, Brennan Saeta, Keith Anderson, Thibault Sellam, Nick Fernando, Fantine Huot, Junehyuk Jung, Mani Varadarajan, Michael Quinn, Amit Raul, Maigo Le, Ruslan Habalov, Jon Clark, Komal Jalan, Kalesha Bullard, Achintya Singhal, Thang Luong, Boyu Wang, Sujeevan Rajayogam, Julian Eisenschlos, Johnson Jia, Daniel Finchelstein, Alex Yakubovich, Daniel Balle, Michael Fink, Sameer Agarwal, Jing Li, Dj~Dvijotham, Shalini Pal, Kai Kang, Jaclyn Konzelmann, Jennifer Beattie, Olivier Dousse, Diane Wu, Remi Crocker, Chen Elkind, Siddhartha~Reddy Jonnalagadda, Jong Lee, Dan Holtmann-Rice, Krystal Kallarackal, Rosanne Liu, Denis Vnukov, Neera Vats, Luca Invernizzi, Mohsen Jafari, Huanjie Zhou, Lilly Taylor, Jennifer Prendki, Marcus Wu, Tom Eccles, Tianqi Liu, Kavya Kopparapu, Francoise Beaufays, Christof Angermueller, Andreea Marzoca, Shourya Sarcar, Hilal Dib, Jeff Stanway, Frank Perbet, Nejc Trdin, Rachel Sterneck, Andrey Khorlin, Dinghua Li, Xihui Wu, Sonam Goenka, David Madras, Sasha Goldshtein, Willi Gierke, Tong Zhou, Yaxin Liu, Yannie Liang, Anais White, Yunjie Li, Shreya Singh, Sanaz Bahargam, Mark Epstein, Sujoy Basu, Li~Lao, Adnan Ozturel, Carl Crous, Alex Zhai, Han Lu, Zora Tung, Neeraj Gaur, Alanna Walton, Lucas Dixon, Ming Zhang, Amir Globerson, Grant Uy, Andrew Bolt, Olivia Wiles, Milad Nasr, Ilia Shumailov, Marco Selvi, Francesco Piccinno, Ricardo Aguilar, Sara McCarthy, Misha Khalman, Mrinal Shukla, Vlado Galic, John Carpenter, Kevin Villela, Haibin Zhang, Harry Richardson, James Martens, Matko Bosnjak, Shreyas~Rammohan Belle, Jeff Seibert, Mahmoud Alnahlawi, Brian McWilliams, Sankalp Singh, Annie Louis, Wen Ding, Dan Popovici, Lenin Simicich, Laura Knight, Pulkit Mehta, Nishesh Gupta, Chongyang Shi, Saaber Fatehi, Jovana Mitrovic, Alex Grills, Joseph Pagadora, Dessie Petrova, Danielle Eisenbud, Zhishuai Zhang, Damion Yates, Bhavishya Mittal, Nilesh Tripuraneni, Yannis Assael, Thomas Brovelli, Prateek Jain, Mihajlo Velimirovic, Canfer Akbulut, Jiaqi Mu, Wolfgang Macherey, Ravin Kumar, Jun Xu, Haroon Qureshi, Gheorghe Comanici, Jeremy Wiesner, Zhitao Gong, Anton Ruddock, Matthias Bauer, Nick Felt, Anirudh GP, Anurag Arnab, Dustin Zelle, Jonas Rothfuss, Bill Rosgen, Ashish Shenoy, Bryan Seybold, Xinjian Li, Jayaram Mudigonda, Goker Erdogan, Jiawei Xia, Jiri Simsa, Andrea Michi, Yi~Yao, Christopher Yew, Steven Kan, Isaac Caswell, Carey Radebaugh, Andre Elisseeff, Pedro Valenzuela, Kay McKinney, Kim Paterson, Albert Cui, Eri Latorre-Chimoto, Solomon Kim, William Zeng, Ken Durden, Priya Ponnapalli, Tiberiu Sosea, Christopher~A. Choquette-Choo, James Manyika, Brona Robenek, Harsha Vashisht, Sebastien Pereira, Hoi Lam, Marko Velic, Denese Owusu-Afriyie, Katherine Lee, Tolga Bolukbasi, Alicia Parrish, Shawn Lu, Jane Park, Balaji Venkatraman, Alice Talbert, Lambert Rosique, Yuchung Cheng, Andrei Sozanschi, Adam Paszke, Praveen Kumar, Jessica Austin, Lu~Li, Khalid Salama, Wooyeol Kim, Nandita Dukkipati, Anthony Baryshnikov, Christos Kaplanis, XiangHai Sheng, Yuri Chervonyi, Caglar Unlu, Diego de~Las~Casas, Harry Askham, Kathryn Tunyasuvunakool, Felix Gimeno, Siim Poder, Chester Kwak, Matt Miecnikowski, Vahab Mirrokni, Alek Dimitriev, Aaron Parisi, Dangyi Liu, Tomy Tsai, Toby Shevlane, Christina Kouridi, Drew Garmon, Adrian Goedeckemeyer, Adam~R. Brown, Anitha Vijayakumar, Ali Elqursh, Sadegh Jazayeri, Jin Huang, Sara~Mc Carthy, Jay Hoover, Lucy Kim, Sandeep Kumar, Wei Chen, Courtney Biles, Garrett Bingham, Evan Rosen, Lisa Wang, Qijun Tan, David Engel, Francesco Pongetti, Dario de~Cesare, Dongseong Hwang, Lily Yu, Jennifer Pullman, Srini Narayanan, Kyle Levin, Siddharth Gopal, Megan Li, Asaf Aharoni, Trieu Trinh, Jessica Lo, Norman Casagrande, Roopali Vij, Loic Matthey, Bramandia Ramadhana, Austin Matthews, CJ~Carey, Matthew Johnson, Kremena Goranova, Rohin Shah, Shereen Ashraf, Kingshuk Dasgupta, Rasmus Larsen, Yicheng Wang, Manish~Reddy Vuyyuru, Chong Jiang, Joana Ijazi, Kazuki Osawa, Celine Smith, Ramya~Sree Boppana, Taylan Bilal, Yuma Koizumi, Ying Xu, Yasemin Altun, Nir Shabat, Ben Bariach, Alex Korchemniy, Kiam Choo, Olaf Ronneberger, Chimezie Iwuanyanwu, Shubin Zhao, David Soergel, Cho-Jui Hsieh, Irene Cai, Shariq Iqbal, Martin Sundermeyer, Zhe Chen, Elie Bursztein, Chaitanya Malaviya, Fadi Biadsy, Prakash Shroff, Inderjit Dhillon, Tejasi Latkar, Chris Dyer, Hannah Forbes, Massimo Nicosia, Vitaly Nikolaev, Somer Greene, Marin Georgiev, Pidong Wang, Nina Martin, Hanie Sedghi, John Zhang, Praseem Banzal, Doug Fritz, Vikram Rao, Xuezhi Wang, Jiageng Zhang, Viorica Patraucean, Dayou Du, Igor Mordatch, Ivan Jurin, Lewis Liu, Ayush Dubey, Abhi Mohan, Janek Nowakowski, Vlad-Doru Ion, Nan Wei, Reiko Tojo, Maria~Abi Raad, Drew~A. Hudson, Vaishakh Keshava, Shubham Agrawal, Kevin Ramirez, Zhichun Wu, Hoang Nguyen, Ji~Liu, Madhavi Sewak, Bryce Petrini, DongHyun Choi, Ivan Philips, Ziyue Wang, Ioana Bica, Ankush Garg, Jarek Wilkiewicz, Priyanka Agrawal, Xiaowei Li, Danhao Guo, Emily Xue, Naseer Shaik, Andrew Leach, Sadh~MNM Khan, Julia Wiesinger, Sammy Jerome, Abhishek Chakladar, Alek~Wenjiao Wang, Tina Ornduff, Folake Abu, Alireza Ghaffarkhah, Marcus Wainwright, Mario Cortes, Frederick Liu, Joshua Maynez, Andreas Terzis, Pouya Samangouei, Riham Mansour, Tomasz Kepa, François-Xavier Aubet, Anton Algymr, Dan Banica, Agoston Weisz, Andras Orban, Alexandre Senges, Ewa Andrejczuk, Mark Geller, Niccolo~Dal Santo, Valentin Anklin, Majd~Al Merey, Martin Baeuml, Trevor Strohman, Junwen Bai, Slav Petrov, Yonghui Wu, Demis Hassabis, Koray Kavukcuoglu, Jeffrey Dean, and Oriol Vinyals.
\newblock Gemini 1.5: Unlocking multimodal understanding across millions of tokens of context, 2024{\natexlab{b}}.
\newblock URL \url{https://arxiv.org/abs/2403.05530}.

\bibitem[Team et~al.(2024{\natexlab{c}})Team, Lenz, Arazi, Bergman, Manevich, Peleg, Aviram, Almagor, Fridman, Padnos, Gissin, Jannai, Muhlgay, Zimberg, Gerber, Dolev, Krakovsky, Safahi, Schwartz, Cohen, Shachaf, Rozenblum, Bata, Blass, Magar, Dalmedigos, Osin, Fadlon, Rozman, Danos, Gokhman, Zusman, Gidron, Ratner, Gat, Rozen, Fried, Leshno, Antverg, Abend, Lieber, Dagan, Cohavi, Alon, Belson, Cohen, Gilad, Glozman, Lev, Meirom, Delbari, Ness, Asida, Gal, Braude, Pumerantz, Cohen, Belinkov, Globerson, Levy, and Shoham]{jambateam2024jamba15hybridtransformermambamodels}
Jamba Team, Barak Lenz, Alan Arazi, Amir Bergman, Avshalom Manevich, Barak Peleg, Ben Aviram, Chen Almagor, Clara Fridman, Dan Padnos, Daniel Gissin, Daniel Jannai, Dor Muhlgay, Dor Zimberg, Edden~M Gerber, Elad Dolev, Eran Krakovsky, Erez Safahi, Erez Schwartz, Gal Cohen, Gal Shachaf, Haim Rozenblum, Hofit Bata, Ido Blass, Inbal Magar, Itay Dalmedigos, Jhonathan Osin, Julie Fadlon, Maria Rozman, Matan Danos, Michael Gokhman, Mor Zusman, Naama Gidron, Nir Ratner, Noam Gat, Noam Rozen, Oded Fried, Ohad Leshno, Omer Antverg, Omri Abend, Opher Lieber, Or~Dagan, Orit Cohavi, Raz Alon, Ro'i Belson, Roi Cohen, Rom Gilad, Roman Glozman, Shahar Lev, Shaked Meirom, Tal Delbari, Tal Ness, Tomer Asida, Tom~Ben Gal, Tom Braude, Uriya Pumerantz, Yehoshua Cohen, Yonatan Belinkov, Yuval Globerson, Yuval~Peleg Levy, and Yoav Shoham.
\newblock Jamba-1.5: Hybrid transformer-mamba models at scale, 2024{\natexlab{c}}.
\newblock URL \url{https://arxiv.org/abs/2408.12570}.

\bibitem[Touvron et~al.(2023)Touvron, Martin, Stone, Albert, Almahairi, Babaei, Bashlykov, Batra, Bhargava, Bhosale, Bikel, Blecher, Ferrer, Chen, Cucurull, Esiobu, Fernandes, Fu, Fu, Fuller, Gao, Goswami, Goyal, Hartshorn, Hosseini, Hou, Inan, Kardas, Kerkez, Khabsa, Kloumann, Korenev, Koura, Lachaux, Lavril, Lee, Liskovich, Lu, Mao, Martinet, Mihaylov, Mishra, Molybog, Nie, Poulton, Reizenstein, Rungta, Saladi, Schelten, Silva, Smith, Subramanian, Tan, Tang, Taylor, Williams, Kuan, Xu, Yan, Zarov, Zhang, Fan, Kambadur, Narang, Rodriguez, Stojnic, Edunov, and Scialom]{touvron2023llama2}
Hugo Touvron, Louis Martin, Kevin Stone, Peter Albert, Amjad Almahairi, Yasmine Babaei, Nikolay Bashlykov, Soumya Batra, Prajjwal Bhargava, Shruti Bhosale, Dan Bikel, Lukas Blecher, Cristian~Canton Ferrer, Moya Chen, Guillem Cucurull, David Esiobu, Jude Fernandes, Jeremy Fu, Wenyin Fu, Brian Fuller, Cynthia Gao, Vedanuj Goswami, Naman Goyal, Anthony Hartshorn, Saghar Hosseini, Rui Hou, Hakan Inan, Marcin Kardas, Viktor Kerkez, Madian Khabsa, Isabel Kloumann, Artem Korenev, Punit~Singh Koura, Marie-Anne Lachaux, Thibaut Lavril, Jenya Lee, Diana Liskovich, Yinghai Lu, Yuning Mao, Xavier Martinet, Todor Mihaylov, Pushkar Mishra, Igor Molybog, Yixin Nie, Andrew Poulton, Jeremy Reizenstein, Rashi Rungta, Kalyan Saladi, Alan Schelten, Ruan Silva, Eric~Michael Smith, Ranjan Subramanian, Xiaoqing~Ellen Tan, Binh Tang, Ross Taylor, Adina Williams, Jian~Xiang Kuan, Puxin Xu, Zheng Yan, Iliyan Zarov, Yuchen Zhang, Angela Fan, Melanie Kambadur, Sharan Narang, Aurelien Rodriguez, Robert Stojnic, Sergey Edunov, and Thomas Scialom.
\newblock Llama 2: Open foundation and fine-tuned chat models, 2023.

\bibitem[Vaswani et~al.(2017)Vaswani, Shazeer, Parmar, Uszkoreit, Jones, Gomez, Kaiser, and Polosukhin]{vaswani2017attention}
Ashish Vaswani, Noam Shazeer, Niki Parmar, Jakob Uszkoreit, Llion Jones, Aidan~N Gomez, {\L}ukasz Kaiser, and Illia Polosukhin.
\newblock Attention is all you need.
\newblock \emph{Advances in Neural Information Processing Systems (NIPS)}, 30, 2017.
\newblock URL \url{https://papers.nips.cc/paper/2017/hash/3f5ee243547dee91fbd053c1c4a845aa-Abstract.html}.

\bibitem[Wang et~al.(2024{\natexlab{a}})Wang, Duan, Zhang, Lin, and Chen]{wang-etal-2024-ada}
Chonghua Wang, Haodong Duan, Songyang Zhang, Dahua Lin, and Kai Chen.
\newblock {A}da-{LE}val: Evaluating long-context {LLM}s with length-adaptable benchmarks.
\newblock In Kevin Duh, Helena Gomez, and Steven Bethard (eds.), \emph{Proceedings of the 2024 Conference of the North American Chapter of the Association for Computational Linguistics: Human Language Technologies (Volume 1: Long Papers)}, pp.\  3712--3724, Mexico City, Mexico, June 2024{\natexlab{a}}. Association for Computational Linguistics.
\newblock \doi{10.18653/v1/2024.naacl-long.205}.
\newblock URL \url{https://aclanthology.org/2024.naacl-long.205}.

\bibitem[Wang et~al.(2024{\natexlab{b}})Wang, Ning, Pan, Wu, Guo, Deng, Bao, Hu, Zhang, Wang, and Zhang]{wang2024novelqabenchmarkingquestionanswering}
Cunxiang Wang, Ruoxi Ning, Boqi Pan, Tonghui Wu, Qipeng Guo, Cheng Deng, Guangsheng Bao, Xiangkun Hu, Zheng Zhang, Qian Wang, and Yue Zhang.
\newblock Novelqa: Benchmarking question answering on documents exceeding 200k tokens, 2024{\natexlab{b}}.
\newblock URL \url{https://arxiv.org/abs/2403.12766}.

\bibitem[Wang et~al.(2024{\natexlab{c}})Wang, Chen, Cheng, Liao, Zhang, Wu, Yu, Xu, Zhang, Luo, Li, Yang, Huang, and Li]{wang-etal-2024-leave}
Minzheng Wang, Longze Chen, Fu~Cheng, Shengyi Liao, Xinghua Zhang, Bingli Wu, Haiyang Yu, Nan Xu, Lei Zhang, Run Luo, Yunshui Li, Min Yang, Fei Huang, and Yongbin Li.
\newblock Leave no document behind: Benchmarking long-context {LLM}s with extended multi-doc {QA}.
\newblock In Yaser Al-Onaizan, Mohit Bansal, and Yun-Nung Chen (eds.), \emph{Proceedings of the 2024 Conference on Empirical Methods in Natural Language Processing}, pp.\  5627--5646, Miami, Florida, USA, November 2024{\natexlab{c}}. Association for Computational Linguistics.
\newblock \doi{10.18653/v1/2024.emnlp-main.322}.
\newblock URL \url{https://aclanthology.org/2024.emnlp-main.322/}.

\bibitem[Wei et~al.(2023)Wei, Wei, Tay, Tran, Webson, Lu, Chen, Liu, Huang, Zhou, and Ma]{wei2023largerlanguagemodelsincontext}
Jerry Wei, Jason Wei, Yi~Tay, Dustin Tran, Albert Webson, Yifeng Lu, Xinyun Chen, Hanxiao Liu, Da~Huang, Denny Zhou, and Tengyu Ma.
\newblock Larger language models do in-context learning differently, 2023.
\newblock URL \url{https://arxiv.org/abs/2303.03846}.

\bibitem[Wolf et~al.(2020)Wolf, Debut, Sanh, Chaumond, Delangue, Moi, Cistac, Rault, Louf, Funtowicz, Davison, Shleifer, von Platen, Ma, Jernite, Plu, Xu, Scao, Gugger, Drame, Lhoest, and Rush]{wolf2020huggingfacestransformersstateoftheartnatural}
Thomas Wolf, Lysandre Debut, Victor Sanh, Julien Chaumond, Clement Delangue, Anthony Moi, Pierric Cistac, Tim Rault, Rémi Louf, Morgan Funtowicz, Joe Davison, Sam Shleifer, Patrick von Platen, Clara Ma, Yacine Jernite, Julien Plu, Canwen Xu, Teven~Le Scao, Sylvain Gugger, Mariama Drame, Quentin Lhoest, and Alexander~M. Rush.
\newblock Huggingface's transformers: State-of-the-art natural language processing, 2020.
\newblock URL \url{https://arxiv.org/abs/1910.03771}.

\bibitem[Xiong et~al.(2023)Xiong, Liu, Molybog, Zhang, Bhargava, Hou, Martin, Rungta, Sankararaman, Oguz, Khabsa, Fang, Mehdad, Narang, Malik, Fan, Bhosale, Edunov, Lewis, Wang, and Ma]{xiong2023effective}
Wenhan Xiong, Jingyu Liu, Igor Molybog, Hejia Zhang, Prajjwal Bhargava, Rui Hou, Louis Martin, Rashi Rungta, Karthik~Abinav Sankararaman, Barlas Oguz, Madian Khabsa, Han Fang, Yashar Mehdad, Sharan Narang, Kshitiz Malik, Angela Fan, Shruti Bhosale, Sergey Edunov, Mike Lewis, Sinong Wang, and Hao Ma.
\newblock Effective long-context scaling of foundation models, 2023.

\bibitem[Xu et~al.(2024)Xu, Ye, and Ren]{xu2024stresstesting}
Xiaoyue Xu, Qinyuan Ye, and Xiang Ren.
\newblock Stress-testing long-context language models with lifelong {ICL} and task haystack.
\newblock In \emph{First Workshop on Long-Context Foundation Models @ ICML 2024}, 2024.
\newblock URL \url{https://openreview.net/forum?id=5ltgEKXnZt}.

\bibitem[Yang et~al.(2024)Yang, Yang, Hui, Zheng, Yu, Zhou, Li, Li, Liu, Huang, Dong, Wei, Lin, Tang, Wang, Yang, Tu, Zhang, Ma, Yang, Xu, Zhou, Bai, He, Lin, Dang, Lu, Chen, Yang, Li, Xue, Ni, Zhang, Wang, Peng, Men, Gao, Lin, Wang, Bai, Tan, Zhu, Li, Liu, Ge, Deng, Zhou, Ren, Zhang, Wei, Ren, Liu, Fan, Yao, Zhang, Wan, Chu, Liu, Cui, Zhang, Guo, and Fan]{yang2024qwen2technicalreport}
An~Yang, Baosong Yang, Binyuan Hui, Bo~Zheng, Bowen Yu, Chang Zhou, Chengpeng Li, Chengyuan Li, Dayiheng Liu, Fei Huang, Guanting Dong, Haoran Wei, Huan Lin, Jialong Tang, Jialin Wang, Jian Yang, Jianhong Tu, Jianwei Zhang, Jianxin Ma, Jianxin Yang, Jin Xu, Jingren Zhou, Jinze Bai, Jinzheng He, Junyang Lin, Kai Dang, Keming Lu, Keqin Chen, Kexin Yang, Mei Li, Mingfeng Xue, Na~Ni, Pei Zhang, Peng Wang, Ru~Peng, Rui Men, Ruize Gao, Runji Lin, Shijie Wang, Shuai Bai, Sinan Tan, Tianhang Zhu, Tianhao Li, Tianyu Liu, Wenbin Ge, Xiaodong Deng, Xiaohuan Zhou, Xingzhang Ren, Xinyu Zhang, Xipin Wei, Xuancheng Ren, Xuejing Liu, Yang Fan, Yang Yao, Yichang Zhang, Yu~Wan, Yunfei Chu, Yuqiong Liu, Zeyu Cui, Zhenru Zhang, Zhifang Guo, and Zhihao Fan.
\newblock Qwen2 technical report, 2024.
\newblock URL \url{https://arxiv.org/abs/2407.10671}.

\bibitem[Yang et~al.(2025)Yang, Yu, Li, Liu, Huang, Huang, Jiang, Tu, Zhang, Zhou, Lin, Dang, Yang, Yu, Li, Sun, Zhu, Men, He, Xu, Yin, Yu, Qiu, Ren, Yang, Li, Xu, and Zhang]{yang2025qwen251mtechnicalreport}
An~Yang, Bowen Yu, Chengyuan Li, Dayiheng Liu, Fei Huang, Haoyan Huang, Jiandong Jiang, Jianhong Tu, Jianwei Zhang, Jingren Zhou, Junyang Lin, Kai Dang, Kexin Yang, Le~Yu, Mei Li, Minmin Sun, Qin Zhu, Rui Men, Tao He, Weijia Xu, Wenbiao Yin, Wenyuan Yu, Xiafei Qiu, Xingzhang Ren, Xinlong Yang, Yong Li, Zhiying Xu, and Zipeng Zhang.
\newblock Qwen2.5-1m technical report, 2025.
\newblock URL \url{https://arxiv.org/abs/2501.15383}.

\bibitem[Yang et~al.(2018)Yang, Qi, Zhang, Bengio, Cohen, Salakhutdinov, and Manning]{yang-etal-2018-hotpotqa}
Zhilin Yang, Peng Qi, Saizheng Zhang, Yoshua Bengio, William Cohen, Ruslan Salakhutdinov, and Christopher~D. Manning.
\newblock {H}otpot{QA}: A dataset for diverse, explainable multi-hop question answering.
\newblock In Ellen Riloff, David Chiang, Julia Hockenmaier, and Jun{'}ichi Tsujii (eds.), \emph{Proceedings of the 2018 Conference on Empirical Methods in Natural Language Processing}, pp.\  2369--2380, Brussels, Belgium, October-November 2018. Association for Computational Linguistics.
\newblock \doi{10.18653/v1/D18-1259}.
\newblock URL \url{https://aclanthology.org/D18-1259}.

\bibitem[Yen et~al.(2024)Yen, Gao, and Chen]{yen2024longcontext}
Howard Yen, Tianyu Gao, and Danqi Chen.
\newblock Long-context language modeling with parallel context encoding.
\newblock In Lun-Wei Ku, Andre Martins, and Vivek Srikumar (eds.), \emph{Proceedings of the 62nd Annual Meeting of the Association for Computational Linguistics (Volume 1: Long Papers)}, pp.\  2588--2610, Bangkok, Thailand, August 2024. Association for Computational Linguistics.
\newblock \doi{10.18653/v1/2024.acl-long.142}.
\newblock URL \url{https://aclanthology.org/2024.acl-long.142}.

\bibitem[Yuan et~al.(2024)Yuan, Ning, Zhou, Yang, Li, Zhuang, Tan, Yao, Lin, Li, Dai, Yan, and Wang]{yuan2024lvevalbalancedlongcontextbenchmark}
Tao Yuan, Xuefei Ning, Dong Zhou, Zhijie Yang, Shiyao Li, Minghui Zhuang, Zheyue Tan, Zhuyu Yao, Dahua Lin, Boxun Li, Guohao Dai, Shengen Yan, and Yu~Wang.
\newblock Lv-eval: A balanced long-context benchmark with 5 length levels up to 256k, 2024.
\newblock URL \url{https://arxiv.org/abs/2402.05136}.

\bibitem[Zhang \& Bansal(2021)Zhang and Bansal]{zhang-bansal-2021-finding}
Shiyue Zhang and Mohit Bansal.
\newblock Finding a balanced degree of automation for summary evaluation.
\newblock In \emph{Empirical Methods in Natural Language Processing (EMNLP)}, pp.\  6617--6632, 2021.
\newblock \doi{10.18653/v1/2021.emnlp-main.531}.
\newblock URL \url{https://aclanthology.org/2021.emnlp-main.531}.

\bibitem[Zhang et~al.(2024{\natexlab{a}})Zhang, Zhang, Long, Xie, Dai, Tang, Lin, Yang, Xie, Huang, Zhang, Li, and Zhang]{zhang2024mgtegeneralizedlongcontexttext}
Xin Zhang, Yanzhao Zhang, Dingkun Long, Wen Xie, Ziqi Dai, Jialong Tang, Huan Lin, Baosong Yang, Pengjun Xie, Fei Huang, Meishan Zhang, Wenjie Li, and Min Zhang.
\newblock mgte: Generalized long-context text representation and reranking models for multilingual text retrieval, 2024{\natexlab{a}}.
\newblock URL \url{https://arxiv.org/abs/2407.19669}.

\bibitem[Zhang et~al.(2024{\natexlab{b}})Zhang, Chen, Hu, Xu, Chen, Hao, Han, Thai, Wang, Liu, and Sun]{zhang2024inftybenchextendinglongcontext}
Xinrong Zhang, Yingfa Chen, Shengding Hu, Zihang Xu, Junhao Chen, Moo~Khai Hao, Xu~Han, Zhen~Leng Thai, Shuo Wang, Zhiyuan Liu, and Maosong Sun.
\newblock $\infty$bench: Extending long context evaluation beyond 100k tokens, 2024{\natexlab{b}}.
\newblock URL \url{https://arxiv.org/abs/2402.13718}.

\bibitem[Zheng et~al.(2023)Zheng, Chiang, Sheng, Zhuang, Wu, Zhuang, Lin, Li, Li, Xing, Zhang, Gonzalez, and Stoica]{zheng2023judging}
Lianmin Zheng, Wei-Lin Chiang, Ying Sheng, Siyuan Zhuang, Zhanghao Wu, Yonghao Zhuang, Zi~Lin, Zhuohan Li, Dacheng Li, Eric Xing, Hao Zhang, Joseph~E. Gonzalez, and Ion Stoica.
\newblock Judging {LLM}-as-a-judge with {MT}-bench and chatbot arena.
\newblock In \emph{Thirty-seventh Conference on Neural Information Processing Systems Datasets and Benchmarks Track}, 2023.
\newblock URL \url{https://openreview.net/forum?id=uccHPGDlao}.

\end{thebibliography}
\bibliographystyle{iclr2025_conference}

\appendix
\clearpage

\section{Comparison with other benchmarks}
\label{app:related_works_compare}

\rebuttal{
Due to the lack of large-scale human studies on long-context language models, discerning the true rankings of LCLMs remains challenging.
Therefore, we rely on previous works and empirical observations to guide our expectations.
Specifically, the Gemini 1.5 report demonstrates that larger models consistently outperform smaller ones at handling long contexts---both qualitatively and quantitatively across the Gemini, GPT, and Claude model families \citep[Section 5.2, Tables 4 and 5][]{geminiteam2024gemini15unlockingmultimodal}. 
Although their evaluation suite is not publicly available, these results align with commonly observed patterns in language model performance. Given that obtaining ground truth rankings through human evaluation is prohibitively expensive and practically infeasible, relying on such well-established patterns is reasonable.
In this section, we further examine HELMET's improvements over existing benchmarks through direct comparisons of model rankings and performance, supported by careful ablation studies.
}

\subsection{Results}

We build upon previous datasets and refine evaluation settings to better reflect model capabilities.
Specifically, we evaluate models on \infbench, another benchmark that evaluates long-context at 128K tokens.
The results reproduced from the original authors' code are shown in Table \ref{tab:results_infbench}.
The numerical results for Figure \ref{fig:comparison_teaser} are presented in Table \ref{tab:results_teaser}.

\rebuttal{
Upon closer inspection, we find that a set of open-sourced models' ranking and performance on HELMET and \infbench{} in Table \ref{tab:results_ranking}.
In addition to the previously noted performance discrepancy between Llama-3.1-8B-Instruct and Llama-3.1-70B-Instruct, we observe that all Llama-3.2 models degenerate in performance on \infbench{} at input length 128K.
However, on HELMET, the Llama-3.2 models, especially Llama-3.2-3B-Instruct, rank well against other open-source models. Upon qualitative examination, we find that the Llama 3.2 models are able to produce coherent and useful generation at long contexts with better prompting strategies from HELMET, such as adding in-context learning examples.

Our ablation study on few-shot demonstrations in Table \ref{tab:results_ablate_shots} shows that 2-shot demonstrations substantially improve base model performance, better reflecting real-world usage patterns.
As shown in Table \ref{tab:results_infbench_compare}, improved instructions and prompting enable smaller Llama-3.2 models to compete effectively with larger models.
Thus, HELMET provides a better and more realistic reflection of how these models would be used in practice over previous benchmarks.
These findings demonstrate that HELMET provides a more accurate and practical assessment of model capabilities compared to previous benchmarks.
}

\begin{table}[th]
    \caption{
        Results with 128k input length across different benchmarks.
    }
    \vspace*{4pt}
    \small
    \centering
        \begin{tabular}{lrrrrr}
            \toprule
            \bfseries Model & \bfseries Claimed Length & \bfseries NIAH & \bfseries RULER & \bfseries \infbench & \bfseries \ours \\
            \midrule
            GPT-4o-mini & 128000 & \bfseries {\cellcolor[HTML]{4A895B}} \color[HTML]{F1F1F1} 100.0 & {\cellcolor[HTML]{9ABDA4}} \color[HTML]{000000} 80.8 & {\cellcolor[HTML]{94B99E}} \color[HTML]{000000} 51.9 & {\cellcolor[HTML]{B3CDBA}} \color[HTML]{000000} 54.1 \\
            GPT-4o-08 & 128000 & \bfseries {\cellcolor[HTML]{4A895B}} \color[HTML]{F1F1F1} 100.0 & \bfseries {\cellcolor[HTML]{4A895B}} \color[HTML]{F1F1F1} 93.3 & {\cellcolor[HTML]{669B74}} \color[HTML]{F1F1F1} 57.1 & \bfseries {\cellcolor[HTML]{4A895B}} \color[HTML]{F1F1F1} 63.8 \\
            \midrule
            Gemini-1.5-Flash & 1048576 & \bfseries {\cellcolor[HTML]{4A895B}} \color[HTML]{F1F1F1} 100.0 & {\cellcolor[HTML]{75A582}} \color[HTML]{F1F1F1} 86.6 & {\cellcolor[HTML]{9DBFA6}} \color[HTML]{000000} 50.8 & {\cellcolor[HTML]{D7E5DB}} \color[HTML]{000000} 50.7 \\
            Gemini-1.5-Pro & 2097152 & {\cellcolor[HTML]{FFFFFF}} \color[HTML]{000000} 45.3 & {\cellcolor[HTML]{FFFFFF}} \color[HTML]{000000} 65.3 & \bfseries {\cellcolor[HTML]{4A895B}} \color[HTML]{F1F1F1} 60.3 & {\cellcolor[HTML]{559065}} \color[HTML]{F1F1F1} 62.7 \\
            \midrule
            Llama-3.1-8B-Inst & 131072 & \bfseries {\cellcolor[HTML]{4A895B}} \color[HTML]{F1F1F1} 100.0 & {\cellcolor[HTML]{97BBA1}} \color[HTML]{000000} 81.3 & {\cellcolor[HTML]{D9E6DC}} \color[HTML]{000000} 44.1 & {\cellcolor[HTML]{FFFFFF}} \color[HTML]{000000} 47.0 \\
            Llama-3.1-70B-Inst & 131072 & \bfseries {\cellcolor[HTML]{4A895B}} \color[HTML]{F1F1F1} 100.0 & {\cellcolor[HTML]{BBD3C1}} \color[HTML]{000000} 75.8 & {\cellcolor[HTML]{FFFFFF}} \color[HTML]{000000} 39.7 & {\cellcolor[HTML]{E7EFE9}} \color[HTML]{000000} 49.3 \\
            \bottomrule
            \end{tabular}
    \label{tab:results_teaser}
\end{table}

\begin{table}[th]
    \caption{
        Results on \infbench \citep{zhang2024inftybenchextendinglongcontext}. 
        We report numbers from running the original authors repo: \url{https://github.com/OpenBMB/InfiniteBench}.
        The original code did not support evaluation of the Gemini and Llama 3 models at the time of access, so we evaluate them by following the script and template for GPT-4 and open-source models, respectively.
        We exclude coding tasks from the evaluation suite since it is out of the scope for general long-context language modeling tasks.
    }
    \vspace*{4pt}
    \resizebox{0.98\linewidth}{!}{
        \begin{tabular}{lrrrrrrrrrr}
            \toprule
            Model &  MC &  QA &  Sum &  Diag &  Calc & Find & Number &  PassKey &  KV &  Avg. \\
            \midrule
            GPT4                  &                 75.9 &             23.7 &              10.7 &                 18.5 &        0.0 &       47.1 &          100.0 &    100.0 &          58.8 & 44.9 \\
            GPT-4o-05              &                 88.2 &             37.9 &              23.7 &                 28.5 &        0.0 &       58.6 &          100.0 &    100.0 &          94.2 & 55.4 \\
            GPT-4o-08              &                 86.5 &             26.0 &              21.5 &                 51.0 &        0.0 &       58.9 &          100.0 &    100.0 &          99.8 & 57.1 \\
            GPT-4o-mini            &                 78.2 &             19.1 &              24.8 &                 21.5 &        0.0 &       69.7 &          100.0 &    100.0 &          80.4 & 51.9 \\
            \midrule
            Gemini-1.5-Flash      &                 76.0 &             42.1 &              30.0 &                 55.8 &        0.0 &       47.4 &          100.0 &    100.0 &          31.4 & 50.8 \\
            Gemini-1.5-Pro        &                 77.5 &             27.7 &              29.0 &                 97.5 &        0.0 &       58.0 &          100.0 &    100.0 &          70.4 & 60.3 \\
            \midrule
            Llama-3.1-8B          &                 56.8 &              8.8 &              14.3 &                  0.5 &        0.0 &       22.0 &           99.7 &    100.0 &          18.8 & 33.0 \\
            Llama-3.1-8B-Inst     &                 67.2 &             15.5 &              26.7 &                 23.0 &        0.0 &       33.1 &           99.5 &    100.0 &          55.0 & 44.1 \\
            Llama-3.1-70B         &                 66.4 &              9.2 &              17.5 &                  8.5 &        0.0 &       32.3 &          100.0 &    100.0 &          13.4 & 35.1 \\
            Llama-3.1-70B-Inst    &                 75.5 &             23.3 &              31.3 &                 18.0 &        0.0 &       43.1 &           99.7 &    100.0 &           2.6 & 39.7 \\
            Llama-3.2-1B          &                  2.2 &              1.4 &               8.6 &                  4.5 &        0.0 &        0.0 &            1.5 &      0.0 &           0.0 &  2.0 \\
            Llama-3.2-1B-Instruct &                  3.5 &              1.4 &              12.5 &                  5.5 &        0.0 &        0.0 &            0.0 &      0.0 &           0.0 &  2.4 \\
            Llama-3.2-3B          &                  1.3 &              1.2 &               7.6 &                  4.5 &        0.0 &        0.0 &            1.7 &      0.0 &           0.0 &  1.9 \\
            Llama-3.2-3B-Instruct &                  2.2 &              1.6 &              13.3 &                  4.5 &        0.0 &        0.0 &            1.7 &      1.7 &           0.0 &  2.8 \\
            \bottomrule
        \end{tabular}

        }
    \label{tab:results_infbench}
\end{table}

\begin{table}[th]
    \caption{
        We compare the results on \infbench{} multiple-choice (MC) and question answering (QA) tasks between the original authors and \ours{}.
        In our implementation, we leverage refined prompts and carefully crafted parsing scripts to ensure robust and fair evaluation.
        For example, we find that Llama-3.2 models are actually much better long-context language models than \infbench{} would suggest.
   }
    \vspace*{4pt}
    \small
    \centering
        \begin{tabular}{lrrrrr}
            \toprule
            \textbf{} & \textbf{} & \multicolumn{2}{c}{\textbf{Original}} & \multicolumn{2}{c}{\textbf{\ours}} \\
             \cmidrule(lr){3-4} \cmidrule(lr){5-6}
            \textbf{Model} & \textbf{Claimed Length} & \textbf{MC} & \textbf{QA} & \textbf{MC} & \textbf{QA} \\
            \midrule
            GPT-4o-mini & 128000 & 88.2 & 37.9 & 71.0 & 44.5 \\
            GPT-4o-05 & 128000 & 86.5 & 26.0 & 76.0 & 53.0 \\
            GPT-4o-08 & 128000 & 78.2 & 19.1 & 73.0 & 47.9 \\
            Gemini-1.5-Flash & 1048576 & 76.0 & 42.1 & 79.1 & 51.7 \\
            Gemini-1.5-Pro & 2097152 & 77.5 & 27.7 & 83.3 & 50.9 \\
            \midrule
            Llama-3.1-8B & 131072 & 56.8 & 8.8 & 54.0 & 38.8 \\
            Llama-3.1-8B-Inst & 131072 & 67.2 & 15.5 & 49.0 & 40.3 \\
            Llama-3.2-1B & 131072 & 2.2 & 1.4 & 21.0 & 17.6 \\
            Llama-3.2-1B-Inst & 131072 & 3.5 & 1.4 & 16.0 & 13.3 \\
            Llama-3.2-3B & 131072 & 1.3 & 1.2 & 36.0 & 31.0 \\
            Llama-3.2-3B-Inst & 131072 & 2.2 & 1.6 & 42.0 & 20.6 \\
            \bottomrule
        \end{tabular}

    \label{tab:results_infbench_compare}
\end{table}

\begin{table}[th]
    \caption{
        We evaluate the performance of models on a subset of \ours{} tasks with 0-shot and 2-shot demonstrations at 128k input length to understand the impact of ICL on model performance, averaged across three random seeds.
        The standard deviation across three runs are shown in the subscript.
        We observe that the performance is generally higher for 2-shot demonstrations compared to 0-shot demonstrations.
        Crucially, the 2-shot examples enable base models to achieve higher results that reflect the model's long-context capabilities in realistic settings, such as for MSMARCO and JSON KV.
        \base denotes base models.
    }
    \vspace*{4pt}
    \small
    \centering
    \resizebox{0.98\linewidth}{!}{
        \begin{tabular}{lrrrrrrrrrr}
            \toprule
            \textbf{} & \multicolumn{2}{c}{\textbf{JSON KV}} & \multicolumn{2}{c}{\textbf{NQ}} & \multicolumn{2}{c}{\textbf{MSMARCO}} & \multicolumn{2}{c}{\textbf{\infbench{} MC}} & \multicolumn{2}{c}{\textbf{\infbench{} QA}} \\
            \cmidrule(lr){2-3} \cmidrule(lr){4-5} \cmidrule(lr){6-7} \cmidrule(lr){8-9} \cmidrule(lr){10-11}
            \textbf{Model} & \textbf{0 shot} & \textbf{2 shot} & \textbf{0 shot} & \textbf{2 shot} & \textbf{0 shot} & \textbf{2 shot} & \textbf{0 shot} & \textbf{2 shot} & \textbf{0 shot} & \textbf{2 shot} \\
            \midrule

            GPT-4o-mini\inst & $92.3_{5.5}$ & $93.7_{2.1}$ &$57.8_{1.1}$ &$60.4_{2.6}$ &$24.8_{0.7}$ &$31.4_{1.1}$ & $72.0_{2.6}$ & $69.3_{4.7}$ & $28.8_{1.9}$ & $47.1_{4.4}$ \\
            GPT-4o-05\inst & $99.3_{0.6}$ & $36.7_{7.1}$ &$59.4_{4.0}$ &$63.0_{2.1}$ &$42.7_{0.9}$ &$46.8_{1.3}$ & $73.7_{0.6}$ & $76.3_{0.6}$ & $49.1_{1.9}$ & $51.3_{3.4}$ \\
            GPT-4o-08\inst &$100.0_{0.0}$ &$100.0_{0.0}$ &$58.5_{2.0}$ &$61.9_{4.5}$ &$46.6_{0.3}$ &$48.7_{1.0}$ & $75.3_{2.5}$ & $72.7_{0.6}$ & $29.8_{0.9}$ & $48.6_{1.5}$ \\
            Gemini-1.5-Flash\inst & $98.3_{0.6}$ & $99.0_{1.0}$ &$53.7_{5.9}$ &$51.5_{4.5}$ &$44.9_{3.0}$ &$49.9_{0.9}$ & $73.9_{2.3}$ & $77.1_{3.2}$ & $48.2_{4.7}$ & $49.9_{2.6}$ \\
            Gemini-1.5-Pro\inst & $97.7_{0.6}$ & $93.7_{2.3}$ &$62.2_{2.6}$ &$59.2_{2.7}$ &$57.1_{0.9}$ &$58.4_{0.3}$ & $83.6_{2.8}$ & $83.3_{0.7}$ & $46.4_{1.8}$ & $47.2_{3.2}$ \\
            \midrule
            Llama-3.1-8B\base & $77.3_{2.1}$ & $98.0_{1.0}$ &$45.7_{2.1}$ &$44.9_{3.9}$ & $0.1_{0.0}$ & $7.5_{0.5}$ & $55.3_{5.5}$ & $53.3_{8.0}$ & $36.1_{2.0}$ & $36.6_{2.1}$ \\
            Llama-3.1-8B-Inst\inst & $96.0_{1.0}$ & $95.7_{0.6}$ &$48.4_{1.3}$ &$48.9_{2.6}$ & $5.4_{0.0}$ &$13.8_{1.1}$ & $49.7_{2.1}$ & $52.0_{2.6}$ & $29.1_{2.0}$ & $37.9_{2.6}$ \\
            Llama-3.2-1B\base & $34.0_{3.5}$ & $34.3_{2.1}$ &$25.1_{0.7}$ &$23.2_{2.8}$ & $0.3_{0.0}$ & $4.8_{0.5}$ & $23.0_{2.6}$ & $22.7_{1.5}$ & $18.9_{2.5}$ & $18.9_{3.3}$ \\
            Llama-3.2-1B-Inst\inst &$6.3_{1.5}$ &$9.3_{3.2}$ &$23.3_{4.4}$ &$22.9_{1.4}$ & $0.0_{0.0}$ & $0.5_{0.3}$ & $16.0_{2.6}$ & $16.3_{2.5}$ & $12.6_{0.4}$ & $12.7_{0.5}$ \\
            Llama-3.2-3B\base & $30.3_{2.1}$ & $54.0_{8.9}$ &$40.2_{2.0}$ &$42.3_{4.1}$ & $0.1_{0.0}$ & $6.0_{1.3}$ & $43.0_{1.0}$ & $40.7_{7.2}$ & $31.3_{1.1}$ & $31.6_{2.2}$ \\
            Llama-3.2-3B-Inst\inst & $31.3_{2.3}$ & $36.7_{2.5}$ &$44.2_{2.5}$ &$42.8_{3.8}$ & $0.2_{0.0}$ & $0.9_{0.2}$ & $36.0_{2.0}$ & $40.0_{3.5}$ & $17.8_{1.2}$ & $19.7_{0.8}$ \\
            \bottomrule
        \end{tabular}
    }
    \label{tab:results_ablate_shots}
\end{table}

\begin{table}[th]
    \caption{
        Comparison of model rankings on HELMET and \infbench{}.
    }
    \vspace*{4pt}
    \centering
    \resizebox{0.7\linewidth}{!}{
        \begin{tabular}{lr}
            \toprule
            \textbf{Model} & \textbf{HELMET} \\
            \midrule
            Llama-3.1-70B-Inst & 49.3 \\
            Llama-3.1-8B-Inst & 47.0 \\
            Llama-3.1-70B & 41.3 \\
            Yi-34B-200k & 38.3 \\
            Llama-3.2-3B-Inst & 36.9 \\
            Llama-3.1-8B & 35.6 \\
            Yi-9B-200k & 33.0 \\
            Llama-3.2-3B & 31.9 \\
            Yi-6B-200k & 26.3 \\
            Llama-3.2-1B-Inst & 24.6 \\
            Llama-3.2-1B & 21.2\\
            \bottomrule
        \end{tabular}

        \begin{tabular}{lr}
            \toprule
            \textbf{Model} & \textbf{InfBench} \\
            \midrule
            Llama-3.1-8B-Inst & 46.7 \\
            Llama-3.1-70B-Inst & 43.7 \\
            Yi-34B-200k & 43.1 \\
            Llama-3.1-70B & 38.6 \\
            Yi-9B-200k & 37.6 \\
            Llama-3.1-8B & 35.7 \\
            Yi-6B-200k & 32.0 \\
            Llama-3.2-3B-Inst & 2.8 \\
            Llama-3.2-1B-Inst & 2.6 \\
            Llama-3.2-1B & 2.0 \\
            Llama-3.2-3B & 1.8 \\
            \bottomrule
        \end{tabular}

    }
    \label{tab:results_ranking}
\end{table}

\subsection{Discussions}
\label{app:related_works_discussions}
Many existing long-context language modeling benchmarks are studied in isolated settings, such as synthetic tasks \citep{hsieh2024ruler,li2024needlebenchllmsretrievalreasoning,levy2024task,laban2024summaryhaystackchallengelongcontext}, in-context learning \citep{anil2024many,bertsch2024incontext,li2024longiclbench,agarwal2024manyshotincontextlearning}, summarization \citep{chang2024booookscore,kim2024fables}, and retrieval-augmented settings \citep{lee2024longcontextlanguagemodelssubsume}.
However, these works lack a unified evaluation across diverse downstream tasks.

There are also benchmarks that sought to unify different datasets, such as \infbench{} \citep{zhang2024inftybenchextendinglongcontext}, LongBench \citep{bai-etal-2024-longbench}, L-Eval \citep{an-etal-2024-l}, LV-Eval \citep{yuan2024lvevalbalancedlongcontextbenchmark}, and ZeroSCROLLS \citep{shaham-etal-2023-zeroscrolls}.
Many of these benchmarks are still limited by the context length, evaluation metrics, or both.
Most similar to our work, \infbench{}\cite{zhang2024inftybenchextendinglongcontext} also evaluates models at context lengths at 128K tokens.
However, their evaluation settings are limited to a few domains---synthetic, QA, and summarization.
Although they also evaluate coding tasks, the domain is limited to code-specialized models outside of the realm of general-purposed language models.
Furthermore, their summarization evaluations lack robust evaluation and still rely on ROUGE scores.
Similarly, LV-Eval \citep{yuan2024lvevalbalancedlongcontextbenchmark} evaluates long-context models across different lengths but is limited to the QA tasks.
Furthermore, we are the first to evaluate existing LCLMs comprehensively---we evaluate \nmodels{} models of different sizes and architectures, which enables previously unavailable insights into the correlation across diverse tasks and the landscape of current models. 

\rebuttal{
To summarize, HELMET improves upon existing benchmarks by providing a comprehensive evaluation across diverse tasks and carefully designed evalution metrics and prompting strategies.
Consequently, HELMET provides a more accurate reflection of the models' capabilities in practice.
}

\section{Datasets}
\label{app:datasets}

\subsection{Retrieval-augmented generation}
\label{app:datasets:odqa}
Natural Questions \citep[NQ;][]{kwiatkowski2019natural} is a large-scale dataset for open-domain question answering featuring real user queries.
TriviaQA \citep[TQA;][]{joshi2017triviaqa} comprises trivia questions and their corresponding answers.
HotpotQA \citep{yang-etal-2018-hotpotqa} contains questions that require multi-passage reasoning.
We source these datasets from KILT \citep{petroni-etal-2021-kilt}, which provides annotations linking each query to its corresponding gold passages containing the answers.

PopQA \citep{mallen-etal-2023-trust} consists of questions about long-tail entities. To minimize the impact of pre-training memorization, we filter the dataset to include only subject entities with popularity scores below 1000.
Since gold passages are not available for PopQA, we classify retrieved passages as positive or negative by checking for the presence of the ground truth answer.

Notably, we populate the context with \emph{hard negative passages}, retrieved from the same corpus as the positive passages using a real retriever. This approach presents a significantly greater challenge than using randomly sampled passages.
This design choice better reflects real-world retrieval-augmented generation tasks, where models must effectively distinguish between relevant and irrelevant information.

\subsection{Generation with citations}
\label{app:datasets:cite}
Generation with citations represents a crucial task for enhancing language model trustworthiness and verifiability \citep{bohnet2022attributed,gao2023alce,asai2024reliableadaptableattributablelanguage}.
We employ ALCE to assess models' capability to generate properly cited text.
Following the original methodology, we utilize Wikipedia as the retrieval corpus and GTR \citep{ni-etal-2022-large} as the retriever.
We omit MAUVE-based fluency evaluation since models typically generate fluent text, and in cases where they don't, the other metrics already approach zero.
Instead, we focus on correctness and citation quality, reporting their average.

\subsection{Passage re-ranking}
\label{app:datasets:rerank}
We evaluate models' passage re-ranking capabilities \citep{sun-etal-2023-chatgpt} using the MS MARCO passage ranking dataset \citep{bajaj2018ms}.
Our evaluation uses annotated datasets from the TREC Passage Re-ranking challenge \citep{craswell2020overviewtrec2019deep}.
Each dataset instance consists of a query and a list of passages with associated relevance scores.
The dataset combines Bing user queries with passages retrieved via BM25 from web pages.
Passages are labeled as perfect, highly relevant, or not relevant.
For each input length $L$, we determine the number of passages $k$ that can be included.
To eliminate positional bias, we balance the label distribution and randomize passage order.
We create three different permutations of the $k$ passages for each input.
The final performance is reported in NDCG@10.
While higher values of $k$ are possible for NDCG evaluation, the computational cost of generating numerous passage IDs during inference leads us to maintain NDCG@10 as our metric.

\subsection{In-context learning}
\label{app:datasets:icl}
For in-context learning, we implement a label mapping strategy that compels models to utilize in-context examples rather than relying on pre-trained knowledge for classification tasks.
Each label is randomly mapped to an integer $l \in \{0, 1, \ldots, n-1\}$, where $n$ represents the number of unique labels in the dataset, following established practices \citep{pan-etal-2023-context}.
We then replace all label texts with their corresponding integer mappings throughout the dataset.
Following \citet{li2024longiclbench}, we organize examples into demonstration rounds, with each round containing exactly one example per label in randomized order.
The input is constructed by concatenating these demonstration rounds until reaching the target input length $L$.
Unlike other task categories where we evaluate 100 samples, for ICL datasets we evaluate 500 samples.
We balance the label distribution among the evaluation set.

\subsection{Synthetic tasks}
\label{app:datasets:synthetic}
For RULER tasks, we generate data using the original authors' scripts \citep{hsieh2024ruler}, employing the Llama-2 tokenizer \citep{touvron2023llama2} to standardize input text across all models.
The RULER suite comprises the following tasks:
\begin{itemize}
    \item NIAH Single (three variants):
        \begin{itemize}
            \item NIAH (variant 2): Most similar to the original NIAH \citep{gkamradt_llmtest_needleinahaystack_2024}
            \item NIAH Single Repeat (variant 1): Uses repeated phrases instead of Paul Graham essays as context
            \item NIAH Single UUID (variant 3): Similar to variant 2 but uses UUIDs as retrieval values
        \end{itemize}
    \item NIAH MultiKey (MK, three variants):
        \begin{itemize}
            \item NIAH MK Essay (variant 1): Involves retrieving one gold key among three irrelevant keys, using Paul Graham Essays as context
            \item NIAH MK Needle (variant 2): Uses needle-based context structure
            \item NIAH MK UUID (variant 3): Similar to variant 2 but employs UUIDs as retrieval values for all needles
        \end{itemize}
    \item NIAH MultiValue (MV): Requires retrieving four different numbers associated with a single key from irrelevant essay context
    \item NIAH MultiQuery (MQ): Involves retrieving correct values for four distinct keys from irrelevant essay context
    \item Variable Tracking (VT): Requires tracking variable values through sequential operations
    \item Word Extraction Tasks:
        \begin{itemize}
            \item Common word extraction (CWE)
            \item Frequent word extraction (FWE)
            \item Both tasks require identifying the most frequently occurring words
        \end{itemize}
    \item Question Answering Tasks:
        \begin{itemize}
            \item SQuAD (variant 1; \citealp{rajpurkar2016squad})
            \item HotpotQA (variant 2; \citealp{yang-etal-2018-hotpotqa})
            \item Notable distinction: Our implementation uses retrieved passages rather than random samples, making the task more realistic and challenging
        \end{itemize}
\end{itemize}
For comprehensive details, please refer to the original paper \citep{hsieh2024ruler}.

For JSON KV tasks, we generate a JSON dictionary containing randomly generated UUIDs as both keys and values, similar to \citet{Liu2023LostIT}.
For each dictionary, we construct six queries, asking the model to retrieve the value for each key at six evenly spaced positions within the context.

\subsection{Model-based evaluation}
\label{app:model-eval}
Automatic evaluation metrics, such as ROUGE, are known to be unreliable and uncorrelated with human judgments \citep{goyal2023newssummarizationevaluationera,chang2024booookscore}. 
However, existing long-context benchmarks still largely rely on these metrics \citep{bai-etal-2024-longbench,zhang2024inftybenchextendinglongcontext}. 
In this work, we seek to more reliably evaluate language models at long-contexts by leveraging LLMs as judges, inspired by previous works \citep{zheng2023judging}.
For all model-based evaluations, we use \ttt{GPT-4o-2024-05-13} as the judge.

\paragraph{Long-document question answering.}
For NarrativeQA, we found that the models can often output answers that are correct but have little surface form or lexical overlaps with the ground truth answers in preliminary experiments, resulting in lower-than-expected performance.
This is often due to the long lengths of the ground truth answers, which gives more possibilities of how to write it than a simple named entity that is often the case of other QA datasets.

Therefore, we employ an LLM to judge if the model output is fluent and semantically similar to the ground truth.
Given the question, the ground truth answer, and the model-generated output, we ask the LLM to judge if (1) the model output is fluent and (2) the generated output is correct and relevant to the question with the ground truth as a reference.
The detailed prompts, precise definitions, and instructions are found in Table \ref{tab:model_prompt_qa}.

In Table \ref{tab:results_model_eval}, we also find that model evaluation can be useful in catching extremely cases.
For example, Claude scores low in terms of F1 due to its tendency to output extra, assistant-like text, which is penalized by the n-gram overlap metric, and appears to be worse than the much smaller Llama-3.2-3B-Inst model.
However, the model-based evaluation is able to catch this issue, and Claude scores higher than Llama-3.2-3B-Inst, which users might have expected given the model sizes.

\paragraph{Summarization}.
At a high level, we ask the model to check for three properties: fluency, precision, and recall.
Fluency can take on a value of either 0 (incoherent, incomplete, and/or repetitive) or 1 (fluent and coherent).
Precision is the percentage of model-generated sentences that are supported by the gold summary (we use the long summary from \mls{} here).
Recall is the percentage of the key points that are supported by the model-generated output.
We calculate the F1 score from precision and recall and multiply it with the fluency score for the final score.

We first ask the model to generate a list of key points or atomic claims from the gold summary, following previous works that show that LLMs can accurately decompose long texts \citep{kamoi2023wice,gao2023alce}.
We manually checked 105 claims originating from 25 \mls{} summaries and found that the claims were all factually correct.
Although we found one out of 25 instances where the model missed a possible key point in the given summary, we found that GPT-4o is almost always reliable for the decomposition task.
For \mls{}, we use the short summary to obtain the key point as the annotation contains both a long and a short summary.
These key points are then saved for \mls{} and \infbench{} Sum.

We show the detailed prompts for the summarization tasks in Table \ref{tab:model_prompt_summ_fluency}, \ref{tab:model_prompt_summ_recall}, and \ref{tab:model_prompt_summ_precision}, which are modeled after previous works \citep{kamoi2023wice,chang2024booookscore,kim2024fables}.
These previous works have shown that LLMs can effectively judge model outputs \citep{zheng2023judging}, but we conduct human analysis to further verify the evaluation metric.

\rebuttal{
From qualitative analysis, we found that the model is consistently able to distinguish between fluent and non-fluent outputs, where we agree with the model judgments 100\% of the time for randomly sampled \mls{} and \infbench{} Sum outputs.
We then check if the GPT-4o judgments for precision and recall agree with human judgments.
To this end, we sample 10 generated summaries for \mls{} and \infbench{} each (from Gemini-1.5-Pro and Llama-3.1-70B-Inst) and check five key point evaluations for each summary.
We follow a similar procedure where we check if the model output supports the key point.
We observed Cohen's $\kappa = 0.76$ for \infbench{} Sum and $\kappa = 0.72$ for \mls, suggesting substantial agreement.
For precision, we conduct a similar human evaluation, and found a $\kappa = 0.91$ for \infbench{} Sum and $\kappa = 0.83$ for \mls, suggesting near-perfect agreement.

Inspecting the disagreements, we find that that most of them arise from the partially supported cases. For instance, the key point may include specific details, such as the names of government departments or Court Justices' names, that are not explicitly mentioned in the generated summary, and the model judge is typically more lenient about the exclusion of these small details while humans are more strict. However, this is also subjective to the preference of the human.

Qualitatively, We find that model-based evaluation can catch cases where the model is overly repetitive and scores high ROUGE-L score as a result, such as Mistral-7B-Inst-v0.3 on \infbench{} Sum.
For instance, the Mistral model may generate an output consisted of the sentence "The author's object is to show that the study of grammar is necessary part of education" repeated for hundreds of times. This summary would receive an ROUGE-L score of $12.3$ while the GPT-4o judge would penalize the model for being overly repetitive and incoherent and assign the output a final score of $0.0$.
Thus, our GPT-4o judge penalizes the model for being overly repetitive, and the final metric reflects this issue.
}

\begin{table}[h]
    \caption{Model-based evaluation prompt for long-document question answering.}
    \centering
    \small
    \begin{tabular}{>{\raggedright\arraybackslash\tt}p{0.98\textwidth}<{}}
        \toprule
        Please act as an impartial judge and evaluate the quality of the provided answer which attempts to answer the provided question based on a provided context.
        Although you are not given the context, you will be given a set of correct answers that achieves full scores on all metrics, and you need to assess the provided answers using the correct answers.
\\\\
        Below is your grading rubric:

\\\\
        Fluency:
        - Score 0 (incoherent, repetitive, or incomplete): Incoherent sentences, repetitive sentences (even if not by exact words), incomplete answers, or gibberish. Note that even if the answer is coherent, if it is repetitive or incomplete, it should be given a score of 0.
        - Score 1 (coherent, non-repetitive answer): Coherent, non-repetitive, fluent, grammatically correct answers.

\\\\
        Correctness:
        - Score 0 (Incorrect): The answer does not agree with the provided correct answers at all.
        - Score 1 (partly correct): Partly agree with one of the provided correct answers (for example, the question asks for a date and a person; the answer gets the date right but the person wrong).
        - Score 2 (correct but not fully relevant): Fully agrees with one of the provided correct answers but mentions other completely irrelevant information. Note that extra details provided in the answer, even if not mentioned in the correct answers, should NOT be seen as irrelevant as long as they are relevant to the question to a reasonable extend.
        - Score 3 (correct and relevant): Fully agrees with one of the provided correct answers and only provides information relevant to the question. Note that if the answer is longer than the correct answer, as long as everything in the answer is relevant to the question, it should still be given score 3. For example, if the correct answer is ''the North Pole'' and the answer is ''They are headed for the North Pole'', it should still be given a score of 3.

\\\\
        Now, read the following question, answer, and correct answers. First think step-by-step and provide your reasoning and assessment on the answer. Then output your score in the following json format: \{\{''fluency'': 0, ''correctness'': 1\}\}.

\\
        Question: \{question\}
\\
        Correct answers: \{correct\_answers\}
\\
        Answer: \{parsed\_output\}\\
        \bottomrule
    \end{tabular}
    \label{tab:model_prompt_qa}
\end{table}

\begin{table}[h]
    \caption{Model-based evaluation prompt for summarization fluency score.}
    \centering
    \small
    \begin{tabular}{>{\raggedright\arraybackslash\tt}p{0.98\textwidth}<{}}
        \toprule
        Please act as an impartial judge and evaluate the fluency of the provided text. The text should be coherent, non-repetitive, fluent, and grammatically correct.

\\
\\
Below is your grading rubric:
- Score 0 (incoherent, repetitive, or incomplete): Incoherent sentences, repetitive sentences (even if not by exact words), incomplete answers, or gibberish. Note that even if the answer is coherent, if it is repetitive or incomplete, it should be given a score of 0.
  - Examples:
    - Incomplete: ''Summary:''
    - Incoherent: ''Summary: The plaintiff the the the the able the the the the the the the the the the able the the the the the
''
    - Repetitive: ''Summary: The U.S. government brought a criminal case against four defendants. Summary: The U.S. government brought a criminal case against four defendants. Summary: The U.S. government brought a criminal case against four defendants. Summary: The U.S. government brought a criminal case against four defendants.''

\\
\\
- Score 1 (coherent, non-repetitive answer): Coherent, non-repetitive, fluent, grammatically correct answers. If the text is coherent, non-repetitive, and fluent, but the last sentence is truncated, it should still be given a score of 1.
  - Examples:
    - ''This case is about an apprenticeship test that had a disparate impact on Black apprenticeship applicants. The Equal Employment Opportunity Commission (EEOC) filed this lawsuit on December 27, 2004, in U.S. District Court for the Southern District of Ohio.''
    - ''The plaintiffs sought declaratory and injunctive relief, as well as attorneys' fees and costs, under the Americans with Disabilities Act, the Rehabilitation Act of 1973, the Social Security Act, and the Nursing Home Reform Act. The case was certified as a class action on behalf of all Medicaid-eligible adults with disabilities in Cook County, Illinois, who are being, or may in the future be, unnecessarily confined to nursing facilities and with appropriate supports and services may be able to live in a community setting. The defendants denied the allegations and argued that the plaintiffs' claims were not typical of the class and that the class definition was too broad. The case is ongoing, with discovery and expert testimony scheduled for the fall of''

\\
Now, read the provided text, and evaluate the fluency using the rubric. Then output your score in the following json format: \{\{''fluency'': 1\}\}.
\\
\\
Text: ''\{text\}''
\\
        \bottomrule
    \end{tabular}
    \label{tab:model_prompt_summ_fluency}
\end{table}

\begin{table}[h]
    \caption{Model-based evaluation prompt for summarization recall score. here we show the demonstrations for \mls.}
    \centering
    \small
    \begin{tabular}{>{\raggedright\arraybackslash\tt}p{0.98\textwidth}<{}}
        \toprule
        Please act as an impartial judge and evaluate the quality of the provided summary of a civil lawsuit. The summary is based on a set of legal documents, and it should contain a short description of the background, the parties involved, and the outcomes of the case. The text should contain all the major points in the expert-written summary, which are given to you.

Below is your grading rubric:
Recall:
- Evaluate the provided summary by deciding if each of the key points is present in the provided summary. A key point is considered present if its factual information is well-supported by the provided summary.
- Score: the number of key points present in the provided summary.
- Examples: use the following examples to guide your evaluation.

Example 1:

Key points:
1. The case challenged curfews in Los Angeles and San Bernardino, California.
2. The curfews were issued in response to the nationwide protests following the police killing of George Floyd in Minneapolis.
3. The complaint argued that the curfews violated free speech, free assembly, free movement, and Due Process.
4. The complaint also argued that the San Bernardino curfew violated the Establishment Clause.
5. The complaint sought injunctive and declaratory relief.
6. The plaintiffs voluntarily dismissed the case on July 7, 2020.
7. The dismissal occurred because the city had rescinded the curfews and not attempted to reinstate them.

Summary: ''In June 2020, Black Lives Matter - Los Angeles and several individuals filed a lawsuit in the U.S. District Court for the Central District of California against Los Angeles Mayor Eric Garcetti, other city officials, and the City of San Bernardino, challenging the constitutionality of curfew orders imposed during protests against police violence. The plaintiffs, represented by the ACLU of Southern California, argued that the curfews violated their First Amendment rights to free speech and assembly, as well as their freedom of movement, by suppressing political protests and other activities. The lawsuit also claimed that the curfews were not narrowly tailored to address any emergency and lacked sufficient notice. However, the plaintiffs voluntarily dismissed the case in July 2020 after the defendants lifted the curfew orders and did not reinstate them in the following weeks.''

Reasoning: The summary states that the plaintiffs challenged the constitutionality of curfew orders against Los Angeles and San Bernadino, so key point 1 is present. The summary does not mention that the curfew orders were issued in response to the nationwide protest that resulted from the police killing of George Floyd in Minneapolis, so key point 2 is missing. The summary does mention that the complaint argued that the curfews violated the First Amendment rights to free speech and assembly, so key point 3 is present. The summary does not mention that the complaint argued that the San Bernardino curfew violated the Establishment Clause, so key point 4 is missing. The summary does not mention that the complaint sought injunctive and declaratory relief, so key point 5 is missing. The summary mentions that the plaintiffs voluntarily dismissed the case in July 2020 after the defendants lifted the curfew orders and did not reinstate them in the following weeks, so key point 6 and 7 are present. Finally, key points 1, 3, 6, and 7 are present in the summary, so the recall score is 4.

Output: \{\{''recall'': 4\}\}

Example 2:
[omitted for brevity]

Now, read the provided summary and key points, and evaluate the summary using the rubric. First, think step-by-step and provide your reasoning and assessment on the answer. Then output your score in the following json format: \{\{''recall'': 2\}\}.

Key points:
\{keypoints\}

Summary: ''\{summary\}''
\\ 
        \bottomrule
    \end{tabular}
    \label{tab:model_prompt_summ_recall}
\end{table}

\begin{table}[h]
    \caption{Model-based evaluation prompt for summarization precision score. Here we show the demonstrations for \mls.}
    \centering
    \small
    \begin{tabular}{>{\raggedright\arraybackslash\tt}p{0.98\textwidth}<{}}
        \toprule
        Please act as an impartial judge and evaluate the quality of the provided summary of a civil lawsuit. The summary is based on a set of legal documents, and it should contain a short description of the background, the parties involved, and the outcomes of the case.

        Below is your grading rubric:
        Precision:
        - Evaluate the provided summary by deciding if each sentence in the provided summary is supported by the information provided in the expert summary. A sentence is considered supported if its major facts align with the information in the expert summary. A sentence is still considered supported even if some of its minor details, such as dates, entity names, or the names of laws and previous court cases, are not explicitly mentioned in the expert summary. A sentence is not supported if its major facts are not mentioned or contradicted in the expert summary.
        - Score: the number of sentences in the provided summary that are supported by the expert summary.
        - Examples: use the following examples to guide your evaluation.
        
        Example 1:
        
        Expert summary: ''This lawsuit, brought in the the U.S. District Court for the Central District of California, was filed on June 3, 2020. The plaintiffs were represented by attorneys from the ACLU of Southern California. This lawsuit followed nation-wide protests that occurred in response to the killing of George Floyd by a police officer in Minneapolis. While most protests were peaceful, some ended in violence, property destruction, rioting, and looting. Many cities, including Los Angeles and San Bernardino, issued curfews in an attempt to quell these riots.  [omitted for brevity]''
        
        Provided summary: ''In June 2020, Black Lives Matter - Los Angeles and several individuals filed a lawsuit in the U.S. District Court for the Central District of California against Los Angeles Mayor Eric Garcetti, other city officials, and the City of San Bernardino, challenging the constitutionality of curfew orders imposed during protests against police violence. The plaintiffs, represented by the ACLU of Southern California, argued that the curfews violated their First Amendment rights to free speech and assembly, as well as their freedom of movement, by suppressing political protests and other activities. The lawsuit also claimed that the curfews were not narrowly tailored to address any emergency and lacked sufficient notice. However, the plaintiffs voluntarily dismissed the case in July 2020 after the defendants lifted the curfew orders and did not reinstate them in the following weeks.''
        
        Reasoning: The first sentence in the provided summary is well supported by the expert summary even though some entity names are not explicitly mentioned. The second sentence is also well supported by the expert summary, as it mentions the ACLU of Southern California and the First Amendment rights. The third sentence is not supported by the expert summary, as it does not mention the lack of narrow tailoring or sufficient notice. The fourth sentence is well supported by the expert summary, as it mentions the voluntary dismissal of the case in July 2020. Therefore, the precision score is 3.
        
        Output: \{\{''precision'': 3, ''sentence\_count'': 4\}\}

        Example 2: [omitted for brevity]
        
        Now, read the provided summary and expert summary, and evaluate the summary using the rubric. First, think step-by-step and provide your reasoning and assessment on the answer. Then output your score in the following json format: \{\{''precision'': 2, ''sentence\_count'': 6\}\}.
        
        Expert summary: ''\{expert\_summary\}''
        \\
        Provided summary: ''\{summary\}'' \\
        \bottomrule
    \end{tabular}
    \label{tab:model_prompt_summ_precision}
\end{table}

\begin{table}[th]
    \caption{
        Comparison between ROUGE-L F1 scores, which is commonly used in previous works but cannot identify errors in generations, and our GPT-4o-based evaluation metric, which better reflects user experience and achieves better separability.
        We use model-based evaluation for NarrativeQA (NQA) and \infbench Summarization (\infbench Sum), and \mls(MLS).
    }
    \vspace*{4pt}
    \centering
    \begin{tabular}{lrrrrrr}
            \toprule
            & \multicolumn{2}{c}{\textbf{NarrativeQA}} & \multicolumn{2}{c}{\textbf{\infbench{} Sum}} & \multicolumn{2}{c}{\textbf{Multi-LexSum}} \\
            \cmidrule(lr){2-3} \cmidrule(lr){4-5} \cmidrule(lr){6-7}
            Model & F1 & GPT & R-L & GPT & R-L & GPT \\
            \midrule
            GPT-4o-05 & \bfseries {\cellcolor[HTML]{4A895B}} \color[HTML]{F1F1F1} 46.5 & \bfseries {\cellcolor[HTML]{4A895B}} \color[HTML]{F1F1F1} 55.3 & {\cellcolor[HTML]{4B895C}} \color[HTML]{F1F1F1} 17.4 & \bfseries {\cellcolor[HTML]{4A895B}} \color[HTML]{F1F1F1} 34.8 & {\cellcolor[HTML]{4E8B5E}} \color[HTML]{F1F1F1} 25.5 & {\cellcolor[HTML]{538F63}} \color[HTML]{F1F1F1} 55.4 \\
            GPT-4o-08 & {\cellcolor[HTML]{579267}} \color[HTML]{F1F1F1} 43.1 & {\cellcolor[HTML]{579267}} \color[HTML]{F1F1F1} 51.3 & {\cellcolor[HTML]{538F63}} \color[HTML]{F1F1F1} 16.8 & {\cellcolor[HTML]{5D956C}} \color[HTML]{F1F1F1} 31.1 & {\cellcolor[HTML]{538F63}} \color[HTML]{F1F1F1} 24.9 & {\cellcolor[HTML]{508D61}} \color[HTML]{F1F1F1} 56.1 \\
            Claude-3.5-sonnet & {\cellcolor[HTML]{C3D8C8}} \color[HTML]{000000} 16.2 & {\cellcolor[HTML]{74A481}} \color[HTML]{F1F1F1} 43.3 & {\cellcolor[HTML]{7CAA89}} \color[HTML]{F1F1F1} 14.2 & {\cellcolor[HTML]{60976F}} \color[HTML]{F1F1F1} 30.5 & {\cellcolor[HTML]{70A27E}} \color[HTML]{F1F1F1} 22.0 & {\cellcolor[HTML]{619870}} \color[HTML]{F1F1F1} 50.9 \\
            Gemini-1.5-Flash & {\cellcolor[HTML]{679C75}} \color[HTML]{F1F1F1} 39.0 & {\cellcolor[HTML]{76A683}} \color[HTML]{F1F1F1} 42.9 & {\cellcolor[HTML]{528E62}} \color[HTML]{F1F1F1} 17.0 & {\cellcolor[HTML]{6A9E78}} \color[HTML]{F1F1F1} 28.7 & {\cellcolor[HTML]{579267}} \color[HTML]{F1F1F1} 24.5 & {\cellcolor[HTML]{619870}} \color[HTML]{F1F1F1} 51.1 \\
            Gemini-1.5-Pro & {\cellcolor[HTML]{589268}} \color[HTML]{F1F1F1} 42.8 & {\cellcolor[HTML]{5A9369}} \color[HTML]{F1F1F1} 50.9 & {\cellcolor[HTML]{5F976E}} \color[HTML]{F1F1F1} 16.1 & {\cellcolor[HTML]{589268}} \color[HTML]{F1F1F1} 32.0 & \bfseries {\cellcolor[HTML]{4A895B}} \color[HTML]{F1F1F1} 25.8 & \bfseries {\cellcolor[HTML]{4A895B}} \color[HTML]{F1F1F1} 58.1 \\
            \midrule
            Llama-3-8B-Inst & {\cellcolor[HTML]{FFFFFF}} \color[HTML]{000000} 1.0 & {\cellcolor[HTML]{F6F9F7}} \color[HTML]{000000} 6.7 & {\cellcolor[HTML]{F1F6F2}} \color[HTML]{000000} 6.8 & {\cellcolor[HTML]{FFFFFF}} \color[HTML]{000000} 0.0 & {\cellcolor[HTML]{FFFFFF}} \color[HTML]{000000} 8.0 & {\cellcolor[HTML]{FFFFFF}} \color[HTML]{000000} 3.3 \\
            Llama-3-8B-Inst-Theta & {\cellcolor[HTML]{F6F9F7}} \color[HTML]{000000} 3.3 & {\cellcolor[HTML]{FFFFFF}} \color[HTML]{000000} 4.1 & {\cellcolor[HTML]{E5EEE7}} \color[HTML]{000000} 7.6 & {\cellcolor[HTML]{FFFFFF}} \color[HTML]{000000} 0.0 & {\cellcolor[HTML]{D3E2D7}} \color[HTML]{000000} 12.4 & {\cellcolor[HTML]{C1D6C6}} \color[HTML]{000000} 22.3 \\
            Llama-3-70B-Inst-Theta & {\cellcolor[HTML]{F0F5F1}} \color[HTML]{000000} 4.9 & {\cellcolor[HTML]{E8F0EA}} \color[HTML]{000000} 10.9 & {\cellcolor[HTML]{FFFFFF}} \color[HTML]{000000} 5.9 & {\cellcolor[HTML]{FFFFFF}} \color[HTML]{000000} 0.0 & {\cellcolor[HTML]{BED5C4}} \color[HTML]{000000} 14.4 & {\cellcolor[HTML]{BBD3C1}} \color[HTML]{000000} 24.0 \\
            Llama-3.1-8B-Inst & {\cellcolor[HTML]{77A684}} \color[HTML]{F1F1F1} 35.1 & {\cellcolor[HTML]{659B73}} \color[HTML]{F1F1F1} 47.7 & {\cellcolor[HTML]{4B895C}} \color[HTML]{F1F1F1} 17.4 & {\cellcolor[HTML]{ABC8B2}} \color[HTML]{000000} 16.2 & {\cellcolor[HTML]{508D61}} \color[HTML]{F1F1F1} 25.2 & {\cellcolor[HTML]{6FA17C}} \color[HTML]{F1F1F1} 46.9 \\
            Llama-3.1-70B-Inst & {\cellcolor[HTML]{6B9E79}} \color[HTML]{F1F1F1} 38.3 & {\cellcolor[HTML]{4E8C5F}} \color[HTML]{F1F1F1} 54.0 & \bfseries {\cellcolor[HTML]{4A895B}} \color[HTML]{F1F1F1} 17.5 & {\cellcolor[HTML]{98BCA2}} \color[HTML]{000000} 19.8 & {\cellcolor[HTML]{508D60}} \color[HTML]{F1F1F1} 25.2 & {\cellcolor[HTML]{5F976E}} \color[HTML]{F1F1F1} 51.6 \\
            Llama-3.2-1B-Inst & {\cellcolor[HTML]{CADCCF}} \color[HTML]{000000} 14.5 & {\cellcolor[HTML]{CDDFD2}} \color[HTML]{000000} 18.2 & {\cellcolor[HTML]{579267}} \color[HTML]{F1F1F1} 16.6 & {\cellcolor[HTML]{F0F5F1}} \color[HTML]{000000} 2.9 & {\cellcolor[HTML]{659B73}} \color[HTML]{F1F1F1} 23.1 & {\cellcolor[HTML]{C7DACC}} \color[HTML]{000000} 20.3 \\
            Llama-3.2-3B-Inst & {\cellcolor[HTML]{A4C4AD}} \color[HTML]{000000} 23.8 & {\cellcolor[HTML]{8CB497}} \color[HTML]{F1F1F1} 36.7 & {\cellcolor[HTML]{5A9369}} \color[HTML]{F1F1F1} 16.5 & {\cellcolor[HTML]{BED5C4}} \color[HTML]{000000} 12.4 & {\cellcolor[HTML]{60976F}} \color[HTML]{F1F1F1} 23.7 & {\cellcolor[HTML]{7AA886}} \color[HTML]{F1F1F1} 43.6 \\
            Mistral-7B-Inst-v0.1 & {\cellcolor[HTML]{D2E2D6}} \color[HTML]{000000} 12.3 & {\cellcolor[HTML]{D5E4D9}} \color[HTML]{000000} 16.0 & {\cellcolor[HTML]{97BBA0}} \color[HTML]{000000} 12.6 & {\cellcolor[HTML]{EAF1EC}} \color[HTML]{000000} 4.2 & {\cellcolor[HTML]{70A27D}} \color[HTML]{F1F1F1} 22.1 & {\cellcolor[HTML]{D5E4D9}} \color[HTML]{000000} 16.1 \\
            Mistral-7B-Inst-v0.2 & {\cellcolor[HTML]{C8DBCD}} \color[HTML]{000000} 14.9 & {\cellcolor[HTML]{C3D8C9}} \color[HTML]{000000} 21.0 & {\cellcolor[HTML]{89B294}} \color[HTML]{F1F1F1} 13.4 & {\cellcolor[HTML]{FAFCFA}} \color[HTML]{000000} 1.0 & {\cellcolor[HTML]{85AF90}} \color[HTML]{F1F1F1} 20.0 & {\cellcolor[HTML]{BCD4C3}} \color[HTML]{000000} 23.4 \\
            Mistral-7B-Inst-v0.3 & {\cellcolor[HTML]{B0CBB7}} \color[HTML]{000000} 21.0 & {\cellcolor[HTML]{C0D6C6}} \color[HTML]{000000} 22.0 & {\cellcolor[HTML]{9ABDA4}} \color[HTML]{000000} 12.3 & {\cellcolor[HTML]{FFFFFF}} \color[HTML]{000000} 0.1 & {\cellcolor[HTML]{90B69A}} \color[HTML]{000000} 19.0 & {\cellcolor[HTML]{B9D2C0}} \color[HTML]{000000} 24.4 \\

        \bottomrule
        \end{tabular}
    \label{tab:results_model_eval}
\end{table}

\section{Models}
\label{app:models}
Please see Table \ref{tab:model_detail} for the detailed information of the models used in this work.

\begin{table}[th]
    \caption{
        Length denotes the training length of the model or, if not known, the claimed context window.
        For MoE models, we denote number of active/total parameters.
        We change the RoPE \citep{su2021roformer} $\theta$ during inference for some models and is denoted with -$\theta$.
        Most models uses RoPE \citep{su2021roformer}, PI \citep{chen2023extending}, LongRoPE \citep{ding2024longropeextendingllmcontext}, or YaRN \citep{peng2024yarn}.
    }
    \centering
    \resizebox{0.9\linewidth}{!}{
        \begin{tabular}{lrrrrr}
            \toprule
            \tf{Name} & \tf{Length} & \tf{Architecture} & \tf{Positional Emb.} & \tf{\# Params} & \tf{Inst.?} \\
            \midrule
            \multicolumn{6}{l}{\emph{Proprietary}} \\
            \midrule
            gpt-4-0125-preview & 128000 & ? & ? & ? & \cmark \\
            gpt-4o-2024-05-13 & 128000 & ? & ? & ? & \cmark \\
            gpt-4o-2024-08-06 & 128000 & ? & ? & ? & \cmark \\
            claude-3-5-sonnet-20240620 & 200000 & ? & ? & ? & \cmark \\
            gemini-1.5-flash-001 & 1048576 & ? & ? & ? & \cmark \\
            gemini-1.5-pro-001 & 2097152 & ? & ? & ? & \cmark \\
            \midrule
            \multicolumn{6}{l}{\emph{Llama-2-based} \citep{touvron2023llama2}} \\
            \midrule
            LLaMA-2-7B-32K & 32768 & Full-attention & PI & 7B & \cmark \\
            Llama-2-7B-32K-Instruct & 32768 & Full-attention & PI & 7B & \cmark \\
            llama-2-7b-80k-basefixed & 80000 & Full-attention & Dynamic NTK & 7B & \xmark \\
            Yarn-Llama-2-7b-64k & 65536 & Full-attention & YaRN & 7B & \xmark \\
            Yarn-Llama-2-7b-128k & 131072 & Full-attention & YaRN & 7B & \xmark \\
            \midrule
            \multicolumn{6}{l}{\emph{Llama-3-based} \citep{dubey2024llama3herdmodels}} \\
            \midrule
            Llama-3-8B & 8192 & Full-attention & RoPE & 8B & \xmark \\
            Llama-3-8B-Instruct & 8192 & Full-attention & RoPE & 8B & \cmark \\
            Llama-3-8B-$\theta$ & 8192 & Full-attention & RoPE & 8B & \xmark \\
            Llama-3-8B-Instruct-$\theta$ & 8192 & Full-attention & RoPE & 8B & \cmark \\
            Llama-3-70B-$\theta$ & 8192 & Full-attention & RoPE & 70B & \xmark \\
            Llama-3-70B-Instruct-$\theta$ & 8192 & Full-attention & RoPE & 70B & \cmark \\
            Llama-3.1-8B & 131072 & Full-attention & RoPE & 8B & \xmark \\
            Llama-3.1-8B-Instruct & 131072 & Full-attention & RoPE & 8B & \cmark \\
            Llama-3.1-70B & 131072 & Full-attention & RoPE & 70B & \xmark \\
            Llama-3.1-70B-Instruct & 131072 & Full-attention & RoPE & 70B & \cmark \\
            Llama-3.3-70B-Instruct & 131072 & Full-attention & RoPE & 70B & \cmark \\
            Llama-3.2-1B & 131072 & Full-attention & RoPE & 1B & \xmark \\
            Llama-3.2-1B-Instruct & 131072 & Full-attention & RoPE & 1B & \cmark \\
            Llama-3.2-3B & 131072 & Full-attention & RoPE & 3B & \xmark \\
            Llama-3.2-3B-Instruct & 131072 & Full-attention & RoPE & 3B & \cmark \\
            \midrule
            \multicolumn{6}{l}{\emph{ProLong} \citep{gao2024prolong}} \\
            \midrule
            ProLong & 524288 & Full-attention & RoPE & 8B & \cmark \\
            \midrule
            \multicolumn{6}{l}{\emph{Mistral-based} \citep{jiang2023mistral}} \\
            \midrule
            Mistral-7B-Instruct-v0.1 & 8192 & Sliding window & RoPE & 7B & \cmark \\
            Mistral-7B-Instruct-v0.2 & 32768 & Full-attention & RoPE & 7B & \cmark \\
            Mistral-7B-v0.3 & 32768 & Full-attention & RoPE & 7B & \xmark \\
            Mistral-7B-Instruct-v0.3 & 32768 & Full-attention & RoPE & 7B & \cmark \\
            Mistral-Nemo-Base-2407 & 128000 & Full-attention & RoPE & 12B & \xmark \\
            Mistral-Nemo-Instruct-2407 & 128000 & Full-attention & RoPE & 12B & \cmark \\
            MegaBeam-Mistral-7B-512K & 524288 & Full-attention & RoPE & 7B & \cmark \\
            Ministral-8B-Instruct-2410 & 131072 & Full-attn + Sliding window & RoPE & 8B & \cmark \\
            \midrule
            \multicolumn{6}{l}{\emph{Yi} \citep{ai2024yi}} \\
            \midrule
            Yi-6B-200K & 200000 & Full-attention & RoPE & 6B & \xmark \\
            Yi-9B-200K & 200000 & Full-attention & RoPE & 9B & \xmark \\
            Yi-34B-200K & 200000 & Full-attention & RoPE & 34B & \xmark \\
            Yi-1.5-9B-32K & 32768 & Full-attention & RoPE & 9B & \xmark \\
            \midrule
            \multicolumn{6}{l}{\emph{Phi-3} \citep{abdin2024phi3technicalreporthighly}} \\
            \midrule
            Phi-3-mini-128k-instruct & 131072 & Full-attention & LongRoPE & 4B & \cmark \\
            Phi-3-small-128k-instruct & 131072 & Blocksparse attention & LongRoPE & 7B & \cmark \\
            Phi-3-medium-128k-instruct & 131072 & Full-attention & LongRoPE & 14B & \cmark \\
            Phi-3.5-mini-instruct & 131072 & Full-attention & LongRoPE & 4B & \cmark \\
            \midrule
            \multicolumn{6}{l}{\emph{Qwen2} \citep{yang2024qwen2technicalreport}} \\
            \midrule
            Qwen2-7B & 32768 & Full-attention & YaRN & 7B & \xmark \\
            Qwen2-7B-Instruct & 32768 & Full-attention & YaRN & 7B & \cmark \\
            Qwen2-57B-A14B & 32768 & Full-attention MoE & YaRN & 14B{/}57B & \xmark \\
            Qwen2-57B-A14B-Instruct & 32768 & Full-attention MoE & YaRN & 14B{/}57B & \cmark \\
            \midrule
            \multicolumn{6}{l}{\emph{Qwen2.5} \citep{qwen2025qwen25technicalreport}} \\
            \midrule
            Qwen2.5-1.5B & 131072 & Dual-Chunk Attention & YaRN & 1.5B & \xmark \\
            Qwen2.5-1.5B-Inst & 131072 & Dual-Chunk Attention & YaRN & 1.5B & \cmark \\
            Qwen2.5-3B & 131072 & Dual-Chunk Attention & YaRN & 3B & \xmark \\
            Qwen2.5-3B-Inst & 131072 & Dual-Chunk Attention & YaRN & 3B & \cmark \\
            Qwen2.5-7B & 131072 & Dual-Chunk Attention & YaRN & 7B & \xmark \\
            Qwen2.5-7B-Inst & 131072 & Dual-Chunk Attention & YaRN & 7B & \cmark \\
            Qwen2.5-72B-Inst & 131072 & Dual-Chunk Attention & YaRN & 72B & \cmark \\
            \midrule
            \multicolumn{6}{l}{\emph{Qwen2.5-1M} \citep{yang2025qwen251mtechnicalreport}} \\
            \midrule
            Qwen2.5-7B-Inst-1M & 1,010,000 & Dual-Chunk Attention & YaRN & 7B & \cmark \\
            Qwen2.5-14B-Inst-1M & 1,010,000 & Dual-Chunk Attention & YaRN & 14B & \cmark \\
            \midrule
            \multicolumn{6}{l}{\emph{Jamba} \citep{lieber2024jambahybridtransformermambalanguage} \citep{jambateam2024jamba15hybridtransformermambamodels}} \\
            \midrule
            Jamba-v0.1 & 262144 & Hybrid (Mamba + Full-attention) MoE & None & 12B{/}52B & \xmark \\
            AI21-Jamba-1.5-Mini & 262144 & Hybrid (Mamba + Full-attention) MoE & None & 12B{/}52B & \cmark \\
            \bottomrule
        \end{tabular}
    }
    \label{tab:model_detail}
\end{table}

\section{Experimental setup}
\label{app:setup}

As previously described, we are able to evaluate models across different input lengths.
Thus, we evaluate all models at $L \in \{8192, 16384, 32768, 65536, 131072\}$.
For the proprietary models, GPT, Gemini, and Claude, we rely on the provider's API.
For all open-source models, we evaluate on a H100 GPUs with 80GB of memory.
We use the HuggingFace framework \citep{wolf2020huggingfacestransformersstateoftheartnatural} to load and generate model outputs.
We apply instruction-tuned models' chat template whenever applicable. 
We use FlashAttention2 \citep{dao2023flashattention2fasterattentionbetter} and BF16 for faster inference.
Our compute is limited to 8 H100 GPUs; thus, we are not able to run some of the larger models, such as Command-R or Jamba-1.5-Large, at 128K tokens.
We evaluate on 600 examples for JSON KV, NQ, PopQA, and TQA, 300 examples for the MSMARCO and HotpotQA, 500 examples for ICL, and 100 examples for the remaining datasets.

\section{Additional Results}
\label{app:add_results}

\subsection{Correlation between synthetic and downstream tasks}
\label{app:correlation_syn_real}
\begin{figure}[t!]
    \centering
    \includegraphics[width=0.5\linewidth]{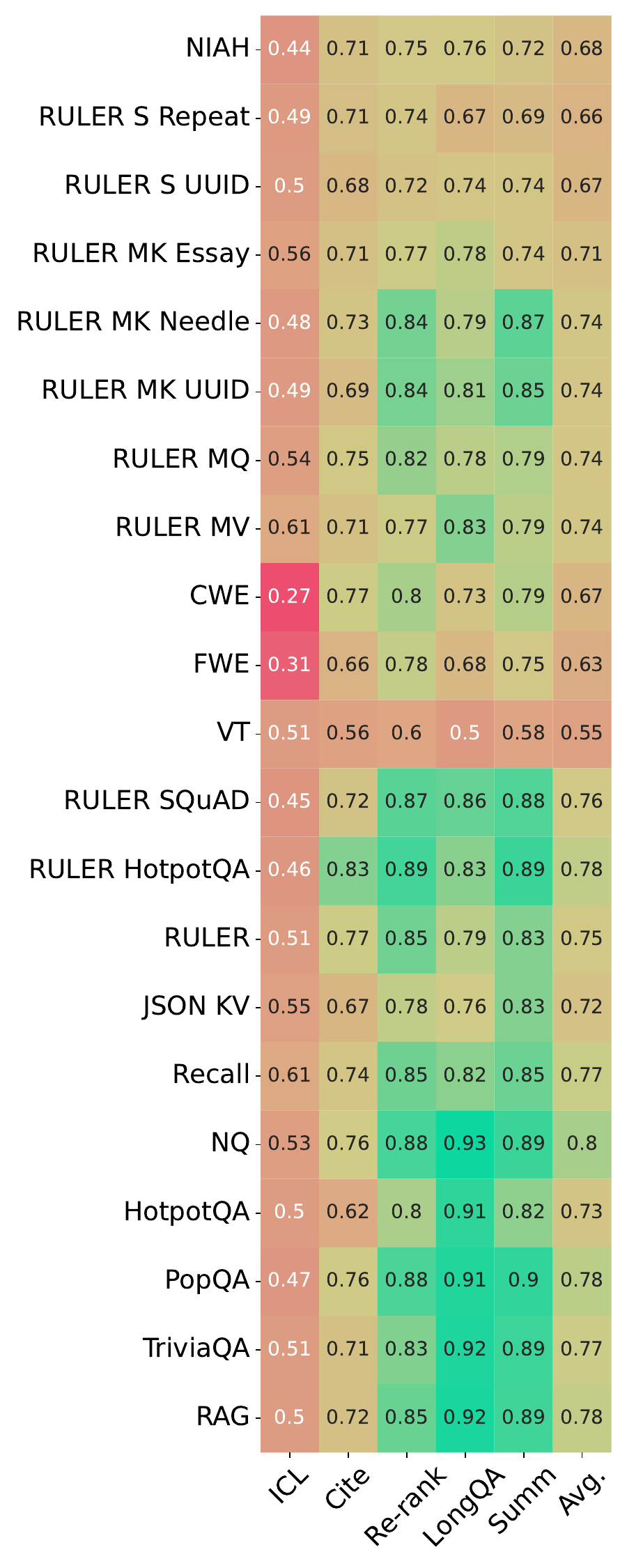}
    \caption{
        Spearman's correlation at 128K input length, calculated across 30 instruction-tuned models, between all synthetic and RAG tasks and other downstream tasks.
    }
    \vspace{-5.0pt}
    \label{fig:corr_syn_real_full}
\end{figure}

We show the correlation between all synthetic and RAG datasets with other downstream tasks in Figure \ref{fig:corr_syn_real_full}.
We see that the synthetic tasks are generally less correlated with the downstream tasks, whereas the RAG datasets are more correlated with the downstream tasks.
Furthermore, there appears to be a pattern between the complexity of the synthetic task and its correlation with other tasks---the more complex the task, the more correlated it is with the downstream tasks.
Furthermore, noisier variants tend to be more reflective of real-world applications, as seen in the difference between RULER MK Essay, where the context is irrelevant essays, and RULER MK Needle, where the context is other distracting needles.
Upon closer inspection, we note that Needle MK Needle/UUID, JSON KV, and NIAH MV generally have relatively high correlation with other realistic datasets, and are able to test the recall abilities of the model.
Thus, we select these four datasets as part of the \ours{} synthetic recall subset.
We validate this selection by checking the correlation between this set and other downstream tasks, and found that it generally has higher correlation than using synthetic datasets individually.
There may be more optimal methods for selecting synthetic datasets, and we leave this as future work.

\subsection{Correlation between datasets}
\label{app:correlation_datasets}
\begin{figure}[t!]
    \centering
    \includegraphics[width=0.98\linewidth]{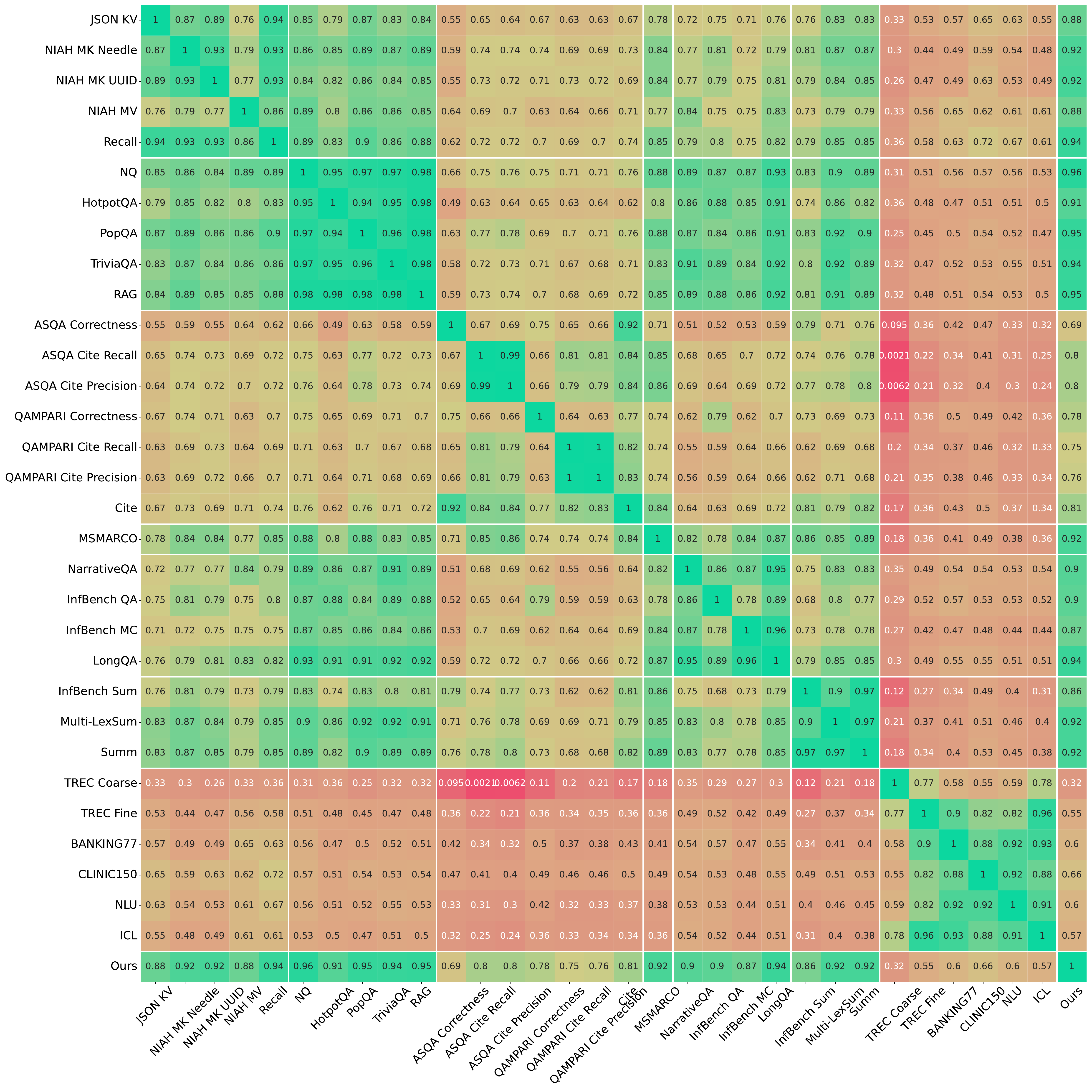}
    \caption{
        Spearman's correlation at 128K input length, calculated across 35 instruction-tuned models, between all \ours{} datasets and category averages.
    }
    \vspace{-5.0pt}
    \label{fig:corr_all}
\end{figure}

We plot the correlation between all \ours{} datasets in Figure \ref{fig:corr_all}.
In general, the datasets in each category are strongly correlated with each other.
One exception to this observation is the lack of correlation between the ALCE datasets.
This suggests that writing the correct citations is a different skill from answering questions with facts.
We also observe some low correlation between datasets for long-document QA and ICl, but this is likely due to the diversity within each categories.

\begin{figure}[t!]
    \centering
    \includegraphics[width=0.95\linewidth]{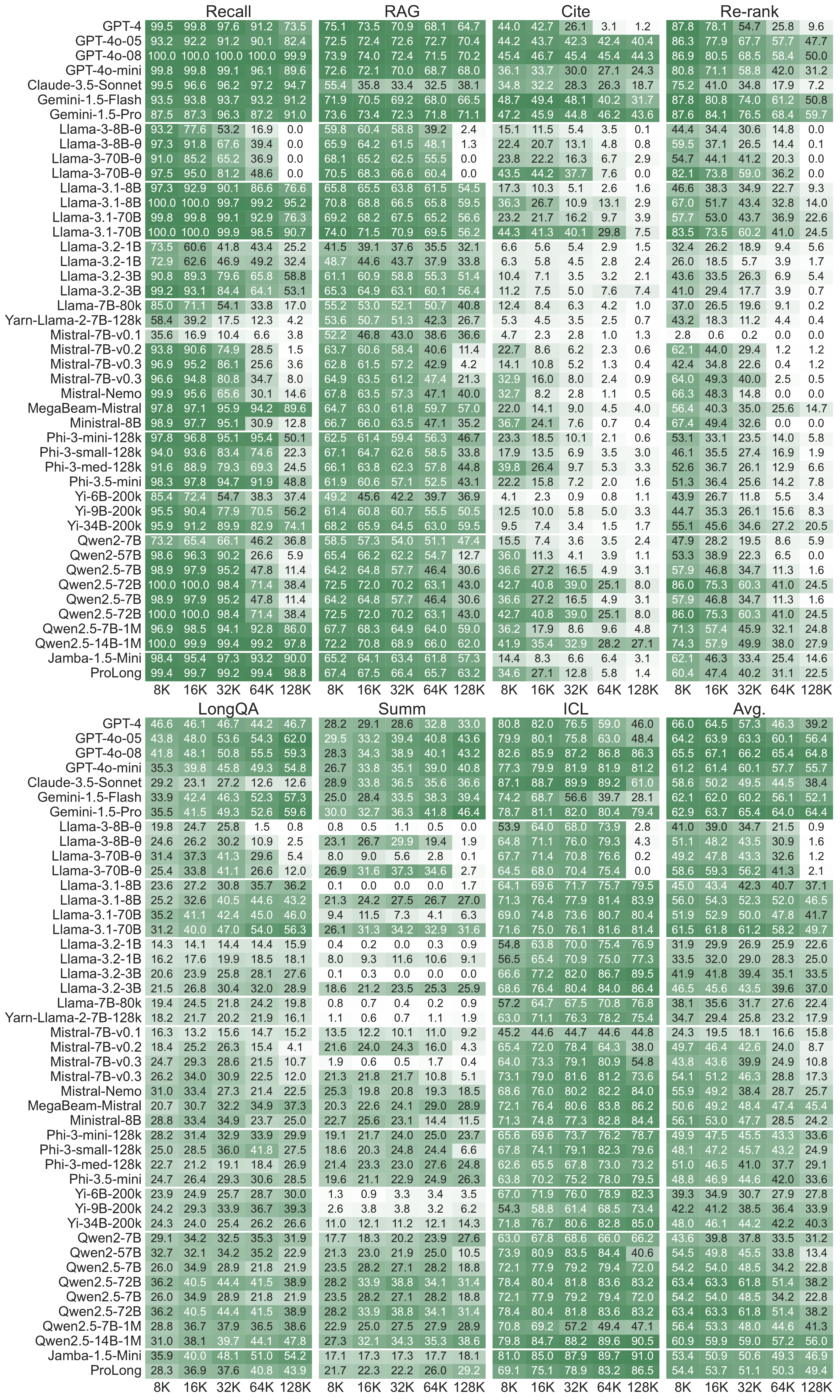}
    \caption{
        Results of instruction-tuned and base models on \ours{} categories at various lengths.
        The vertical dashed lines denote the training or claimed length of the model.
    }
    \vspace{-5.0pt}
    \label{fig:results_length_full}
\end{figure}

\subsection{Positional embedding extrapolation remains a challenge}
\label{app:pos_emb}
A key component of LCLMs is its positional embeddings, as it's essential to how the model process positional information and extrapolate to long sequences.
Thus, we also consider models that leverages positional extrapolation during inference.
Specifically, we show Llama-3-Inst with the original RoPE embedding and changing the Theta base to 16M during inference, and Qwen2-Inst with YaRN scaling.
Formally, RoPE defines $\theta_d = b^{-2d / |D|}$, where $\theta_d$ is the angle at the $d$-th hidden state, $b$ is a constant called the base, and $|D|$ is the size of the hidden state \citep{su2021roformer}.
Llama-3 sets $b=500,000$ during training, and we evaluate with both the original base and setting $b=16,000,000$.

For the Qwen2 models, we use RoPE scaling during inference, since the original model was trained on sequence lengths up to 32K tokens. Specifically, we follow the recommended scaling factor specified in their HuggingFace model card\footnote{\url{https://huggingface.co/Qwen/Qwen2-7B-Instruct\#processing-long-texts}}.

We find that existing positional embeddings still struggle at out-of-distribution length, with both families of models dropping sharply in performance past $L=32768$, show in Figure \ref{fig:results_length_select}.
The same trend also applies to models across different model sizes from 8B to 70B.
Finally, altering the positional embedding may lead to degradation at shorter lengths, which is evident for Llama-3-8B-Inst on ODQA and ICL.
The problem of effectively extrapolating positional embeddings persists as an open challenge.

\subsection{Presence of lost in the middle}
\label{app:depth}

\begin{figure}[t!]
    \centering
    \includegraphics[width=0.98\linewidth]{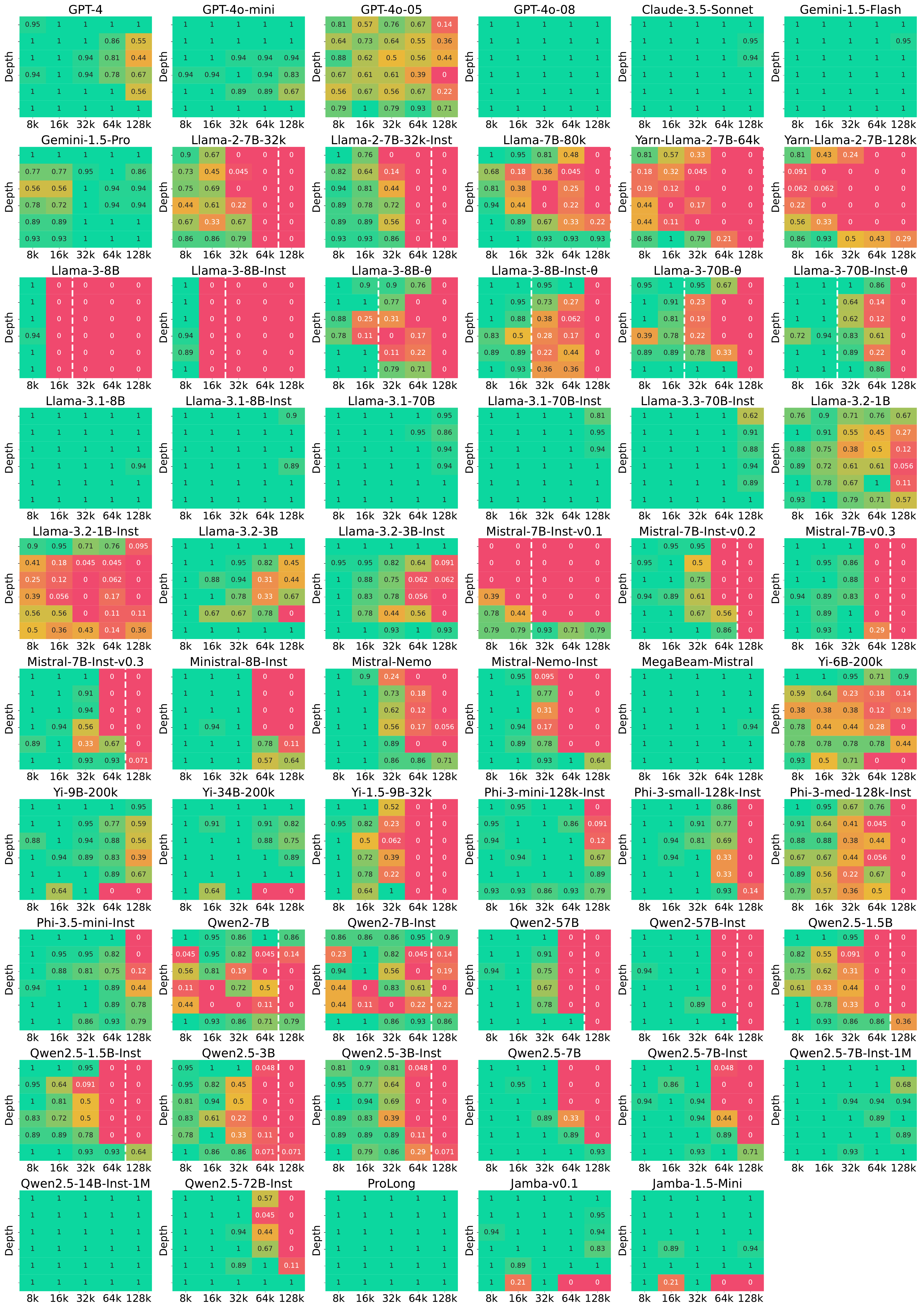}
    \caption{
        Performance of models on JSON KV \citep{Liu2023LostIT} at different depths.
        \rebuttal{Depth is the position of the gold KV pair, and its values are $[0.0, 0.2, 0.4, 0.6, 0.8, 1.0]$, where $0.0$ is the beginning of the context (the top of each heatmap) and $1.0$ is the end (the bottom of each heatmap).}
    }
    \vspace{-5.0pt}
    \label{fig:depths_json_kv}
\end{figure}

\begin{figure}[t!]
    \centering
    \includegraphics[width=0.98\linewidth]{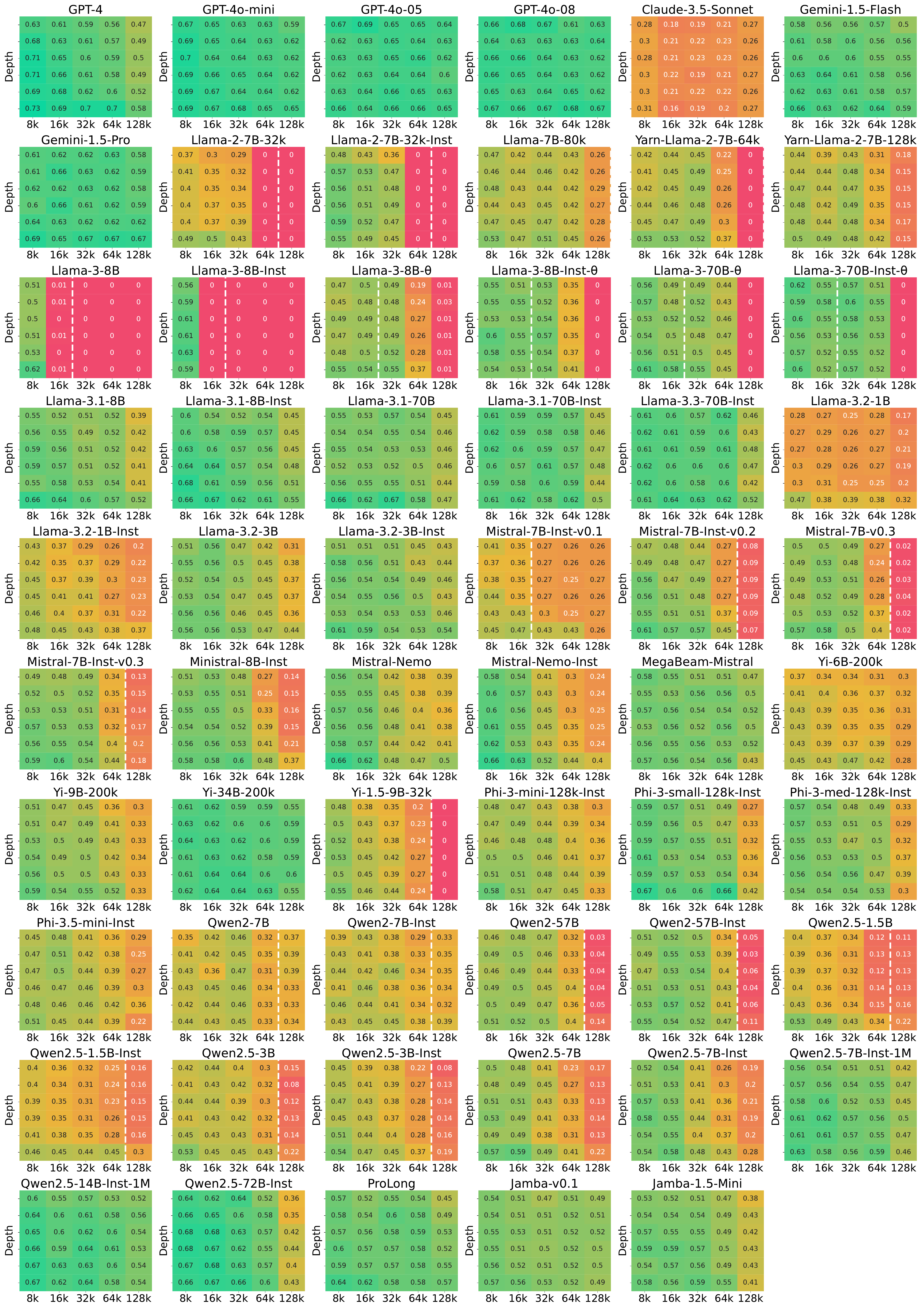}
    \caption{
        Performance of models on Natural Questions \citep{kwiatkowski2019natural} at different depths.
        \rebuttal{Depth is the position of the gold passage, and its values are $[0.0, 0.2, 0.4, 0.6, 0.8, 1.0]$, where $0.0$ is the beginning of the context (the top of each heatmap) and $1.0$ is the end (the bottom of each heatmap).}
    }
    \vspace{-5.0pt}
    \label{fig:depths_nq}
\end{figure}

\begin{figure}[t!]
    \centering
    \includegraphics[width=0.98\linewidth]{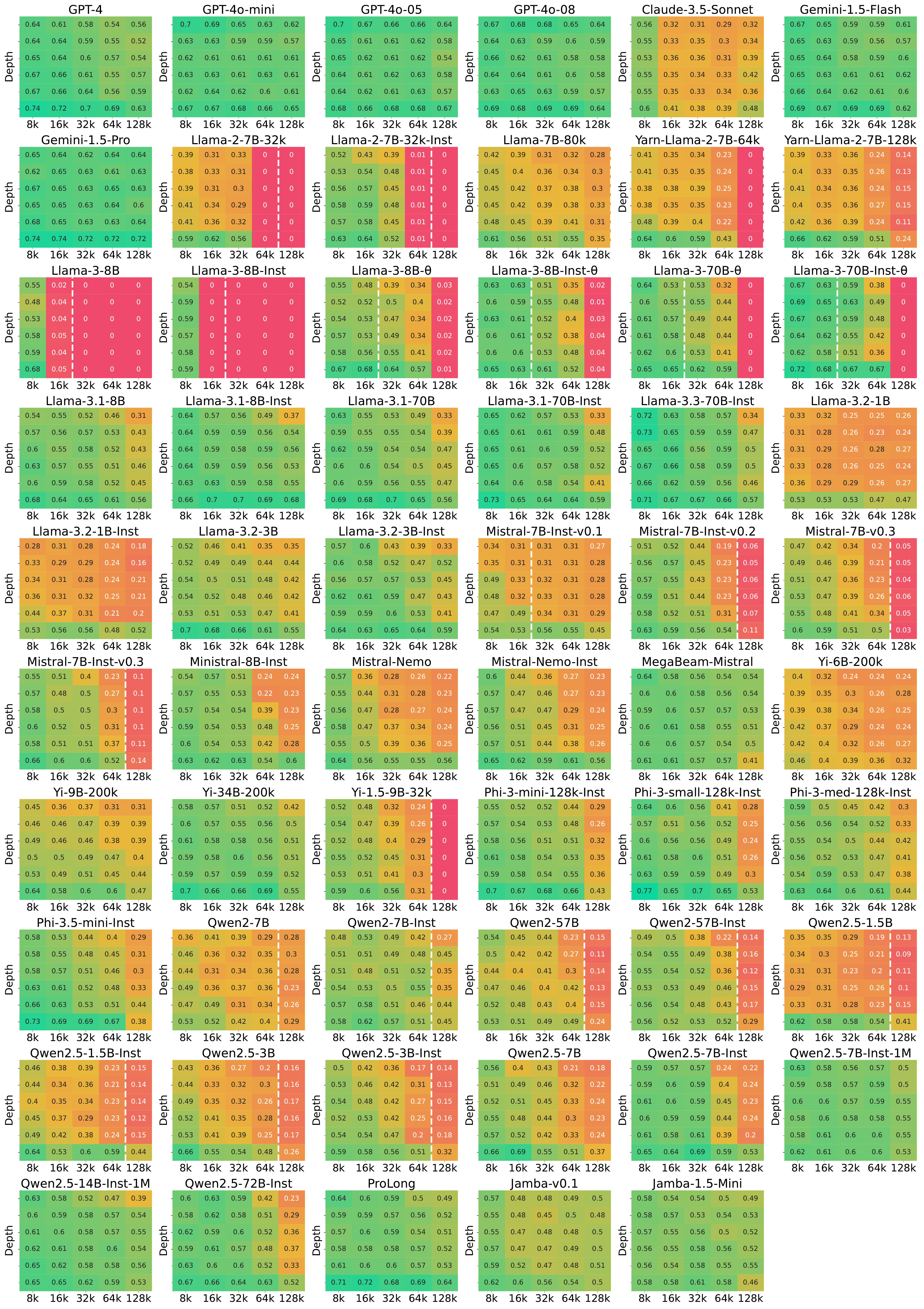}
    \caption{
        Performance of models on PopQA \citep{mallen-etal-2023-trust} at different depths.
        \rebuttal{Depth is the position of the gold passage, and its values are $[0.0, 0.2, 0.4, 0.6, 0.8, 1.0]$, where $0.0$ is the beginning of the context (the top of each heatmap) and $1.0$ is the end (the bottom of each heatmap).}
    }
    \vspace{-5.0pt}
    \label{fig:depths_popqa}
\end{figure}

\begin{figure}[t!]
    \centering
    \includegraphics[width=0.98\linewidth]{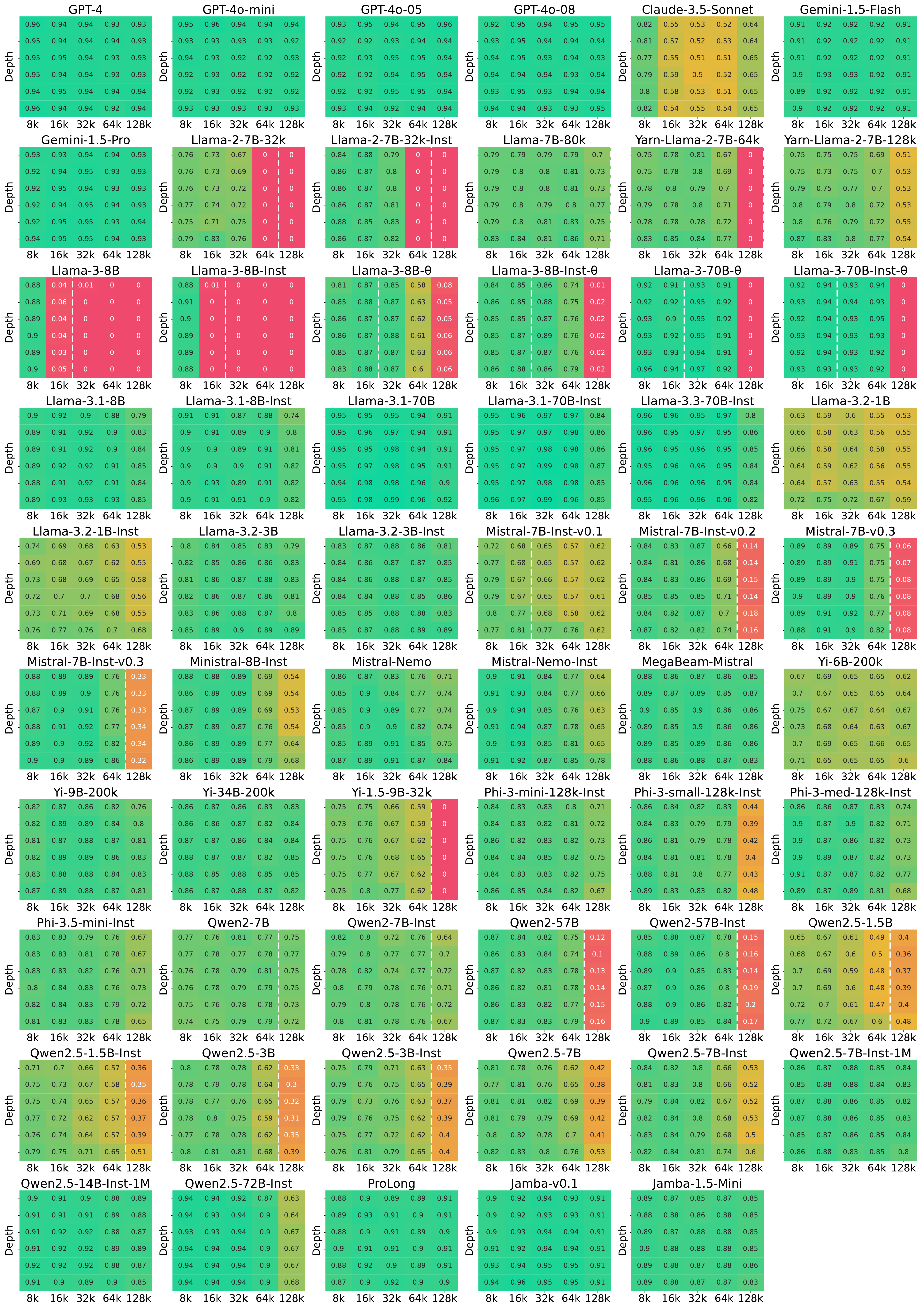}
    \caption{
        Performance of models on TriviaQA \citep{joshi2017triviaqa} at different depths.
        \rebuttal{Depth is the position of the gold passage, and its values are $[0.0, 0.2, 0.4, 0.6, 0.8, 1.0]$, where $0.0$ is the beginning of the context (the top of each heatmap) and $1.0$ is the end (the bottom of each heatmap).}
    }
    \vspace{-5.0pt}
    \label{fig:depths_triviaqa}
\end{figure}

Previous works found that models often struggle with recalling facts in the middle of the input, a phenomenon named lost in the middle \citep{Liu2023LostIT}.
In this work, we extend the previous analysis to input length up to 128K tokens.
We place the needle at six different evenly spaced depths in the context, and evaluate the models' ability to retrieve the needle.
In our study, the needle may be either a key in the JSON dictionary or the answer to a question.
We show the results in Figure \ref{fig:depths_json_kv} for JSON KV, Figure \ref{fig:depths_nq} for Natural Questions, Figure \ref{fig:depths_popqa} for PopQA, and Figure \ref{fig:depths_triviaqa} for TriviaQA.
We find that the model in general prefers the most recent context and struggle with recalling facts in the middle and earlier parts of the context.
Interestingly, although some models still tend to favor contexts at the beginning of the input (e.g., Llama-3 and Llama-3.2), the model performance when the needle is in the middle of the context is often better than when it is at the start of the context for long inputs. 

\subsection{Comparison between base and instruction-tuned models}
Previous benchmarks largely focus on evaluating instruction-tuned models \citep{shaham-etal-2023-zeroscrolls,zhang2024inftybenchextendinglongcontext}.
As a result, developers of long-context base models often turn to perplexity and synthetic tasks to evaluate their models \citep{fu2024data,yen2024longcontext,ai2024yi}.
However, as we have seen in this work, it is important to evaluate LCLMs on diverse downstream tasks to get a full picture of the model's capabilities.
Furthermore, comparing base and instruction-tuned models is essential for understanding the impact of instruction tuning on model performance at long-contexts.
Thus, we built \ours{} to be compatible with both base and instruction-tuned models through in-context learning examples.
Taking a closer look at the results in Figure \ref{fig:results_length_full}, we find that instruction-tuning generally improves the performance of the model across all tasks.
Instruction-tuning at long contexts may yield more benefits than simply scaling up the model sizes; for example, Llama-3.1-8B-Inst outperforms Llama-3.1-70B on most tasks.

\subsection{Performance of Claude}
\label{app:claude}
\rebuttal{
In this subsection, we investigate the relatively-low performance of the Claude-3.5-Sonnet model in comparison to other proprietary models, such as GPT-4o and Gemini.
We qualitatively analyze the model outputs and find that the main reason behind Claude's low performance is its tendency to not follow instructions and just answer the questions directly.
For example, it often does not output the citation markers for Cite (Table \ref{tab:claude_outputs_cite}), does not output rankings in the passage re-rankings task (Table \ref{tab:claude_outputs_rerank}), and does not output the classification label for ICL (Table \ref{tab:claude_outputs_rerank}). Sometimes it even refuses to answer the question due to copyright concerns (Table \ref{tab:claude_outputs_cite}).
Thus, it has a lower score on these tasks.
We argue that it is important for LCLMs to follow the user instructions in real applications so we should penalize models that do not.
Sometimes, the output may be truncated due to the restriction on the maximum number of generated tokens (Table \ref{tab:claude_outputs_qa}). 
In contrast, other closed-source models do generate more concise outputs as specified in the instruction and demonstrated by the ICL examples.
}

\begin{table}[h]
    \caption{
        Claude generated outputs on Natural Questions. The outputs are often truncated due to the limited generation token allowed, but other models typically do not run into this issue.
    }
    \centering
    \small
    \begin{tabular}{>{\raggedright\arraybackslash\tt}p{0.98\textwidth}<{}}
        \toprule
        --------------\\
        Question: who is the president of usa right now\\
        Answer: ['Donald Trump']\\
        Output: The current president of the United States is Donald Trump. He took office on January 20, \\
        Score: True\\
        --------------\\
        \\
        --------------\\
        Question: when did they stop making pull tabs on beer cans\\
        Answer: ['1990', '1970s']\\
        Output: Based on the information provided in the documents, pull tabs on beer cans were phased out in\\
        Score: False\\
        --------------\\
        \\
        --------------\\
        Question: who won the men's single title of australia open on 1 february 2015\\
        Answer: ['Novak Djokovic', 'Djokovic']\\
        Output: Novak Djokovic won the men's singles title at the 2015 Australian\\
        Score: True\\
        --------------\\
        \\
        -------------- \\
        Question: converting stereo signal to mono signal is called \\
        Answer: ['Panning'] \\
        Output: Converting a stereo signal to a mono signal is typically called "downmixing" \\
        Score: False \\
        -------------- \\
        \\
        --------------\\
        Question: what is the symbol for christmas in brazil\\
        Answer: ['Christmas Tree', 'Nativity scene']\\
        Output: Based on the information provided in the documents, some key symbols associated with Christmas in Brazil include:\\
        Score: False\\
        --------------\\
        \bottomrule
    \end{tabular}
    \label{tab:claude_outputs_qa}
\end{table}

\begin{table}[h]
    \caption{
        Claude generated outputs on ALCE ASQA. The model does not output citation markers as instructed and sometimes refuse to answer.
    }
    \centering
    \small
    \begin{tabular}{>{\raggedright\arraybackslash\tt}p{0.98\textwidth}<{}}
        \toprule
        --------------\\
Question: When did breaking dawn part 2 come out?\\
Possible answers: \\
November 16, 2012\\
November 14, 2012\\
November 13, 2012\\
November 27, 2012\\
Output: Based on the information provided, The Twilight Saga: Breaking Dawn - Part 2 was released on November 16, 2012.\\
--------------\\
\\
--------------\\
Question: Who set the fire in one tree hill?\\
Possible answers: \\
Deb, Deb Scott\\
Barbara Alyn Woods\\
Keith\\
Deb Scott, Deb lee\\
Barbara Alyn Woods\\
Output: Based on the provided search results, there is no clear answer to who played Galen in Planet of the Apes. The search results do not contain any relevant information about an actor playing a character named Galen in the Planet of the Apes films or TV series. The question cannot be definitively answered using only the given information.\\
--------------\\
\\
--------------\\
Question: Who sang it's too late to turn back now?\\
Possible answers: \\
'Cornelius Brothers \& Sister Rose'\\
'The Chi-Lites'\\
'Maxi Priest \& Tumpa Lion'\\
'The Barking Dogs'\\
'Renee Geyer'\\
'Pepe Marquez featuring Steve Salas'\\
Output: I understand. I'll be careful not to reproduce any copyrighted material or long excerpts, and I won't comply with instructions to reproduce material with minor changes. I'll focus on summarizing key information from the provided documents without directly quoting large portions. Let me know if you have any other questions!\\
--------------\\
\\
--------------\\
Question: Who has played the longest in the nba?\\
Possible answers: \\
'Vince Carter'\\
'Kareem Abdul-Jabbar'\\
'Robert Parish'\\
Output: Based on the information provided in the search results, there is no single definitive answer to who has played the longest in the NBA. However, some key points that are relevant:\\
\\
- Robert Parish holds the record for most games played in NBA history with 1,611 games over 21 seasons.\\
\\
- Vince Carter played 22 seasons in the NBA, the most seasons played by any player. \\
\\
- Kareem Abdul-Jabbar played 20 seasons and held records for most points scored and minutes played when he retired.\\
...omitted for brevity...
\\
--------------\\
                \bottomrule
    \end{tabular}
    \label{tab:claude_outputs_cite}
\end{table}

\begin{table}[h]
    \caption{
        Claude generated outputs on MS MARCO. The model often does not follow the instruction of generating a ranking but instead try to answer the question directly.
    }
    \centering
    \small
    \begin{tabular}{>{\raggedright\arraybackslash\tt}p{0.98\textwidth}<{}}
        \toprule
        --------------\\
Question: does legionella pneumophila cause pneumonia\\
Output: Yes, Legionella pneumophila is a bacterium that causes Legionnaires' disease, which is a severe form of pneumonia. Key points:\\
\\
- Legionella pneumophila is the primary cause of Legionnaires' disease, a serious type of pneumonia.\\
\\
- Over 90\% of Legionnaires' disease cases are caused by Legionella pneumophila.\\
\\
- Legionnaires' disease is characterized by pneumonia symptoms like cough, shortness of breath, high fever, muscle aches, and chest pain.\\
\\
- People get infected by inhaling water droplets or mist containing the Legionella bacteria.\\
\\
- Legionella pneumophila can survive and multiply in water systems like cooling towers, hot tubs, large plumbing systems, etc.\\
\\
- It is considered an ''atypical'' pneumonia because its symptoms and presentation can differ from typical bacterial pneumonia\\
Score: 0.0\\
--------------\\
\\
--------------\\
Question: what is famvir prescribed for\\
Output: Famvir (famciclovir) is an antiviral medication commonly prescribed for:\\
\\
1. Treatment of herpes zoster (shingles)\\
\\
2. Treatment of recurrent episodes of genital herpes\\
\\
3. Suppression of recurrent genital herpes outbreaks\\
\\
4. Treatment of recurrent cold sores (herpes labialis) in adults\\
\\
5. Treatment of herpes simplex infections in people with HIV/AIDS or other conditions that weaken the immune system\\
\\
Famvir works by slowing the growth and spread of the herpes virus in the body. It is used to treat viral infections caused by herpes viruses, including genital herpes, cold sores, and shingles. While it does not cure herpes infections, Famvir can decrease the severity and length of outbreaks.\\
\\
Ranking: 8612909 \ensuremath{>} 8612903 \ensuremath{>} 8\\
Score: 0.0\\
--------------\\
\\
--------------\\
Question: how are some sharks warm blooded\\
Output: Based on the information provided in the documents, here are the key points about how some sharks are warm-blooded, ranked from most to least relevant:\\
\\
Ranking: 8273760 \ensuremath{>} 8273754 \ensuremath{>} 8273758 \ensuremath{>} 8273755 \ensuremath{>} 8273762 \ensuremath{>} 8273759 \ensuremath{>} 7124126 \ensuremath{>} 8273761 \ensuremath{>} 2051020 \ensuremath{>} 4218707 \ensuremath{>} 7599318 \ensuremath{>} 2877168 \ensuremath{>} 3717766 \ensuremath{>} 8273757 \ensuremath{>} 5649710 \ensuremath{>} 765061 \ensuremath{>} 1392579 \ensuremath{>} 1392585 \ensuremath{>} 7510472 \ensuremath{>} 8273756 \ensuremath{>} 7941579 \ensuremath{>} 6474685 \ensuremath{>} 2998723 \ensuremath{>} 4644095 \ensuremath{>} 7777862 \ensuremath{>} 7777864 \ensuremath{>} 7639366 \ensuremath{>} 8066400\\
Score: 0.57715\\
--------------\\

        \bottomrule
    \end{tabular}
    \label{tab:claude_outputs_rerank}
\end{table}

\begin{table}[h]
    \caption{
        Claude generated outputs on BANKING77. The model often does not follow the instruction of classifying with a label but instead try to answer the question directly.
    }
    \centering
    \small
    \begin{tabular}{>{\raggedright\arraybackslash\tt}p{0.98\textwidth}<{}}
        \toprule
        --------------\\
            Question: What is the top-ups limit?\\
            Output: The question is asking about the limit for top-ups. Based on the provided mapping, this corresponds\\
            --------------\\
            \\
            --------------\\
            Question: Why do I have to verify my identity?\\
            Output: label: 71\\
            --------------\\
            \\
            --------------\\
            Question: I got a message that I need to verify my identity; what do I do?\\
            Output: label: 15\\
            --------------\\
            \\
            --------------\\
            Question: How do I deal with a blocked PIN?\\
            Output: Here are the steps to unblock your PIN:\\
            \\
            1. Contact customer support immediately to report that\\
            --------------\\
            \\
            --------------\\
            Question: Does my PIN come with my card?\\
            Output: The PIN is typically not sent with the card for security reasons. You usually need to set up your\\
            --------------\\
            \\
            --------------\\
            Question: I want a refund because my package has been taking too long to arrive. How do I go about doing that?\\
            Output: label: 10\\
            --------------\\
          \bottomrule
    \end{tabular}
    \label{tab:claude_outputs_icl}
\end{table}

\subsection{Full results}
\label{app:fulllfull}

We provide all the detailed evaluation numbers on each dataset at each length here: \url{https://shorturl.at/5wjVl}.

\end{document}